\documentclass[preprint,12pt]{elsarticle}

\usepackage{amssymb}
\usepackage{url}
\usepackage{hyperref}

\usepackage{graphicx}
\usepackage{multirow}
\usepackage{adjustbox}
\usepackage{pdflscape}
\usepackage{geometry}
\usepackage{lineno}
\usepackage{tabularx}
\usepackage{array}
\usepackage{makecell}
\usepackage{algorithm}
\usepackage{algorithmic}

\usepackage{listings}
\usepackage{float}

\usepackage{xcolor}
\usepackage{caption}
\usepackage{booktabs}
\usepackage{amsmath}

\journal{}

\begin{document}
\begin{frontmatter}



\title{Automatic identification of diagnosis from hospital discharge letters via weakly supervised Natural Language Processing}


\author[inst1]{Vittorio Torri\corref{cor1}}

\affiliation[inst1]{organization={MOX - Modelling and Scientific Computing Lab, Department of Mathematics, Politecnico di Milano},
            city={Milan},
            country={Italy}}
\cortext[cor1]{Corresponding Author. \\Accepted for publication in \textit{Scientific Reports}. The final version is available at DOI: \href{https://doi.org/10.1038/s41598-026-56721-0}{10.1038/s41598-026-56721-0}.}

\author[inst2]{Elisa Barbieri}
\author[inst3,inst4]{Anna Cantarutti}

\affiliation[inst2]{organization={Division of Pediatric Infectious Diseases, Department for Woman and Child Health, University of Padova},
            city={Padova},
            country={Italy}}

\affiliation[inst3]{organization={Unit of Biostatistics, Epidemiology and Public Health, Department of Statistics and Quantitative Methods, University of Milano-Bicocca},
            city={Milan},
            country={Italy}}

\affiliation[inst4]{organization={National Centre for Healthcare Research and Pharmacoepidemiology, Department of Statistics and Quantitative Methods, University of Milano-Bicocca},
            city={Milan},
            country={Italy}}
            
\author[inst2]{Carlo Giaquinto}

\author[inst1,inst4,inst5]{Francesca Ieva}

\affiliation[inst5]{organization={HDS - Health Data Science Centre, Human Technopole},
            city={Milan},
            country={Italy}}

\begin{abstract}
Identifying patient diagnoses from hospital discharge letters is essential for large-scale cohort selection and epidemiological research, but traditional supervised approaches require extensive manual annotation, which is often impractical for large textual datasets. We present a weakly supervised Natural Language Processing (NLP) pipeline for classifying Italian discharge letters without document-level manual annotation. The method extracts diagnosis-related sentences, generates semantic embeddings using a transformer model further pre-trained on Italian medical documents, and applies a two-level clustering procedure to derive weak labels that are then used to train a document-level classifier.

The approach was evaluated in a case study on bronchiolitis using 33,176 discharge letters of children admitted to 44 emergency rooms or hospitals in the Veneto Region, Italy, between 2017 and 2020. The best weakly supervised model achieved an AUROC of 77.68\% ($\pm4.30\%$), an AUPRC of 73.13\% ($\pm4.93\%$), and an F1-score of 78.14\% ($\pm4.89\%$) against manually annotated data. Performance surpassed unsupervised baselines and approached fully supervised models, while reducing the need for manual annotation by more than 1,500 hours for a dataset of this size. Similar model rankings were observed in a secondary validation on a smaller bronchitis dataset (3,188 discharge letters, 2020-2025), where the best weakly supervised model achieved an AUPRC of 76.72\% ($\pm 5.02\%$).

These results suggest the potential of weakly supervised NLP methods for scalable disease identification from clinical discharge letters.
\end{abstract}

\begin{keyword}
Natural Language Processing \sep weakly-supervised learning \sep discharge letters \sep diagnosis identification
\MSC 68T50
\end{keyword}

\end{frontmatter}

\section*{Introduction}
\label{sec:introduction}
Electronic Health Records (EHR) contain a vast and continuously growing amount of patient information, but a significant portion of this data remains unstructured in formats like clinical notes, reports, discharge letters and referrals. Despite containing valuable information, this unstructured data is often underutilized in the downstream analysis due to their complexity and heterogeneity~\cite{tayefi2021challenges,cowie2017electronic}. While Natural Language Processing (NLP) has made significant advancements in extracting and classifying information from text~\cite{nadkarni2011natural, khurana2023natural}, applying these techniques to clinical narratives remains challenging. Medical documents are characterized by a highly specialized lexicon, with ambiguous abbreviations and frequent misspellings, a lack of standard templates across hospitals and a limited amount of manually labelled data~\cite{hossain2023natural, friedman2014natural, iroju2015systematic}. Additionally, most NLP tools are optimized for high-resource languages such as English, limiting their transferability to languages like Italian, where datasets and domain-specific linguistic resources are scarce~\cite{neveol2018clinical}. 

Among the various types of clinical documents, hospital discharge letters play a particularly important role: they summarize the patient's condition, diagnoses, and treatments, representing a key source of diagnostic information for both clinical follow-up and population-based studies. Although international standards such as the ICD-CM~\cite{world2016international} exist for encoding diagnoses, physicians often prefer to provide textual descriptions of their diagnoses because they are faster to write, more expressive, and often the only option available in document templates~\cite{walsh2004clinician, millares2021consultation}. Even when ICD-CM codes are included, previous research has shown that they can be incomplete or inaccurate~\cite{walsh2018icd, cozzolino2018accuracy, cipparone2015inaccuracy}.
Automatic methods capable of identifying diagnoses directly from unstructured discharge letters are therefore crucial for cohort selection and epidemiological research~\cite{dong2022automated}, reducing the need for manual review, which is labor-intensive and impractical at scale.

In Italy, every hospital or emergency room (ER) admission generates a discharge letter that should contain at least one diagnosis describing the reason for admission or discharge. However, these letters vary widely in format and terminology, often mixing narrative and structured components. Developing robust NLP systems for such data is challenging: traditional approaches rely on supervised classification models that require a substantial amount of manually annotated data~\cite{mehrafarin2022importance}, an effort that is costly and time-consuming for clinicians. This highlights a significant gap in the field: the need for methods that can effectively exploit large, unannotated datasets to extract clinically meaningful information without the burden of manual labelling.

Previous studies applying NLP to Italian clinical texts have not addressed discharge letters. Lanera et al.~\cite{lanera2020use, lanera2022deep} worked on supervised models to identify varicella cases from notes of family pediatricians, while Sciannameo et al.~\cite{veronica2022deep} used a BERT-based model to predict rehospitalizations for elderly patients from medication prescriptions and hospitalization records. Hammami et al.~\cite{hammami2021automated} classified pathology reports for cancer morphology using a rule-based approach, and Viani et al. \cite{viani2018information, viani2019supervised} extracted entities from cardiology reports through ontologies and recurrent neural networks.
Most international work on discharge letters has been conducted in English using resources such as the MIMIC database~\cite{johnson2016mimic, johnson2023mimic} (see \cite{edin2023automated, li2020icd, li2018automated, falter2024using} for some examples), but research has also been conducted in other European languages with limited resources, such as Bulgarian~\cite{boytcheva2011automatic, velichkov2020automatic}, Dutch~\cite{bagheri2020automatic, van2024using}, Czech~\cite{lenc2024czech} and Swedish~\cite{remmer2021multi}.
These studies, however, all relied on large annotated datasets, making them difficult to reproduce in settings where manual labelling is infeasible.

Recent work has explored weakly supervised approaches to mitigate the need for large annotated clinical corpora. Many of these methods rely on task-specific heuristics, such as disease-related keywords, regular expressions, or the transfer of labels from structured data like diagnostic codes or medication records~\cite{sanyal2022weakly,wang2019clinical,cusick2021using,wang2022leveraging,humbert2022strategies, greco2026weakly}. In parallel, the EHR phenotyping literature has proposed frameworks that combine surrogate signals extracted from structured and textual data to infer patient-level disease status, including approaches such as PheNorm~\cite{yu2018enabling}, sureLDA~\cite{ahuja2020surelda}, Anchor~\cite{halpern2016electronic}, Polar~\cite{wagholikar2020polar} and XPRESS~\cite{agarwal2016learning}. These methods typically aggregate information across multiple encounters and rely on structured EHR variables (e.g., diagnosis codes, medications, laboratory tests or concept counts extracted from notes) to construct surrogate features for phenotyping. While effective in large longitudinal EHR datasets, such assumptions are not always satisfied in practice. In particular, structured clinical variables may be incomplete or unavailable, especially in emergency department admissions that do not lead to hospitalization, where discharge summaries often constitute the primary source of diagnostic information. In these settings, the task becomes one of document-level diagnosis identification directly from free-text clinical narratives, which remains comparatively less explored. More recently, large language models (LLMs) have also been used to support weak supervision pipelines~\cite{smith2024language}; however, in clinical settings, effective deployment often still relies on the integration of domain-specific knowledge~\cite{das2025weakly} or the use of proprietary models~\cite{de2025assessing}.

To address these limitations, we propose a weakly supervised NLP pipeline for automatically identifying patient diagnoses directly from Italian hospital discharge letters, operating solely on free-text clinical documents without requiring structured EHR variables or patient-level aggregation.
Our approach eliminates the dependency on document-level manual annotations and offers a more generalizable solution than previous rule-based methods. 

The pipeline first identifies diagnosis-related sentences from a subset of documents and represents them through contextual embeddings generated by a transformer model, previously pre-trained with public data in Italian related to the medical domain. These embeddings are then grouped through a two-level clustering procedure, after which the resulting clusters are semantically summarized and mapped to diseases of interest to produce weak labels. The weakly labelled data are finally used to train a transformer-based classifier capable of recognizing those diseases directly from discharge letters. This design enables large-scale, disease-agnostic classification of clinical narratives while avoiding the costs of large-scale document-level manual labelling.

We evaluate the proposed framework through a real-world case study focused on bronchiolitis, one of the leading causes of hospitalizations (20 \%), emergency department visits (18 \%) and office visits (15 \%) among young children and a major driver of seasonal pediatric respiratory epidemics~\cite{hall2009burden}. The dataset includes 33,176 discharge letters of 15,196 pediatric patients admitted to 44 emergency rooms or hospitals across the Veneto Region (Italy) between January 2017 and December 2020, collected through the Pedianet network~\cite{cantarutti2021pedianet}. To the best of our knowledge, there are no prior studies addressing the identification of bronchiolitis directly from medical text, whereas existing works have relied on structured data for diagnosis prediction~\cite{durani2008clinical,gebremedhin2022developing,vartiainen2023risk}. 

To assess the robustness and generalizability of the proposed pipeline, we additionally conduct a secondary validation study on bronchitis using another smaller dataset from the same Pedianet network. This dataset includes 3,188 discharge letters collected between January 2020 and May 2025, involving 2,364 children living in Veneto Region, accessing 31 different hospitals. Although substantially smaller than the bronchiolitis dataset, this additional cohort provides an independent validation setting to assess whether the proposed weakly supervised framework can generalize to a different respiratory condition and data distribution.

In our evaluation, we test the weakly supervised pipeline against manually labelled data, considering different classification backbones and comparing its performance with both fully unsupervised and fully supervised approaches. The proposed method achieves competitive performance, outperforming other unsupervised techniques and showing only a limited gap compared with supervised classifiers. Its performance is robust to weak labels selection, and the analysis of the intermediate steps indicates a strong potential for generalization across diseases.

\section*{Methods}
\label{sec:methods}

\begin{figure}
    \centering
    \includegraphics[width=\textwidth]{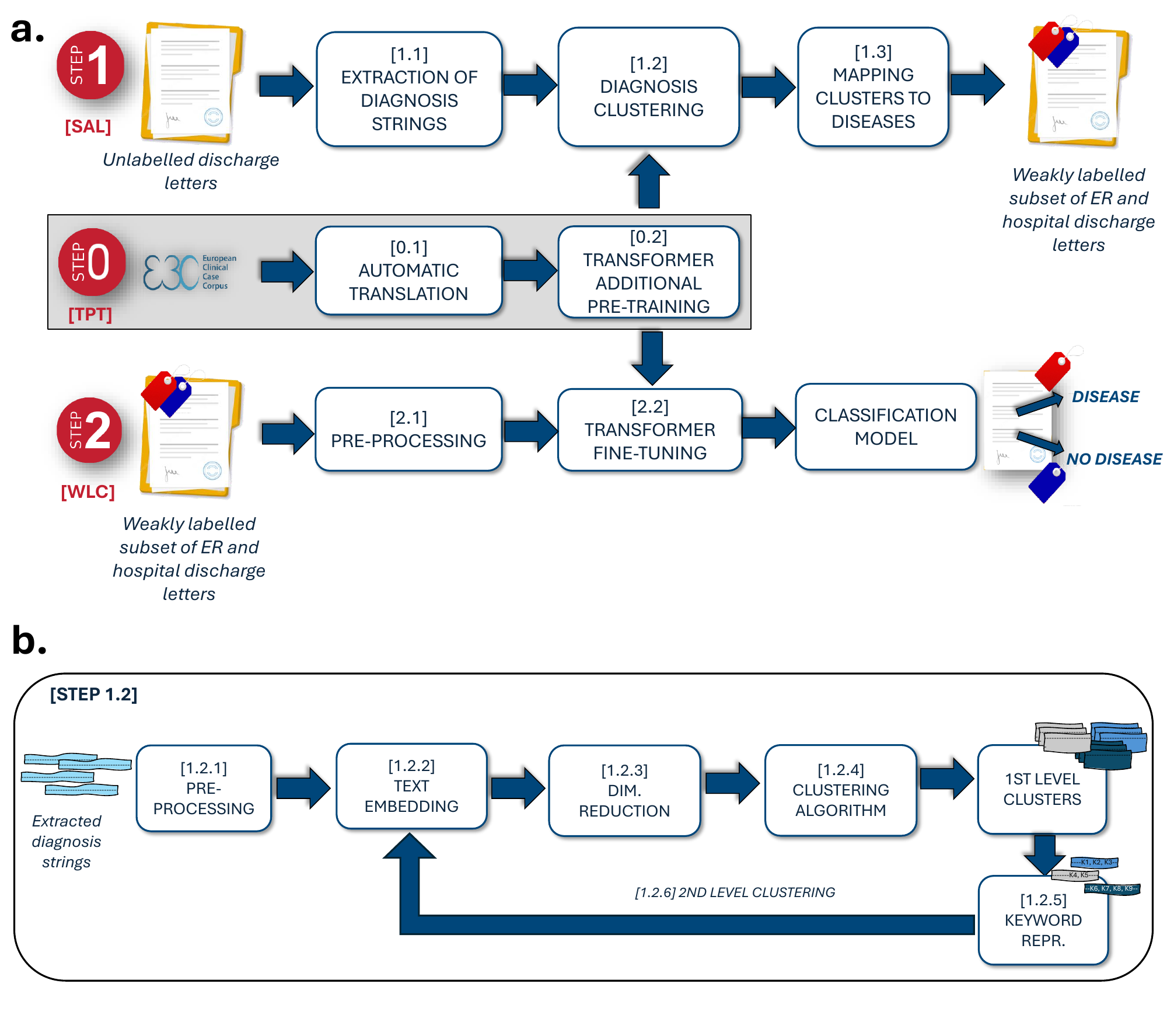}
    \caption{a. Diagram of the full pipeline for weakly supervised classification of discharge letters. Step 0 (TPT) consists in additional pre-training of a transformer-based model on a dataset of medical documents in Italian, including documents auomatically translated into Italian. Step 1 (SAL) is the core of the pipeline, consisting in the automatic extraction of diagnosis strings from the letters, their clustering (exploiting the model from TPT) and the mapping of the clusters to the disease labels. This procedure leads to a weakly labelled subset of data that is used in Step 2 (WLC) to train a weakly supervised classificaton model. b. Detailed diagram of the clustering step for diagnosis strings, corresponding to Step 1.2 of the full pipeline. The extracted diagnoses are embedded with a transformer-based model, and, after a dimensionality reduction step, clustered. Clusters are summarized with relevant keywords and the clustering is repeated on these clusters' representations.}
    \label{fig:global-pipeline}
\end{figure}

In this section, we describe in detail the pipeline we propose for the weakly supervised diagnosis identification. The pipeline, depicted in Figure~\ref{fig:global-pipeline}, can be divided into three main steps:
\begin{enumerate}
    \item[0] \textbf{TPT} - Pre-Training of a Transformer-based model for Italian medical documents.
    \item[1] \textbf{SAL} - Semi-automatic labelling of the dataset.
    \item[2] \textbf{WLC} - Classification of discharge letters using the weakly labelled dataset.
\end{enumerate}

\noindent The next sections describe in detail the three steps. A summary of the main notation used throughout these sections is reported in \ref{app:B}.

\subsection*{Pre-training of a Transformer-based model for Italian medical documents (TPT)}
\label{sec:bert-pretraining}
\noindent Statistical and machine learning models for textual data require a mathematical representation of text. 
Let $\mathcal{X}$ denote the space of textual documents. A text encoder can be formalized as a mapping $f_{\theta}: \mathcal{X} \rightarrow \mathbb{R}^d$, where $f_\theta$ is a parametric function with parameters $\theta$ and $d$ is the embedding dimension. Classical vocabulary-based representations (e.g., bag-of-words or TF--IDF) produce sparse vectors in $\mathbb{R}^{|{V}|}$, where ${V}$ denotes the corpus vocabulary. While effective in certain settings, these representations do not explicitly encode semantic similarity between distinct lexical forms.

Neural network-based embeddings address this limitation. Static word embeddings, such as Word2Vec~\cite{mikolov2013efficient}, define a fixed mapping $w \mapsto v_w \in \mathbb{R}^d$ for each token $w$, where $v_w$ is a dense, continuous vector, used for every occurrence of $w$, and $d \ll |V|$. In contrast, contextual embeddings compute representations that depend on the full input sequence. Transformer-based models~\cite{vaswani2017attention}, including BERT~\cite{devlin2019bert} and RoBERTa~\cite{liu2019roberta}, define contextual encoders where the embedding of each token depends also on the surrounding tokens, capturing different meanings that can depend on the context.

Several transformer models have been specialized for the biomedical domain in English, such as BioBERT~\cite{lee2020biobert} and ClinicalBERT~\cite{alsentzer2019publicly}. For Italian, however, most available transformer models are trained primarily on general-domain corpora~\cite{tamburini2020bertology}, with limited exposure to biomedical or clinical terminology.

To improve domain adaptation, we perform additional pre-training on the European Clinical Case Corpus~(E3C)~\cite{magnini2020e3c}, a multilingual medical dataset containing documents from journal articles, specialty-school admission tests, patient information leaflets, and medical-science thesis abstracts (a summary of its composition is provided in \ref{app:e3c}).

Let $\mathcal{C} = \bigcup_{\ell \in \mathcal{L}} \mathcal{C}_\ell$ denote the full E3C corpus, where $\mathcal{L}$ is the set of languages represented in the dataset and $\mathcal{C}_\ell$ denotes the subset of documents in language $\ell$. Let $\ell = it$ denote Italian. Since the corpus contains documents in multiple languages, we translate all non-Italian subsets into Italian using the Google Translate API. Automatic translation may introduce lexical or syntactic noise. However, the translated documents are used only for masked language modeling during domain-adaptive pre-training. As a result, occasional translation errors are unlikely to systematically bias the downstream task, while the enlarged corpus helps expose the model to a broader range of medical terminology. 

Formally, for $\ell \neq it$, we define $T_\ell : \mathcal{C}_\ell \rightarrow \mathcal{C}_{it}$ (Step 0.1 in Figure~\ref{fig:global-pipeline}.a).

The resulting Italian medical corpus used for domain-adaptive pre-training is
\begin{equation}
\mathcal{C}^{it}
=
\mathcal{C}_{it}
\;\cup\;
\bigcup_{\ell \in \mathcal{L} \setminus \{it\}} T_\ell(\mathcal{C}_\ell).
\end{equation}

Starting from a pre-trained transformer encoder $f_{\theta_0}$, we perform additional masked language modeling (MLM) pre-training on $\mathcal{C}^{it}$. The objective function is
\begin{equation}
\mathcal{L}_{\text{MLM}}(\theta)
=
- \mathbb{E}_{(c,t)}
\log P_\theta(w_t \mid c_{\setminus t}),
\end{equation}
where $w_t$ denotes a masked token in document $c$, and $c_{\setminus t}$ denotes the surrounding context.

The optimized parameters are obtained as $\hat{\theta} = \arg\min_{\theta} \mathcal{L}_{\text{MLM}}(\theta)$ (Step 0.2 in Figure~\ref{fig:global-pipeline} .a), resulting in the domain-adapted encoder $f_{\hat{\theta}} : \mathcal{X} \rightarrow \mathbb{R}^{d}$ where $d = 768$ for the BERT-base encoders that we use. This encoder is subsequently used both in the semi-automatic labelling (SAL, Step 1) and in the weakly supervised classification (WLC, Step 2), which are described in the following sections.

\subsection*{Semi-automatic labelling pipeline (SAL)}
\label{subsec:labelling-pipeline}
Developing a weakly supervised classification pipeline for discharge letters requires a procedure to derive weak labels for at least a subset of data. Weak labels are labels that can be automatically associated to the data but that might contain some noise. 
This section presents a methodology to construct these weak labels by extracting diagnosis strings from discharge letters and clustering them. This allows the creation of groups of diagnoses related to the same disease or symptoms that can be exploited as labels. This step corresponds to Step 1 (SAL) of Figure \ref{fig:global-pipeline}.a.

Our approach combines regular expressions for extracting diagnosis strings and a customized clustering pipeline based on a transformer model. 
In the following sections, we describe in detail these sub-steps.

\subsubsection*{Extraction of diagnosis strings}
\label{subsec:diagnosis-extraction}

The first step (1.1 in Figure~\ref{fig:global-pipeline}.a) consists of extracting diagnosis strings from discharge letters.
Let $\mathcal{D} = \{x_i\}_{i=1}^N$ denote the set of discharge letters, where each $x_i \in \mathcal{X}$ is a textual document.
Italian discharge letters do not follow a standardized template, as each hospital has its own, and even different departments within the same hospital may personalize it further. Nevertheless, we have observed that these templates often include a designated section for diagnoses, where physicians report the conditions related to the current hospitalization. This section may contain one or multiple diagnosis entries (e.g., primary and secondary diagnoses referring to the present admission), which typically contain the information required to determine the document label.

We formalize the extraction step as a deterministic function
$
g : \mathcal{X} \rightarrow \mathcal{S} \cup \{\emptyset\},
$
where $\mathcal{S}$ denotes the space of diagnosis strings. For each document $x_i$, the output is
$
s_i = g(x_i),
$
where $s_i \in \mathcal{S}$ if a diagnosis section is detected and $\emptyset$ otherwise.
Three main challenges arise in handling diagnosis sections:
\begin{enumerate}
    \item Not all discharge summaries include this indication, as it may be absent or left empty.
    \item Different templates use varying terminologies and position this indication differently in the text.
    \item Diagnosis strings are typically written as free text, with the same disease described by different strings and levels of detail
\end{enumerate} 
The first point is the primary reason why our pipeline focuses on classifying the text of a discharge letter rather than relying solely on the extracted diagnosis strings.
To address the second point, the function $g$ is implemented using a predefined set of regular expressions that identify and extract the diagnosis section across heterogeneous templates. Regular expressions provide a flexible and reproducible pattern-matching mechanism for detecting structured fields with limited syntactic variation. The full set of patterns is reported in \ref{app:diagnosis}.

After this step, for the majority of the letters we obtain a free-text diagnosis string describing the conditions associated with the current hospitalization. This extraction step focuses on diagnoses explicitly reported in the structured diagnosis section and does not aim to capture conditions mentioned exclusively in other narrative sections (e.g., past medical history). These strings are not directly used as labels due to their heterogeneity (point 3 above); instead, they are clustered, as detailed in the following section.

\subsubsection*{Diagnosis Clustering}
\label{subsec:clustering-pipeline}
Once the diagnosis strings are extracted, we cluster them to identify semantically coherent groups corresponding to diseases or symptoms . This corresponds to Step 1.2 in panel a of Figure \ref{fig:global-pipeline} and is expanded in panel b of the same figure.

Let ${I} = \{ i \in \{1,\dots,N\} : s_i \neq \emptyset \}$ denote the subset of documents for which a diagnosis string was successfully extracted. In this clustering step, we consider only the set $\{ s_i \}_{i \in {I}}$, ignoring documents with $s_i = \emptyset$.

Each diagnosis string undergoes minor normalization, including removal of punctuation, dates, and time expressions, through a deterministic pre-processing function $p_s : \mathcal{S} \rightarrow \mathcal{S}$. The normalized string is denoted $\bar{s}_i = p_s(s_i)$ (Step 1.2.1).

Each normalized diagnosis string is embedded using the domain-adapted transformer encoder $f_{\hat{\theta}}$ defined in Section~\hyperref[sec:bert-pretraining]{Pre-training of a Transformer-based model for Italian medical documents (TPT)}. The resulting embedding is
\begin{equation}
z_i = f_{\hat{\theta}}(\bar{s}_i)~\in~\mathbb{R}^{d},~\quad i~\in~{I}.
\end{equation}
Let $Z = \{ z_i \}_{i \in {I}}$ denote the set of all diagnosis embeddings (Step 1.2.2).

Since high-dimensional embeddings are not directly suitable for density-based clustering, we apply Principal Component Analysis (PCA) to $Z$ (Step~1.2.3). Let $W \in \mathbb{R}^{d \times k}$ denote the projection matrix obtained by fitting PCA on $Z$, where in our case studies $d=768, k=100$, retaining more than 80\% of the variance in the embedding space. The projected embeddings are
$
\tilde{z}_i = W^\top z_i \in \mathbb{R}^{k}.
$

After this step, a clustering algorithm can be applied to the resulting embedding vectors (Step~1.2.4). We selected HDBSCAN~\cite{campello2013density} because it does not require specifying the number of clusters a priori, allows leaving part of data unclustered, and can identify clusters with heterogeneous shapes and densities. These properties are well suited to transformer-based text embeddings, which often lie on complex, nonlinear manifolds and may form clusters with varying densities, irregular geometries, and overlapping boundaries. Several commonly used clustering algorithms, such as K-Means, Gaussian Mixture Models, and standard hierarchical clustering, are less suitable for this type of data~\cite{asyaky2021improving, yeasmin2022transformer}, as further supported by a comparative analysis reported in \ref{app:clustering}. 

We apply HDBSCAN to the set $\{ \tilde{z}_i \}_{i \in {I}}$. The clustering defines an assignment function
$
h : \mathbb{R}^{k} \rightarrow \{1,\dots,K\} \cup \{-1\},
$
where $h(\tilde{z}_i) = k$ indicates membership in cluster $C_k$, and $h(\tilde{z}_i) = -1$ denotes noise points left unclustered. HDBSCAN is parameterized by a minimum cluster size and a minimum number of samples, selected via grid search as detailed in \ref{app:clustering}.

The first-level clustering may produce multiple compact clusters corresponding to the same disease category. To merge semantically related clusters, we construct a keyword-based representation for each cluster.

For each cluster $C_k$, we define a set of representative keywords $\mathrm{Keywords}(C_k)$ obtained by applying frequency-based filtering to the tokens appearing in $C_k$ after tokenization and stopword removal. These keywords provide a compact textual summary of the cluster. The complete keyword extraction procedure is described in \ref{app:clustering}.

Each cluster is then represented by embedding its keyword summary using the same encoder $f_{\hat{\theta}}$, yielding a cluster-level embedding $u_k \in \mathbb{R}^{d}$. A second-level clustering procedure is applied to $\{u_k\}$ to merge clusters that are semantically close~(Step~1.2.6).

\subsubsection*{Mapping clusters to diseases}
Step 1.3 consists of mapping clusters to weak labels that will be used to train the classification model.

Let $\{C_k\}_{k=1}^K$ denote the clusters obtained in Step~1.2 and let $\mathrm{Keywords}(C_k)$ be the keyword set associated with cluster $C_k$.

For each disease of interest, we specify one or more definitions. Each definition $d$ consists of:
\begin{itemize}
    \item a set of positive keywords $P_d = \{p_1, \dots, p_r\}$ that must be present,
    \item optionally, a set of negative keywords $N_d = \{n_1, \dots, n_s\}$ that must be absent.
\end{itemize}

A cluster $C_k$ is selected as positive for the disease if there exists at least one definition $d$ such that
$
P_d \subseteq \mathrm{Keywords}(C_k)
\quad \text{and} \quad
N_d \cap \mathrm{Keywords}(C_k) = \emptyset.
$

Let 
$
L = \{ k \in \{1,\dots,K\} : C_k \text{ satisfies at least one } d \}
$ denote the set of clusters selected for the disease, $h(i)$ the cluster assignment of diagnosis string $s_i$ (Step~1.2), and 
$
I_c = \{ i \in I : h(i) \neq -1 \}
$ the subset of documents whose diagnosis strings were assigned to a cluster (i.e., excluding noise points identified by HDBSCAN). The weak label is defined as
\begin{equation}
\tilde{y}_i =
\begin{cases}
1 & \text{if } h(i) \in L, \\
0 & \text{if } h(i) \notin L,
\end{cases}
\quad i \in I_c .
\end{equation}
Documents with $h(i) = -1$ (i.e., diagnosis strings left unclustered) or $i \notin I$ (no diagnosis string extracted) do not receive a weak label and are excluded from the weakly labelled training subset.

Thus, among the documents in $I_c$, discharge letters whose extracted diagnosis strings belong to selected clusters are assigned a positive weak label, while the remaining clustered documents receive a negative weak label.

This procedure is fully deterministic once the disease definitions are specified and does not depend on gold annotations. The only domain input required is the specification of disease-related keywords, which is substantially less demanding than document-level manual annotation.
In practice, keyword definitions are specified in collaboration with clinicians, with attention to the terminology typically used in discharge letters for the disease of interest. This step does not require document-level review and is performed independently of gold labels. While the coverage of weak labels depends on the completeness of the keyword definitions, the cluster-level matching strategy reduces sensitivity to minor lexical variations. 

A keyword-based selection at the cluster level is more robust than applying keywords directly to the raw diagnosis strings extracted in Step~1.1. Individual diagnosis strings are often heterogeneous, contain variable phrasing, and may include negations or references to past medical history. Direct keyword matching at the string level would therefore introduce a higher rate of false positives and false negatives. By contrast, clustering first aggregates semantically similar diagnoses, allowing keyword matching to operate on stabilized and semantically coherent cluster representations.
Similarly, manually reviewing all extracted diagnosis strings would be impractical due to their large number.

\subsection*{Classification of discharge letters using the weakly labelled dataset (WLC)}
\label{subsec:classification-pipeline}
The classification pipeline that exploits the weakly labelled data produced by Step 1 corresponds to Step 2 (WLC) in Figure~\ref{fig:global-pipeline}.a. 
Let $\tilde{\mathcal{D}} = \{(x_i, \tilde{y}_i) : i \in {I_c}\}$ denote the weakly labelled dataset constructed in Step 1.3. Only documents with an extracted diagnosis string are included in this training set.

Each document undergoes a deterministic cleaning function $p : \mathcal{X} \rightarrow \mathcal{X}$ that removes headers and footers containing non-informative administrative content (see \ref{app:preproc} for details). The processed document is denoted $\bar{x}_i = p(x_i)$ (Step 2.1).

Since RoBERTa-based models are limited to 512 input tokens, we define a truncation operator $\tau_{512}(\cdot)$ that retains the first 512 tokens of the processed document. The final input to the classifier is therefore
$
x_i^{*} = \tau_{512}(\bar{x}_i).
$
In our dataset, approximately 90\% of discharge letters contain fewer than 512 tokens after preprocessing, so truncation rarely affects the input. For datasets with a larger proportion of long documents, alternative strategies could be considered, such as splitting the document into segments of at most 512 tokens and aggregating segment-level probabilities, or using architectures designed for longer contexts, as discussed in studies on long-document classification with BERT-based models~\cite{limsopatham2021effectively}.

The classification model is based on the domain-adapted transformer encoder $f_{\hat{\theta}}$ defined in Section~\hyperref[sec:bert-pretraining]{Pre-training of a Transformer-based model for Italian medical documents (TPT)}. A linear classification layer with parameters $w \in \mathbb{R}^{d}$ and bias $b \in \mathbb{R}$ is added on top of the encoder.

For an input document $x_i^{*}$, the predicted probability of the disease is
$
\hat{p}_i = \sigma\!\left(w^\top f_{\hat{\theta}}(x_i^{*}) + b \right),
$
where $\sigma(z) = 1/(1+e^{-z})$ is the logistic function.

Model parameters $\phi = (\hat{\theta}, w, b)$ are optimized by minimizing the binary cross-entropy loss over the weakly labelled dataset:
\begin{equation}
\mathcal{L}_{\text{WLC}}(\phi)
=
- \sum_{i \in {I_c}}
\left[
\tilde{y}_i \log \hat{p}_i
+
(1-\tilde{y}_i) \log (1-\hat{p}_i)
\right].
\end{equation}

\subsection*{Experimental outline}
\label{subsec:eval}
The study uses anonymized discharge letters obtained from the Pedianet network~\cite{cantarutti2021pedianet}. Inclusion in the Pedianet database is voluntary and parents/legal guardians must provide informed consent for their children's anonymized data to be used for research purposes. The study was performed in accordance with the Declaration of Helsinki. Ethical approval of the study and access to the database were granted by the Internal Scientific Committee of So.Se.Te. Srl, the legal owner of Pedianet.

The main evaluation is conducted at the final classification stage of the pipeline, where predictions are compared against manually annotated gold labels $y_i^{\text{gold}}$. In addition, intermediate components of the pipeline—diagnosis extraction (Step~1.1) and clustering (Steps~1.2–1.3)—are evaluated on a randomly selected subset of 1{,}000 diagnosis strings for each dataset, manually annotated by domain experts. Each string was labelled as \textit{correct}, \textit{partially correct}, or \textit{wrong} with respect to (i) completeness of the extracted diagnosis string and (ii) correctness of its assignment to first- and second-level clusters. 

At the classification stage, we compare our pipeline with:
\begin{itemize}
    \item Two unsupervised rule-based baselines:
    \begin{itemize}
        \item \textit{Rule-based full-text}: keyword matching applied to the entire discharge letter;
        \item \textit{Rule-based diagnosis}: keyword matching applied only to extracted diagnosis strings.
    \end{itemize}
    The keywords correspond to the disease definitions introduced in Step~1.3.
    
    \item An \textit{XPRESS-style weak supervision baseline} (\textit{WS-XPRESS-UmBERTo-TPT}). Following the XPRESS paradigm~\cite{agarwal2016learning}, we generate weak labels directly applying the keyword definitions defined in Step~1.3 (positive and negative sets) to the discharge letters and train the UmBERTo-TPT model used in our pipeline on these weak labels.
    \item A zero-shot large language model (LLM) for Italian (\textit{LLaMAntino-3-ANITA-8B-Inst-DPO-ITA})~\cite{polignano2024advanced}, prompted to classify each discharge letter as positive or negative for bronchiolitis.
\end{itemize}

To assess the contribution of the transformer backbone and domain-adaptive pre-training (TPT), we replace the classifier in Step~2.2 with:
\begin{enumerate}
    \item UmBERTo without TPT (\textit{WS-UmBERTo})~\cite{tamburini2020bertology};
    \item XLM-RoBERTa-base~\cite{conneau2020unsupervised}, with (\textit{WS-XLM-RoBERTa-TPT}) and without (\textit{WS-XLM-RoBERTa}) TPT;
    \item RemBERT~\cite{chung2020rethinking}, with (\textit{WS-RemBERT-TPT}) and without (\textit{WS-RemBERT-BPT}) TPT;
    \item An LSTM-based recurrent neural network using Italian Word2Vec embeddings (\textit{WS-LSTM})~\cite{mikolov2013efficient}.
\end{enumerate}

The three transformer-based models with TPT are also evaluated in a fully supervised setting (\textit{FS-UmBERTo-TPT}, \textit{FS-XLM-RoBERTa-TPT}, \textit{FS-RemBERT-TPT}) trained directly on gold labels, providing an upper bound on performance under full supervision.

Since weak labels are derived from clustered diagnosis strings, classification models may over-rely on diagnosis sections. To assess this potential bias, we evaluate models under two conditions:
\begin{enumerate}
    \item Using the full discharge letter;
    \item Removing the diagnosis strings used to construct weak labels (when present).
\end{enumerate}

To better understand the effect of weak supervision, we perform additional analyses. First, we assess the robustness of the results with respect to the construction of weak labels by removing individual clusters from the set of positive weak labels and re-evaluating the models. Second, we analyze the impact of label noise by replacing varying proportions of gold labels with weak labels and measuring the resulting performance degradation. Finally, we evaluate classification performance against the weak labels themselves, allowing us to quantify the discrepancy between weak and gold supervision (in \ref{app:classification}).

Classification performance is evaluated using precision (P), recall (R), F1-score (F1) for the positive class, area under the ROC curve (AUROC), and area under the precision–recall curve (AUPRC). For classifiers producing a continuous score, the classification threshold is selected within training folds using Youden’s index~\cite{youden1950index}, and all evaluation metrics are computed on held-out test folds reflecting the original class prevalence.

All metrics are computed using stratified cross-validation. For the bronchiolitis dataset, we use 10-fold cross-validation. For the smaller bronchitis validation dataset, which contains 97 positive cases, we use 5-fold cross-validation to ensure a sufficient number of positive examples in each test fold. For each metric, we report the mean and standard deviation across folds.

To assess statistical significance of performance differences, we perform paired permutation (sign-flip) tests on fold-wise differences between our method and each other model. All tests are two-sided. To control for multiple comparisons across model variants, p-values are adjusted using the Holm–Bonferroni procedure within each metric. Differences are considered statistically significant when the Holm-adjusted p-value is below 0.05.

AUROC and AUPRC are not reported for rule-based classifiers, as they do not produce probabilistic outputs. Fully supervised models are not evaluated on weak labels.

Since data originate from multiple hospitals and Local Health Units (LHUs), additional experiments assessing inter-site variability are reported in \ref{app:classification}.

\subsection*{Training details}
Additional pre-training of the transformer-based models on the E3C corpus was performed using the masked language modeling objective described in Section~\hyperref[sec:bert-pretraining]{Pre-training of a Transformer-based model for Italian medical documents (TPT)}. The models were optimized using a learning rate of $2 \cdot 10^{-5}$ and weight decay of $0.01$. A validation set consisting of 20\% of the corpus was used for early stopping.

The LSTM baseline consists of a frozen Word2Vec embedding layer followed by a 32-unit LSTM layer, a dropout layer, a 16-unit fully connected layer, an additional dropout layer, and the final output layer. The model is trained using the Adam optimizer with a learning rate scheduler starting from $10^{-2}$, decay rate $0.7$, and decay step size of 100. Early stopping is applied based on validation performance. 

The transformer-based classifiers are trained using the AdamW optimizer with a linear learning rate scheduler starting from $10^{-3}$ and weight decay of $0.01$. These hyperparameters were selected on training folds of the bronchiolitis dataset using grid search. The same configuration was applied unchanged to the bronchitis dataset. Training proceeds for a number of epochs determined by early stopping (6 epochs for UmBERTo models, 4 for XLM-RoBERTa models, and 3 for RemBERT models).

To stabilize optimization, a gradual unfreezing strategy is applied. Initially, only the final classification layer is trained. After two epochs, the last transformer layer is unfrozen, and after four epochs (if reached), the last but one transformer layer is also unfrozen.

To mitigate class imbalance during training, negative examples are undersampled within each training fold, limiting their number to at most twice that of the positive class. No undersampling is applied to test folds. Additional results supporting this choice are reported in \ref{app:classification}.

All experiments were implemented in Python. Training was performed on a single NVIDIA Tesla V100 GPU with 32\,GB of VRAM.

\section*{Results}
\label{sec:results}
\subsection*{Datasets}
\label{subsec:data}
The dataset for the main case study comprises 33,176 discharge letters related to 15,196 children who were admitted to emergency rooms and/or hospitalized in the Veneto Region between January 2017 and December 2020, with a family pediatrician adhering to the Pedianet network~\cite{cantarutti2021pedianet}. 

The dataset covers 44 different hospitals in the region, which can be grouped into 9 Local Health Units (LHUs) and 2 university hospitals. 
Each discharge letter includes the complete text of the letter, along with the date of the letter and an anonymized identifier of the patient. A manual review of discharge letters was conducted to extract relevant diagnoses of bronchiolitis, resulting in 335 positive cases within this dataset. This review was performed by a data manager with a clinical background, with the support of clinicians for uncertain cases. 
These labels are the gold standard to validate our pipeline, and we do not make direct use of them in the pipeline training (except for the fully supervised experiments). 
The median length of these letters is 4,515 characters (1st-3rd quartiles 3,240-5,806).

As an additional validation setting, we analyze a smaller dataset of 3,188 discharge letters related to 2,364 children who were admitted to emergency rooms and/or hospitalized in the Veneto Region between January 2020 and May 2025, with a family pediatrician adhering to the Pedianet network. This dataset covers 31 hospitals and 11 LHUs. Manual review, following the same approach of the bronchiolitis dataset, identified 97 cases of bronchitis, which serve as gold labels for evaluation. The median length of these letters is 5,387 characters (1st-3rd quartiles 2,601-6,759).

\ref{app:data-plots} reports additional plots with data distributions among hospitals and LHUs for the two datasets. An example of a fictitious letter is reported in \ref{app:letter-example}. 

A secondary annotator independently reviewed a stratified sample of 30 positive and 30 negative discharge letters for each dataset, resulting in perfect inter-annotator agreement for bronchiolitis ($\kappa=1.00$) and substantial agreement for bronchitis ($\kappa=0.73$), with discrepancies primarily involving less certain cases.

\subsection*{Diagnosis extraction}
\label{sub:diagnosis-results}
Our diagnosis extraction module (Step 1.1) successfully extracted a diagnosis sentence from 89\% of the discharge letters in the bronchiolitis dataset. We extracted 15,038 unique diagnosis sentences, of which 11,279 occurred only once, confirming the need for clustering them in wider groups that can be more easily related to known diseases and symptoms.
ICD-9-CM codes are present only in 737 diagnosis strings (4.9\%), covering 3,436 discharge letters (10.4\%). 
The most common strings are the simplest, while more articulate ones have single occurrences. The median length of the extracted diagnosis strings is 49 characters, with first and third quartiles of 17 and 156 characters.
Among the 335 bronchiolitis-labelled letters, we extracted 138 unique diagnosis strings from 306 letters.

The same extraction procedure was applied to the bronchitis validation dataset. Diagnosis strings were successfully extracted from 90\% of the discharge letters, yielding 1,376 unique diagnosis strings, of which 1,092 occurred only once. Considering the 97 bronchitis-labelled letters, diagnosis strings were extracted from all of them, producing 88 unique diagnosis expressions. 

\begin{figure}[hp]
    \centering
    \includegraphics[width=.8\linewidth]{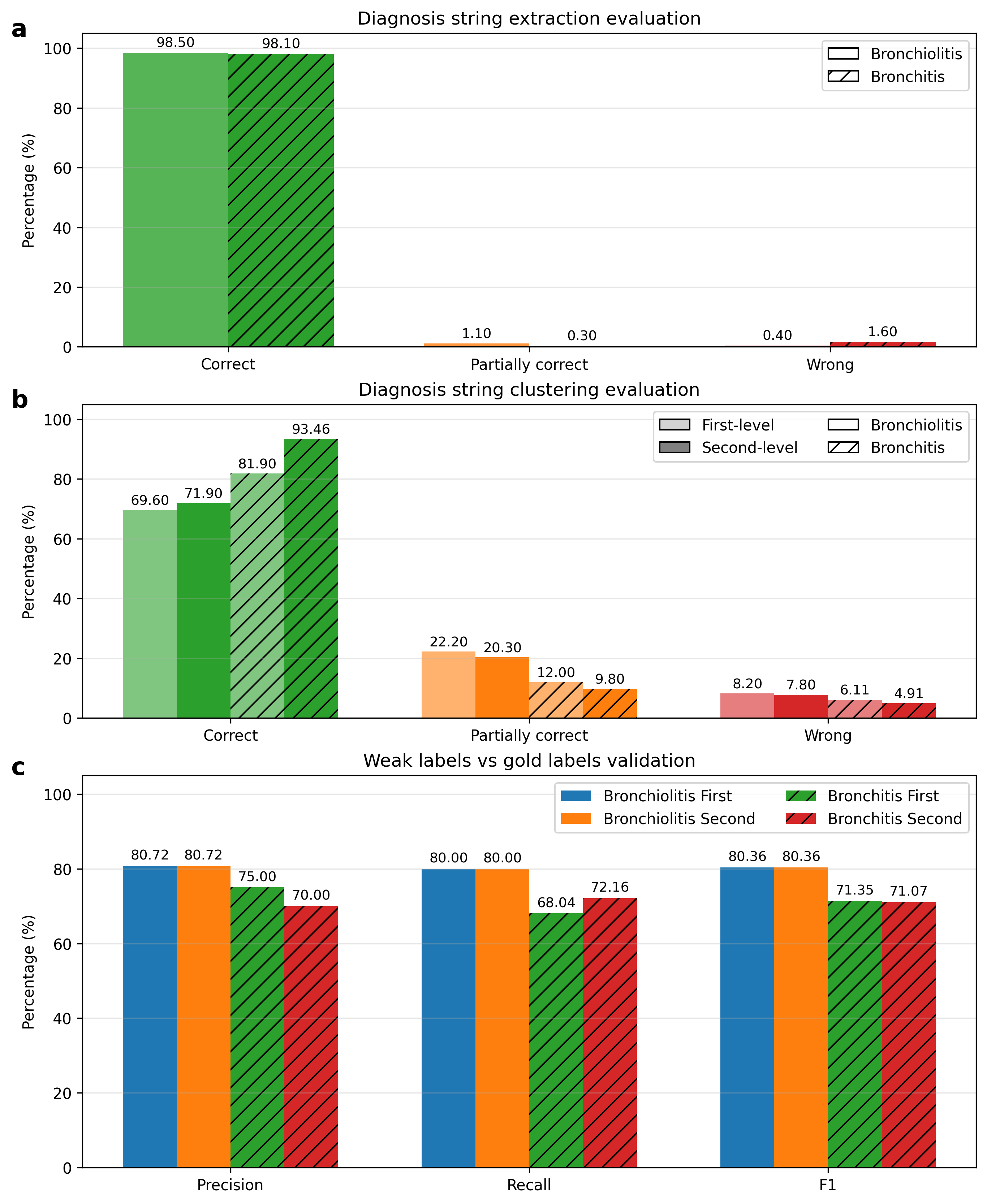}
    \caption{Evaluation of the intermediate steps of the pipeline on the bronchiolitis and bronchitis datasets. a. Evaluation of diagnosis string extraction on 1,000 manually annotated diagnosis strings per dataset, reporting the percentage of extracted strings classified as correct, partially correct, or wrong. b. Evaluation of diagnosis clustering after first-level and second-level clustering on the same set of annotated diagnosis strings. c. Validation of weak labels against gold labels in terms of precision, recall, and F1-score, reported separately for labels obtained from first-level and second-level clustering.}
    \label{fig:labels-eval}
\end{figure}

The manual evaluation results of this step on a random sample of 1,000 strings for each dataset are summarized in Figure~\ref{fig:labels-eval}.a. The number of extracted strings that are not diagnosis strings or that are incomplete is very limited ($\leq 2\%$).

\subsection*{Diagnosis clustering}
\label{sub:clustering-results}
In the bronchiolitis dataset, the 15,038 extracted diagnosis strings were grouped into 590 clusters. Cluster sizes varied significantly, from a large cluster of 270 strings to several smaller clusters containing 5 strings. The subsequent second-level clustering reduced the number of clusters to 135. In the bronchitis dataset clusters were 77 and 37, respectively. A summary of the main clusters is reported in \ref{app:clustering}.

The quality of the clustering assignments was evaluated on a random sample of 1,000 diagnosis strings per dataset. Figure~\ref{fig:labels-eval}.b reports the proportion of assignments that were manually classified as correct, partially correct, or wrong at both clustering levels.
Most of the \textit{partially correct} cluster assignments are due to minor variations in disease specification, e.g., \textit{fracture of right hand} assigned to a cluster for \textit{fracture of left hand}, \textit{severe influenza} assigned to a cluster for \textit{mild influenza}.  A smaller number of cases correspond to diagnoses mentioning multiple symptoms or diseases that only partially match the cluster definition, such as \textit{vomiting and fever} assigned to a cluster describing \textit{fever} alone.

\begin{table}[t]
\centering
\caption{Keyword-based representation of the first 15 first-level clusters related to respiratory infections in each dataset, with the \% of cases in each cluster, the \% to the total number of cases in the dataset and the indication of the clusters that would be selected as weak labels at this stage.}
\label{tab:clusters1_resp}
\resizebox{\textwidth}{!}{%
\begin{tabular}{|lllrr|c|}
\hline
\multicolumn{1}{|l|}{\textbf{Cluster \#}} &
  \multicolumn{1}{l|}{\textbf{Cluster keywords {[}ITA{]}}} &
  \multicolumn{1}{l|}{\textbf{Cluster keywords {[}ENG{]}}} &
  \multicolumn{1}{r|}{\textbf{\begin{tabular}[c]{@{}r@{}}Within cluster \\ cases \%\end{tabular}}} &
  \textbf{\begin{tabular}[c]{@{}r@{}}\% of total cases \\ covered\end{tabular}} &
  \textbf{\begin{tabular}[c]{@{}c@{}}Positive\\ weak label\end{tabular}} \\ \hline
\multicolumn{5}{|c|}{\textbf{Bronchiolitis}} &
   \\ \hline
\multicolumn{1}{|l|}{1} &
  \multicolumn{1}{l|}{bronchiolite lieve} &
  \multicolumn{1}{l|}{mild bronchiolitis} &
  \multicolumn{1}{r|}{85.19} &
  6.87 &
  y \\ \hline
\multicolumn{1}{|l|}{2} &
  \multicolumn{1}{l|}{acuta bronchiolite iniziale lieve} &
  \multicolumn{1}{l|}{acute mild initial bronchiolitis} &
  \multicolumn{1}{r|}{85.14} &
  44.48 &
  y \\ \hline
\multicolumn{1}{|l|}{3} &
  \multicolumn{1}{l|}{acuta bronchiolite corso insufficienza respiratoria} &
  \multicolumn{1}{l|}{acute bronchiolitis undergoing respiratory failure} &
  \multicolumn{1}{r|}{74.05} &
  11.94 &
  y \\ \hline
\multicolumn{1}{|l|}{4} &
  \multicolumn{1}{l|}{acuto broncospasmo corso febbre paziente} &
  \multicolumn{1}{l|}{acute bronchospasm ongoing fever patient} &
  \multicolumn{1}{r|}{70.00} &
  15.02 &
  y \\ \hline
\multicolumn{1}{|l|}{5} &
  \multicolumn{1}{l|}{bronchiolite difficolta' alimentazione} &
  \multicolumn{1}{l|}{bronchiolitis feeding difficulties} &
  \multicolumn{1}{r|}{50.00} &
  2.09 &
  y \\ \hline
\multicolumn{1}{|l|}{6} &
  \multicolumn{1}{l|}{acuta infezione insufficienza respiratoria transitoria virosi} &
  \multicolumn{1}{l|}{acute respiratory infection transient virosis} &
  \multicolumn{1}{r|}{12.70} &
  2.39 &
  n \\ \hline
\multicolumn{1}{|l|}{7} &
  \multicolumn{1}{l|}{broncospasmo infezione respiratoria} &
  \multicolumn{1}{l|}{bronchospasm respiratory infection} &
  \multicolumn{1}{r|}{6.67} &
  0.30 &
  n \\ \hline
\multicolumn{1}{|l|}{8} &
  \multicolumn{1}{l|}{broncospasmo disidratazione lieve} &
  \multicolumn{1}{l|}{bronchospasm mild dehydration} &
  \multicolumn{1}{r|}{1.17} &
  0.60 &
  n \\ \hline
\multicolumn{1}{|l|}{9} &
  \multicolumn{1}{l|}{flogosi alte vie respiratorie febbre} &
  \multicolumn{1}{l|}{high airways phlogosis fever} &
  \multicolumn{1}{r|}{0.00} &
  0.00 &
  n \\ \hline
\multicolumn{1}{|l|}{10} &
  \multicolumn{1}{l|}{broncospasmo infezione vie respiratorie} &
  \multicolumn{1}{l|}{bronchospasm airways infection} &
  \multicolumn{1}{r|}{0.00} &
  0.00 &
  n \\ \hline
\multicolumn{1}{|l|}{11} &
  \multicolumn{1}{l|}{acuto broncospasmo} &
  \multicolumn{1}{l|}{acute bronchospasm} &
  \multicolumn{1}{r|}{0.00} &
  0.00 &
  n \\ \hline
\multicolumn{1}{|l|}{12} &
  \multicolumn{1}{l|}{broncospasmo flogosi vie} &
  \multicolumn{1}{l|}{bronchospasm airways phlogosis} &
  \multicolumn{1}{r|}{0.00} &
  0.00 &
  n \\ \hline
\multicolumn{1}{|l|}{13} &
  \multicolumn{1}{l|}{broncospasmo polmonite} &
  \multicolumn{1}{l|}{bronchospasm pneumonia} &
  \multicolumn{1}{r|}{0.00} &
  0.00 &
  n \\ \hline
\multicolumn{1}{|l|}{14} &
  \multicolumn{1}{l|}{broncospasmo corso} &
  \multicolumn{1}{l|}{ongoing bronchospasm} &
  \multicolumn{1}{r|}{0.00} &
  0.00 &
  n \\ \hline
\multicolumn{1}{|l|}{15} &
  \multicolumn{1}{l|}{broncopolmonite sinistra} &
  \multicolumn{1}{l|}{left bronchopneumonia} &
  \multicolumn{1}{r|}{0.00} &
  0.00 &
  n \\ \hline
\multicolumn{5}{|c|}{\textbf{Bronchitis}} &
   \\ \hline
\multicolumn{1}{|l|}{1} &
  \multicolumn{1}{l|}{bronchite acuta} &
  \multicolumn{1}{l|}{acute bronchitis} &
  \multicolumn{1}{r|}{90.48} &
  39.18 &
  y \\ \hline
\multicolumn{1}{|l|}{2} &
  \multicolumn{1}{l|}{broncospasmo acuto parainfettivo} &
  \multicolumn{1}{l|}{acute parainfectious bronchospasm} &
  \multicolumn{1}{r|}{22.73} &
  5.15 &
  n \\ \hline
\multicolumn{1}{|l|}{3} &
  \multicolumn{1}{l|}{broncospasmo acuto} &
  \multicolumn{1}{l|}{acute bronchospasm} &
  \multicolumn{1}{r|}{16.67} &
  5.15 &
  n \\ \hline
\multicolumn{1}{|l|}{4} &
  \multicolumn{1}{l|}{bronchite} &
  \multicolumn{1}{l|}{bronchitis} &
  \multicolumn{1}{r|}{84.62} &
  22.68 &
  y \\ \hline
\multicolumn{1}{|l|}{5} &
  \multicolumn{1}{l|}{nausea vomito febbre} &
  \multicolumn{1}{l|}{nausea vomit fever} &
  \multicolumn{1}{r|}{5.00} &
  1.03 &
  n \\ \hline
\multicolumn{1}{|l|}{6} &
  \multicolumn{1}{l|}{insufficienza respiratoria infezione basse} &
  \multicolumn{1}{l|}{respiratory insufficiency lower infection} &
  \multicolumn{1}{r|}{30.00} &
  6.19 &
  y \\ \hline
\multicolumn{1}{|l|}{7} &
  \multicolumn{1}{l|}{rinite polmonite basale destra} &
  \multicolumn{1}{l|}{rhinitis right basal pneumonia} &
  \multicolumn{1}{r|}{10.00} &
  3.09 &
  n \\ \hline
\multicolumn{1}{|l|}{8} &
  \multicolumn{1}{l|}{broncospasmo} &
  \multicolumn{1}{l|}{bronchospasm} &
  \multicolumn{1}{r|}{9.09} &
  2.06 &
  n \\ \hline
\multicolumn{1}{|l|}{9} &
  \multicolumn{1}{l|}{polmonite sinistra insufficienza respiratoria} &
  \multicolumn{1}{l|}{left pneumonia respiratory insufficiency} &
  \multicolumn{1}{r|}{9.09} &
  1.03 &
  n \\ \hline
\multicolumn{1}{|l|}{10} &
  \multicolumn{1}{l|}{otite media acuta} &
  \multicolumn{1}{l|}{acute otitis media} &
  \multicolumn{1}{r|}{0.00} &
  0.00 &
  n \\ \hline
\multicolumn{1}{|l|}{11} &
  \multicolumn{1}{l|}{infezione basse alte vie respiratorie} &
  \multicolumn{1}{l|}{respiratory infection lower upper} &
  \multicolumn{1}{r|}{33.33} &
  4.12 &
  n \\ \hline
\multicolumn{1}{|l|}{12} &
  \multicolumn{1}{l|}{laringite ipoglottica} &
  \multicolumn{1}{l|}{hypoglottic laryngitis} &
  \multicolumn{1}{r|}{0.00} &
  0.00 &
  n \\ \hline
\multicolumn{1}{|l|}{13} &
  \multicolumn{1}{l|}{bronchiolite lieve} &
  \multicolumn{1}{l|}{mild bronchiolitis} &
  \multicolumn{1}{r|}{4.76} &
  1.03 &
  n \\ \hline
\multicolumn{1}{|l|}{14} &
  \multicolumn{1}{l|}{infezione} &
  \multicolumn{1}{l|}{infection} &
  \multicolumn{1}{r|}{3.57} &
  1.03 &
  n \\ \hline
\multicolumn{1}{|l|}{15} &
  \multicolumn{1}{l|}{febbre flogosi alte vie} &
  \multicolumn{1}{l|}{fever upper tract phlogosis} &
  \multicolumn{1}{r|}{3.23} &
  2.06 &
  n \\ \hline
\end{tabular}
}
\end{table}

Table~\ref{tab:clusters1_resp} reports the main first-level clusters related to respiratory infections for both datasets, including the number of bronchiolitis or bronchitis cases covered by each cluster. Given space constraints, we report here only the 15 largest clusters mentioning bronchiolitis, bronchitis, bronchospasm, pneumonia, respiratory infections, otitis, or fever, since these are the clusters most closely related to the respiratory diseases considered in our case studies.
For the bronchiolitis dataset, four clusters (\#1, \#2, \#3, \#5) contain the keyword \textit{``bronchiolitis"}. Three of them include a large majority of bronchiolitis cases, whereas the fourth cluster (\#5) has only half of its diagnoses labelled as bronchiolitis. These four clusters together cover 65.38\% of all bronchiolitis cases.
Additionally, 13 clusters include the keyword \textit{``bronchospasm"}, a symptom frequently associated with bronchiolitis. According to physicians who annotated the data, bronchospasm is associated with bronchiolitis only if accompanied by fever. Only one of these 13 clusters~(\#4) has both \textit{``bronchospasm"} and \textit{``fever"} keywords, and indeed, it contains a high proportion (70\%) of bronchiolitis cases, accounting for an additional 15\% of the total bronchiolitis cases. Thus, these five clusters are those that, without using the gold labels, would be selected to define the weak labels for the training set, according to the keyword-based definitions \textit{\{positive=\{bronchiolitis\},negative=\{\}\}} and \textit{\{positive=\{bronchospasm, fever\},negative=\{\}\}}. 
Among the respiratory infection clusters, three additional clusters contain bronchiolitis cases (\#6, \#7, \#8), although in a small number.
For the bronchitis dataset, three clusters are selected at the first level (\#1, \#4, \#6), according to the definitions \textit{\{positive=\{bronchitis\},negative=\{\}\}} and \textit{\{positive=\{lower, tract, respiratory, infection\},negative=\{upper\}\}}. These cover 68.05\% of bronchitis cases.

Table~\ref{tab:clusters2} reports the second-level clusters obtained from first-level clusters reported in Table~\ref{tab:clusters1_resp}. In particular, several clusters related to \textit{bronchiolitis}, \textit{bronchospasm}, \textit{bronchitis}, and \textit{respiratory infections} are aggregated into broader second-level clusters.

In the bronchiolitis case study the selection of weak labels for training remains unchanged, as second-level clusters \#1, \#2, \#3, and \#4 cover exactly the clusters that would have been selected at the first level. Consequently, a downstream comparison between first- and second-level clustering would yield identical weak labels and identical classification results in this specific setting. In the bronchitis dataset, the second-level clustering merges two clusters that are partially related to \textit{lower tract respiratory infections}, adding them to the set of positive weak labels. 

\newcolumntype{Y}{>{\raggedright\arraybackslash}X} 
\newcolumntype{R}{>{\raggedleft\arraybackslash}p{1.2cm}}
\newcolumntype{C}{>{\centering\arraybackslash}p{1.1cm}}

\begin{table}[!htbp]
\centering
\tiny

\setlength{\tabcolsep}{5pt}      
\renewcommand{\arraystretch}{0.85}
\setlength{\extrarowheight}{1pt}

\caption{Keyword-based representation of the second-level clusters that are obtained as merge of first level-clusters reported in Table~1, with the \% of cases in each cluster, the \% to the total number of cases in the dataset and the indication of those selected as positive weak labels.}
\label{tab:clusters2}

\resizebox{\textwidth}{!}{%
\begin{tabularx}{\linewidth}{|R|Y|Y|R|Y|Y|R|R|C|}
\hline
\makecell[tl]{\textbf{1st lvl}\\\textbf{cluster}} &
\makecell[tl]{\textbf{2nd level}\\\textbf{cluster}\\\textbf{keywords}\\\textbf{[ITA]}} &
\makecell[tl]{\textbf{2nd level}\\\textbf{cluster}\\\textbf{keywords}\\\textbf{[ENG]}} &
\makecell[tl]{\textbf{1st lvl}\\\textbf{cluster}} &
\makecell[tl]{\textbf{1st level}\\\textbf{cluster}\\\textbf{keywords}\\\textbf{[ITA]}} &
\makecell[tl]{\textbf{1st level}\\\textbf{cluster}\\\textbf{keywords}\\\textbf{[ENG]}} &
\makecell[tl]{\textbf{Within}\\\textbf{cluster}\\\textbf{bronch. \%}} &
\makecell[tl]{\textbf{\% of total}\\\textbf{bronch.}\\\textbf{covered}} &
\makecell[tc]{\textbf{Positive}\\\textbf{weak}\\\textbf{label}} \\
\hline

\multicolumn{9}{|c|}{\textbf{Bronchiolitis}} \\ \hline

\multirow[t]{2}{*}{1} &
\multirow[t]{2}{=}{bronchiolite lieve} &
\multirow[t]{2}{=}{mild bronchiolitis} &
1 & acuta bronchiolite iniziale lieve & acute mild initial bronchiolitis &
\multirow[t]{2}{*}{85.16} & \multirow[t]{2}{*}{51.34} & \multirow[t]{2}{*}{y} \\
\cline{4-6}
& & & 2 & bronchiolite lieve & mild bronchiolitis & & & \\
\hline

\multirow[t]{2}{*}{2} &
\multirow[t]{2}{=}{broncospasmo infezione respiratoria} &
\multirow[t]{2}{=}{bronchospasm respiratory infection} &
7 & broncospasmo infezione vie respiratorie & bronchospasm infection airways &
\multirow[t]{2}{*}{2.90} & \multirow[t]{2}{*}{0.30} & \multirow[t]{2}{*}{n} \\
\cline{4-6}
& & & 10 & broncospasmo infezione respiratoria & bronchospasm respiratory infection & & & \\
\hline

\multirow[t]{8}{*}{3} &
\multirow[t]{8}{=}{broncospasmo} &
\multirow[t]{8}{=}{bronchospasm} &
11 & acuto broncospasmo & acute bronchospasm &
\multirow[t]{8}{*}{0.00} & \multirow[t]{8}{*}{0.00} & \multirow[t]{8}{*}{n} \\
\cline{4-6}
& & & 12 & broncospasmo flogosi vie & bronchospasm airways flogosis & & & \\
\cline{4-6}
& & & 13 & broncospasmo polmonite & bronchospasm pneumonia & & & \\
\cline{4-6}
& & & 14 & broncospasmo corso & ongoing bronchospasm & & & \\
\cline{4-6}
& & & 22 & broncospasmo otite & bronchospasm otitis & & & \\
\cline{4-6}
& & & 23 & broncospasmo parainfettivo & bronchospasm parainfective & & & \\
\cline{4-6}
& & & 24 & broncospasmo virosi & bronchospasm virosis & & & \\
\cline{4-6}
& & & 17 & broncospasmo episodio & bronchospasm episode & & & \\
\hline

\multirow[t]{5}{*}{4} &
\multirow[t]{5}{=}{infezione vie respiratorie} &
\multirow[t]{5}{=}{airways infection} &
18 & infezione alte vie respiratorie & high airways infection &
\multirow[t]{5}{*}{0.00} & \multirow[t]{5}{*}{0.00} & \multirow[t]{5}{*}{n} \\
\cline{4-6}
& & & 19 & flogosi infezione alte vie respiratorie & flogosis high airways infection & & & \\
\cline{4-6}
& & & 27 & altre infezioni acute vie respiratorie & other acute respiratory infection airways & & & \\
\cline{4-6}
& & & 26 & crisi epilessia infezione paziente polmonare respiratoria & seizure epilepsy respiratory infection lung patient & & & \\
\cline{4-6}
& & & 16 & febbricola febbrile infezione respiratoria virosi & febrile fever respiratory infection virosis & & & \\
\hline
\multicolumn{9}{|c|}{\textbf{Bronchitis}} \\ \hline

\multirow[t]{2}{*}{1} &
\multirow[t]{2}{=}{bronchite} &
\multirow[t]{2}{=}{bronchitis} &
1 & bronchite acute & acute bronchitis &
\multirow[t]{2}{*}{88.24} & \multirow[t]{2}{*}{61.86} & \multirow[t]{2}{*}{y} \\
\cline{4-6}
& & & 4 & bronchite & bronchitis & & & \\
\hline

\multirow[t]{3}{*}{2} &
\multirow[t]{3}{=}{broncospasmo} &
\multirow[t]{3}{=}{bronchospasm} &
2 & broncospasmo acuto parainfettivo & acute parainfectious bronchospasm &
\multirow[t]{3}{*}{16.22} & \multirow[t]{3}{*}{12.37} & \multirow[t]{3}{*}{n} \\
\cline{4-6}
& & & 3 & broncospasmo acuto & acute bronchospasm & & & \\
\cline{4-6}
& & & 8 & broncospasmo & bronchospasm & & & \\
\hline

\multirow[t]{2}{*}{3} &
\multirow[t]{2}{=}{infezione basse vie respiratorie} &
\multirow[t]{2}{=}{lower tract respiratory infection} &
6 & insufficienza respiratoria infezione basse & respiratory insufficiency lower infection &
\multirow[t]{2}{*}{31.25} & \multirow[t]{2}{*}{10.31} & \multirow[t]{2}{*}{y} \\
\cline{4-6}
& & & 11 & infezione alte basse vie respiratorie & respiratory infection lower upper tract & & & \\
\hline
\end{tabularx}
}
\end{table}
Second-level clustering substantially reduces the number of clusters, to 20--50\% of the original clusters in our datasets. In the specific case studies, Figure~\ref{fig:labels-eval}.c shows no effect on bronchiolitis weak labels and an increase in recall, at the cost of a decrease in precision, for bronchitis. Beyond these specific case studies, Figure~\ref{fig:labels-eval}.b provides an independent evaluation of clustering quality on a randomly selected sample of 1,000 diagnosis strings. In this broader assessment, second-level clustering increases the proportion of correct assignments and reduces partially correct or incorrect mappings in both datasets. This indicates a systematic refinement of cluster coherence, supporting the role of second-level clustering in improving weak label quality and interpretability, although this does not necessarily translate into substantial gains in downstream classification performance.

The selected clusters cover 80\% of the original bronchiolitis cases and 72\% of the bronchitis cases, as illustrated in Figure~\ref{fig:labels-eval}.c, which compares weak labels derived from clustering with the manually annotated gold labels.

To better understand the false negatives, we reviewed 20 discharge letters labelled as bronchiolitis in the gold standard but not covered by our weak labels. We found 10 cases where the diagnosis refers to other symptoms or diseases (e.g., otitis, dyspnea, fever) while bronchiolitis is mentioned elsewhere in the letter; 4 cases where bronchiolitis appears in the diagnosis alongside other diseases, leading to assignment to a cluster related to the other condition; 3 cases of respiratory issues without explicit mention of bronchiolitis, respiratory syncytial virus, or bronchospasm with fever; 2 cases with no diagnosis string present; and 1 case referring to bronchiolitis in a family member rather than the patient. These results suggest that most false negatives arise from the absence of explicit bronchiolitis indications in the diagnosis strings, rather than from clustering errors.

To assess the effect of the TPT step on clustering results, \ref{app:clustering} reports the clusters obtained using the original UmBERTo model in our clustering pipeline, on the bronchiolitis dataset. The number of clusters related to respiratory infections is higher without TPT (34 vs 27), and their keywords appear more 
heterogeneous. Selecting weak labels by applying the same criterion we used on our clusters, we would cover only 47\% of bronchiolitis cases with the clusters from the original UmBERTo model. Moreover, these clusters are more 
heterogeneous since they introduce more false negatives in the weak labels (precision decreases to 72 \%). 

Additional results on the effect of clustering algorithms and hyperparameters are reported in\ref{app:clustering}. 

\subsection*{Disease classification}
\label{sub:classification-result}

\begin{table}[!htbp]
\centering
\scriptsize
\caption{Classification results on gold labels reported as mean (standard deviation) across cross-validation folds. Bold indicates best overall performance per metric, per dataset. Underline indicates best performance among non-fully-supervised models. Statistical significance was assessed using two-sided paired permutation tests across folds, comparing each model to WS-UmBERTo-TPT. P-values were adjusted using the Holm correction within each metric ($^{*}p_{adj} < 0.05$, $^{**}p_{adj} < 0.01$).}
\label{tab:gold-results}
\resizebox{\textwidth}{!}{%
\begin{tabular}{|l|lrrrrr|rrrrr|}
\hline
\multicolumn{1}{|l|}{\multirow{2}{*}{\textbf{Learning}}} &
  \multicolumn{1}{l|}{\multirow{2}{*}{\textbf{Classifier}}} &
  \multicolumn{5}{c|}{\textbf{Bronchiolitis}} &
  \multicolumn{5}{c|}{\textbf{Bronchitis}} \\ \cline{3-12} 
 &
  \multicolumn{1}{l|}{} &
  \multicolumn{1}{r|}{\textbf{P {[}\%{]}}} &
  \multicolumn{1}{r|}{\textbf{R {[}\%{]}}} &
  \multicolumn{1}{r|}{\textbf{F1 {[}\%{]}}} &
  \multicolumn{1}{r|}{\textbf{AUROC {[}\%{]}}} &
  \multicolumn{1}{r|}{\textbf{AUPRC {[}\%{]}}} &
  \multicolumn{1}{r|}{\textbf{P {[}\%{]}}} &
  \multicolumn{1}{r|}{\textbf{R {[}\%{]}}} &
  \multicolumn{1}{r|}{\textbf{F1 {[}\%{]}}} &
  \multicolumn{1}{r|}{\textbf{AUROC {[}\%{]}}} &
  \textbf{AUPRC {[}\%{]}} \\ \hline
\multicolumn{12}{|c|}{\textbf{With diagnosis string}} \\ \hline
\multirow[t]{8}{*}{Weakly supervised} &
  \multicolumn{1}{|l|}{WS-UmBERTo-TPT} &
  \multicolumn{1}{r|}{\begin{tabular}[c]{@{}r@{}}\underline{78.34}\\ (6.77)\end{tabular}} &
  \multicolumn{1}{r|}{\begin{tabular}[c]{@{}r@{}}\underline{77.57}\\ (6.89)\end{tabular}} &
  \multicolumn{1}{r|}{\begin{tabular}[c]{@{}r@{}}\underline{78.14}\\ (4.89)\end{tabular}} &
  \multicolumn{1}{r|}{\begin{tabular}[c]{@{}r@{}}\underline{77.68}\\ (4.30)\end{tabular}} &
  \multicolumn{1}{r|}{\begin{tabular}[c]{@{}r@{}}\underline{73.13}\\ (4.93)\end{tabular}} &
  \multicolumn{1}{r|}{\begin{tabular}[c]{@{}r@{}}\underline{73.06}\\ (5.95)\end{tabular}} &
  \multicolumn{1}{l|}{\begin{tabular}[c]{@{}l@{}}\underline{82.11}\\ (5.82)\end{tabular}} &
  \multicolumn{1}{r|}{\begin{tabular}[c]{@{}r@{}}\underline{78.02}\\ (5.25)\end{tabular}} &
  \multicolumn{1}{r|}{\begin{tabular}[c]{@{}r@{}}\underline{82.30}\\ (5.10)\end{tabular}} &
  \begin{tabular}[c]{@{}r@{}}\underline{76.72}\\ (5.02)\end{tabular} \\ \cline{2-12}
 &
  \multicolumn{1}{l|}{WS-XLM-RoBERTa-TPT} &
  \multicolumn{1}{r|}{\begin{tabular}[c]{@{}r@{}}69.54$^{*}$\\ (5.45)\end{tabular}} &
  \multicolumn{1}{r|}{\begin{tabular}[c]{@{}r@{}}71.33\\ (5.12)\end{tabular}} &
  \multicolumn{1}{r|}{\begin{tabular}[c]{@{}r@{}}70.85$^{*}$\\ (4.52)\end{tabular}} &
  \multicolumn{1}{r|}{\begin{tabular}[c]{@{}r@{}}69.24$^{*}$\\ (3.89)\end{tabular}} &
  \multicolumn{1}{r|}{\begin{tabular}[c]{@{}r@{}}62.69$^{*}$\\ (3.67)\end{tabular}} &
  \multicolumn{1}{r|}{\begin{tabular}[c]{@{}r@{}}71.57\\ (5.94)\end{tabular}} &
  \multicolumn{1}{l|}{\begin{tabular}[c]{@{}l@{}}75.26$^{*}$\\ (5.46)\end{tabular}} &
  \multicolumn{1}{r|}{\begin{tabular}[c]{@{}r@{}}73.37$^{*}$\\ (5.16)\end{tabular}} &
  \multicolumn{1}{r|}{\begin{tabular}[c]{@{}r@{}}76.41$^{*}$\\ (5.32)\end{tabular}} &
  \begin{tabular}[c]{@{}r@{}}71.01\\ (5.45)\end{tabular} \\ \cline{2-12}
 &
  \multicolumn{1}{l|}{WS-RemBERT-TPT} &
  \multicolumn{1}{r|}{\begin{tabular}[c]{@{}r@{}}67.41$^{*}$\\ (7.01)\end{tabular}} &
  \multicolumn{1}{r|}{\begin{tabular}[c]{@{}r@{}}70.11\\ (6.82)\end{tabular}} &
  \multicolumn{1}{r|}{\begin{tabular}[c]{@{}r@{}}69.23$^{*}$\\ (5.24)\end{tabular}} &
  \multicolumn{1}{r|}{\begin{tabular}[c]{@{}r@{}}68.00$^{*}$\\ (5.03)\end{tabular}} &
  \multicolumn{1}{r|}{\begin{tabular}[c]{@{}r@{}}60.51$^{*}$\\ (3.47)\end{tabular}} &
  \multicolumn{1}{r|}{\begin{tabular}[c]{@{}r@{}}69.23\\ (5.28)\end{tabular}} &
  \multicolumn{1}{l|}{\begin{tabular}[c]{@{}l@{}}74.23$^{*}$\\ (6.12)\end{tabular}} &
  \multicolumn{1}{r|}{\begin{tabular}[c]{@{}r@{}}71.64$^{*}$\\ (5.87)\end{tabular}} &
  \multicolumn{1}{r|}{\begin{tabular}[c]{@{}r@{}}76.15$^{*}$\\ (5.17)\end{tabular}} &
  \begin{tabular}[c]{@{}r@{}}68.39$^{*}$\\ (5.56)\end{tabular} \\ \cline{2-12}
 &
  \multicolumn{1}{l|}{WS-UmBERTo} &
  \multicolumn{1}{r|}{\begin{tabular}[c]{@{}r@{}}68.29\\ (6.53)\end{tabular}} &
  \multicolumn{1}{r|}{\begin{tabular}[c]{@{}r@{}}75.99\\ (6.41)\end{tabular}} &
  \multicolumn{1}{r|}{\begin{tabular}[c]{@{}r@{}}72.15\\ (4.93)\end{tabular}} &
  \multicolumn{1}{r|}{\begin{tabular}[c]{@{}r@{}}71.21\\ (4.21)\end{tabular}} &
  \multicolumn{1}{r|}{\begin{tabular}[c]{@{}r@{}}63.29\\ (5.66)\end{tabular}} &
  \multicolumn{1}{r|}{\begin{tabular}[c]{@{}r@{}}73.27\\ (5.73)\end{tabular}} &
  \multicolumn{1}{l|}{\begin{tabular}[c]{@{}l@{}}76.29$^{*}$\\ (6.19)\end{tabular}} &
  \multicolumn{1}{r|}{\begin{tabular}[c]{@{}r@{}}74.75\\ (5.45)\end{tabular}} &
  \multicolumn{1}{r|}{\begin{tabular}[c]{@{}r@{}}80.31\\ (5.36)\end{tabular}} &
  \begin{tabular}[c]{@{}r@{}}75.29\\ (5.98)\end{tabular} \\ \cline{2-12}
 &
  \multicolumn{1}{l|}{WS-XLM-RoBERTa} &
  \multicolumn{1}{r|}{\begin{tabular}[c]{@{}r@{}}67.49$^{*}$\\ (5.28)\end{tabular}} &
  \multicolumn{1}{r|}{\begin{tabular}[c]{@{}r@{}}68.67$^{*}$\\ (5.09)\end{tabular}} &
  \multicolumn{1}{r|}{\begin{tabular}[c]{@{}r@{}}68.02$^{*}$\\ (4.32)\end{tabular}} &
  \multicolumn{1}{r|}{\begin{tabular}[c]{@{}r@{}}67.65$^{*}$\\ (3.65)\end{tabular}} &
  \multicolumn{1}{r|}{\begin{tabular}[c]{@{}r@{}}61.11$^{*}$\\ (3.23)\end{tabular}} &
  \multicolumn{1}{r|}{\begin{tabular}[c]{@{}r@{}}70.14\\ (6.14)\end{tabular}} &
  \multicolumn{1}{l|}{\begin{tabular}[c]{@{}l@{}}73.96$^{*}$\\ (5.65)\end{tabular}} &
  \multicolumn{1}{r|}{\begin{tabular}[c]{@{}r@{}}72.00$^{*}$\\ (5.49)\end{tabular}} &
  \multicolumn{1}{r|}{\begin{tabular}[c]{@{}r@{}}78.84\\ (5.91)\end{tabular}} &
  \begin{tabular}[c]{@{}r@{}}72.24$^{*}$\\ (6.11)\end{tabular} \\ \cline{2-12}
 &
  \multicolumn{1}{l|}{WS-RemBERT} &
  \multicolumn{1}{r|}{\begin{tabular}[c]{@{}r@{}}65.24$^{*}$\\ (6.81)\end{tabular}} &
  \multicolumn{1}{r|}{\begin{tabular}[c]{@{}r@{}}63.82$^{*}$\\ (6.84)\end{tabular}} &
  \multicolumn{1}{r|}{\begin{tabular}[c]{@{}r@{}}64.96$^{*}$\\ (5.13)\end{tabular}} &
  \multicolumn{1}{r|}{\begin{tabular}[c]{@{}r@{}}65.74$^{*}$\\ (4.92)\end{tabular}} &
  \multicolumn{1}{r|}{\begin{tabular}[c]{@{}r@{}}59.02$^{*}$\\ (3.16)\end{tabular}} &
  \multicolumn{1}{r|}{\begin{tabular}[c]{@{}r@{}}68.41$^{*}$\\ (5.81)\end{tabular}} &
  \multicolumn{1}{l|}{\begin{tabular}[c]{@{}l@{}}73.21$^{*}$\\ (5.47)\end{tabular}} &
  \multicolumn{1}{r|}{\begin{tabular}[c]{@{}r@{}}70.73$^{*}$\\ (5.66)\end{tabular}} &
  \multicolumn{1}{r|}{\begin{tabular}[c]{@{}r@{}}76.52$^{*}$\\ (6.06)\end{tabular}} &
  \begin{tabular}[c]{@{}r@{}}69.20$^{*}$\\ (5.41)\end{tabular} \\ \cline{2-12}
 &
  \multicolumn{1}{l|}{WS-LSTM} &
  \multicolumn{1}{r|}{\begin{tabular}[c]{@{}r@{}}57.39$^{**}$\\ (8.31)\end{tabular}} &
  \multicolumn{1}{r|}{\begin{tabular}[c]{@{}r@{}}60.86$^{**}$\\ (11.52)\end{tabular}} &
  \multicolumn{1}{r|}{\begin{tabular}[c]{@{}r@{}}59.87$^{**}$\\ (9.69)\end{tabular}} &
  \multicolumn{1}{r|}{\begin{tabular}[c]{@{}r@{}}63.29$^{**}$\\ (7.36)\end{tabular}} &
  \multicolumn{1}{r|}{\begin{tabular}[c]{@{}r@{}}49.94$^{**}$\\ (5.13)\end{tabular}} &
  \multicolumn{1}{r|}{\begin{tabular}[c]{@{}r@{}}60.78$^{**}$\\ (6.25)\end{tabular}} &
  \multicolumn{1}{l|}{\begin{tabular}[c]{@{}l@{}}63.92$^{**}$\\ (5.66)\end{tabular}} &
  \multicolumn{1}{r|}{\begin{tabular}[c]{@{}r@{}}62.31$^{**}$\\ (6.28)\end{tabular}} &
  \multicolumn{1}{r|}{\begin{tabular}[c]{@{}r@{}}70.12$^{**}$\\ (6.24)\end{tabular}} &
  \begin{tabular}[c]{@{}r@{}}60.55$^{**}$\\ (5.54)\end{tabular} \\ \cline{2-12}
 &
  \multicolumn{1}{l|}{WS-XPRESS-UmBERTo-TPT} &
  \multicolumn{1}{r|}{\begin{tabular}[c]{@{}r@{}}16.57$^{**}$\\ (2.21)\end{tabular}} &
  \multicolumn{1}{r|}{\begin{tabular}[c]{@{}r@{}}60.52$^{**}$\\ (9.18)\end{tabular}} &
  \multicolumn{1}{r|}{\begin{tabular}[c]{@{}r@{}}29.02$^{**}$\\ (9.04)\end{tabular}} &
  \multicolumn{1}{r|}{\begin{tabular}[c]{@{}r@{}}54.18$^{**}$\\ (10.24)\end{tabular}} &
  \multicolumn{1}{r|}{\begin{tabular}[c]{@{}r@{}}19.91$^{**}$\\ (1.74)\end{tabular}} &
  \multicolumn{1}{r|}{\begin{tabular}[c]{@{}r@{}}38.46$^{**}$\\ (5.96)\end{tabular}} &
  \multicolumn{1}{l|}{\begin{tabular}[c]{@{}l@{}}51.55$^{**}$\\ (5.81)\end{tabular}} &
  \multicolumn{1}{r|}{\begin{tabular}[c]{@{}r@{}}44.05$^{**}$\\ (5.80)\end{tabular}} &
  \multicolumn{1}{r|}{\begin{tabular}[c]{@{}r@{}}56.89$^{**}$\\ (5.77)\end{tabular}} &
  \begin{tabular}[c]{@{}r@{}}42.84$^{**}$\\ (5.45)\end{tabular} \\ \hline
\multicolumn{1}{|l|}{Unsupervised} &
  \multicolumn{1}{l|}{Rule-based full-text} &
  \multicolumn{1}{r|}{\begin{tabular}[c]{@{}r@{}}65.57$^{*}$\\ (6.56)\end{tabular}} &
  \multicolumn{1}{r|}{\begin{tabular}[c]{@{}r@{}}68.76$^{*}$\\ (4.00)\end{tabular}} &
  \multicolumn{1}{r|}{\begin{tabular}[c]{@{}r@{}}67.22$^{*}$\\ (6.19)\end{tabular}} &
  \multicolumn{1}{r|}{-} &
  \multicolumn{1}{r|}{-} &
  \multicolumn{1}{r|}{\begin{tabular}[c]{@{}r@{}}15.79$^{**}$\\ (5.51)\end{tabular}} &
  \multicolumn{1}{l|}{\begin{tabular}[c]{@{}l@{}}32.88$^{**}$\\ (5.67)\end{tabular}} &
  \multicolumn{1}{r|}{\begin{tabular}[c]{@{}r@{}}21.33$^{**}$\\ (5.81)\end{tabular}} &
  \multicolumn{1}{r|}{-} &
  - \\ \cline{2-12}
 &
  \multicolumn{1}{l|}{Rule-based diagnosis} &
  \multicolumn{1}{r|}{\begin{tabular}[c]{@{}r@{}}74.61\\ (4.23)\end{tabular}} &
  \multicolumn{1}{r|}{\begin{tabular}[c]{@{}r@{}}67.42$^{*}$\\ (3.15)\end{tabular}} &
  \multicolumn{1}{r|}{\begin{tabular}[c]{@{}r@{}}70.74$^{*}$\\ (3.41)\end{tabular}} &
  \multicolumn{1}{r|}{-} &
  \multicolumn{1}{r|}{-} &
  \multicolumn{1}{r|}{\begin{tabular}[c]{@{}r@{}}21.74$^{**}$\\ (6.15)\end{tabular}} &
  \multicolumn{1}{l|}{\begin{tabular}[c]{@{}l@{}}20.55$^{**}$\\ (5.68)\end{tabular}} &
  \multicolumn{1}{r|}{\begin{tabular}[c]{@{}r@{}}21.13$^{**}$\\ (5.87)\end{tabular}} &
  \multicolumn{1}{r|}{-} &
  - \\ \cline{2-12}
 &
  \multicolumn{1}{l|}{LLaMAntino-3} &
  \multicolumn{1}{r|}{\begin{tabular}[c]{@{}r@{}}78.26\\ (5.72)\end{tabular}} &
  \multicolumn{1}{r|}{\begin{tabular}[c]{@{}r@{}}74.64$^{*}$\\ (7.48)\end{tabular}} &
  \multicolumn{1}{r|}{\begin{tabular}[c]{@{}r@{}}75.12\\ (4.29)\end{tabular}} &
  \multicolumn{1}{r|}{\begin{tabular}[c]{@{}r@{}}71.67$^{*}$\\ (3.48)\end{tabular}} &
  \multicolumn{1}{r|}{\begin{tabular}[c]{@{}r@{}}67.92$^{*}$\\ (4.47)\end{tabular}} &
  \multicolumn{1}{r|}{\begin{tabular}[c]{@{}r@{}}65.21$^{*}$\\ (5.51)\end{tabular}} &
  \multicolumn{1}{l|}{\begin{tabular}[c]{@{}l@{}}72.61$^{*}$\\ (5.89)\end{tabular}} &
  \multicolumn{1}{r|}{\begin{tabular}[c]{@{}r@{}}68.71$^{*}$\\ (5.67)\end{tabular}} &
  \multicolumn{1}{r|}{\begin{tabular}[c]{@{}r@{}}75.29$^{*}$\\ (5.68)\end{tabular}} &
  \begin{tabular}[c]{@{}r@{}}68.13$^{*}$\\ (4.74)\end{tabular} \\ \hline
\multicolumn{1}{|l|}{Supervised} &
  \multicolumn{1}{l|}{FS-UmBERTo-TPT} &
  \multicolumn{1}{r|}{\begin{tabular}[c]{@{}r@{}}\textbf{80.30}\\ (6.24)\end{tabular}} &
  \multicolumn{1}{r|}{\begin{tabular}[c]{@{}r@{}}\textbf{81.52}$^{*}$\\ (6.59)\end{tabular}} &
  \multicolumn{1}{r|}{\begin{tabular}[c]{@{}r@{}}\textbf{80.83}\\ (4.12)\end{tabular}} &
  \multicolumn{1}{r|}{\begin{tabular}[c]{@{}r@{}}\textbf{83.45}$^{*}$\\ (4.02)\end{tabular}} &
  \multicolumn{1}{r|}{\begin{tabular}[c]{@{}r@{}}\textbf{77.57}$^{*}$\\ (2.02)\end{tabular}} &
  \multicolumn{1}{r|}{\begin{tabular}[c]{@{}r@{}}\textbf{74.49}\\ (4.45)\end{tabular}} &
  \multicolumn{1}{l|}{\begin{tabular}[c]{@{}l@{}}\textbf{82.47}\\ (4.68)\end{tabular}} &
  \multicolumn{1}{r|}{\begin{tabular}[c]{@{}r@{}}\textbf{78.28}\\ (4.51)\end{tabular}} &
  \multicolumn{1}{r|}{\begin{tabular}[c]{@{}r@{}}\textbf{85.93}\\ (4.47)\end{tabular}} &
  \begin{tabular}[c]{@{}r@{}}\textbf{79.48}\\ (3.52)\end{tabular} \\ \cline{2-12}
 &
  \multicolumn{1}{l|}{FS-XLM-RoBERTa-TPT} &
  \multicolumn{1}{r|}{\begin{tabular}[c]{@{}r@{}}75.29$^{*}$\\ (5.21)\end{tabular}} &
  \multicolumn{1}{r|}{\begin{tabular}[c]{@{}r@{}}78.31\\ (4.36)\end{tabular}} &
  \multicolumn{1}{r|}{\begin{tabular}[c]{@{}r@{}}76.54\\ (4.14)\end{tabular}} &
  \multicolumn{1}{r|}{\begin{tabular}[c]{@{}r@{}}78.54\\ (3.42)\end{tabular}} &
  \multicolumn{1}{r|}{\begin{tabular}[c]{@{}r@{}}72.23\\ (2.70)\end{tabular}} &
  \multicolumn{1}{r|}{\begin{tabular}[c]{@{}r@{}}74.29\\ (5.55)\end{tabular}} &
  \multicolumn{1}{l|}{\begin{tabular}[c]{@{}l@{}}80.41\\ (5.01)\end{tabular}} &
  \multicolumn{1}{r|}{\begin{tabular}[c]{@{}r@{}}77.23\\ (4.74)\end{tabular}} &
  \multicolumn{1}{r|}{\begin{tabular}[c]{@{}r@{}}83.82\\ (5.54)\end{tabular}} &
  \begin{tabular}[c]{@{}r@{}}76.10\\ (4.61)\end{tabular} \\ \cline{2-12}
 &
  \multicolumn{1}{l|}{FS-RemBERT-TPT} &
  \multicolumn{1}{r|}{\begin{tabular}[c]{@{}r@{}}73.24\\ (6.58)\end{tabular}} &
  \multicolumn{1}{r|}{\begin{tabular}[c]{@{}r@{}}75.28\\ (5.93)\end{tabular}} &
  \multicolumn{1}{r|}{\begin{tabular}[c]{@{}r@{}}74.21$^{*}$\\ (4.68)\end{tabular}} &
  \multicolumn{1}{r|}{\begin{tabular}[c]{@{}r@{}}76.39\\ (4.31)\end{tabular}} &
  \multicolumn{1}{r|}{\begin{tabular}[c]{@{}r@{}}68.33\\ (5.76)\end{tabular}} &
  \multicolumn{1}{r|}{\begin{tabular}[c]{@{}r@{}}72.66\\ (5.52)\end{tabular}} &
  \multicolumn{1}{l|}{\begin{tabular}[c]{@{}l@{}}76.70\\ (5.54)\end{tabular}} &
  \multicolumn{1}{r|}{\begin{tabular}[c]{@{}r@{}}74.62\\ (6.12)\end{tabular}} &
  \multicolumn{1}{r|}{\begin{tabular}[c]{@{}r@{}}83.14\\ (5.73)\end{tabular}} &
  \begin{tabular}[c]{@{}r@{}}75.26\\ (5.61)\end{tabular} \\ \hline
\multicolumn{12}{|c|}{\textbf{Without diagnosis string}} \\ \hline
\multicolumn{1}{|l|}{Weakly supervised} &
  \multicolumn{1}{l|}{WS-UmBERTo-TPT} &
  \multicolumn{1}{r|}{\begin{tabular}[c]{@{}r@{}}\underline{74.25}\\ (6.54)\end{tabular}} &
  \multicolumn{1}{r|}{\begin{tabular}[c]{@{}r@{}}73.52\\ (6.12)\end{tabular}} &
  \multicolumn{1}{r|}{\begin{tabular}[c]{@{}r@{}}\underline{73.89}\\ (5.64)\end{tabular}} &
  \multicolumn{1}{r|}{\begin{tabular}[c]{@{}r@{}}\underline{74.27}\\ (4.83)\end{tabular}} &
  \multicolumn{1}{r|}{\begin{tabular}[c]{@{}r@{}}\underline{69.27}\\ (4.36)\end{tabular}} &
  \multicolumn{1}{r|}{\begin{tabular}[c]{@{}r@{}}\underline{72.09}\\ (5.86)\end{tabular}} &
  \multicolumn{1}{l|}{\begin{tabular}[c]{@{}l@{}}\underline{\textbf{79.90}}\\ (5.81)\end{tabular}} &
  \multicolumn{1}{r|}{\begin{tabular}[c]{@{}r@{}}\underline{75.79}\\ (5.80)\end{tabular}} &
  \multicolumn{1}{r|}{\begin{tabular}[c]{@{}r@{}}\underline{79.26}\\ (5.74)\end{tabular}} &
  \begin{tabular}[c]{@{}r@{}}\underline{74.71}\\ (5.18)\end{tabular} \\ \cline{2-12}
 &
  \multicolumn{1}{l|}{WS-XLM-RoBERTa-TPT} &
  \multicolumn{1}{r|}{\begin{tabular}[c]{@{}r@{}}66.83$^{*}$\\ (5.69)\end{tabular}} &
  \multicolumn{1}{r|}{\begin{tabular}[c]{@{}r@{}}67.52$^{*}$\\ (5.88)\end{tabular}} &
  \multicolumn{1}{r|}{\begin{tabular}[c]{@{}r@{}}67.19$^{*}$\\ (4.57)\end{tabular}} &
  \multicolumn{1}{r|}{\begin{tabular}[c]{@{}r@{}}67.25$^{*}$\\ (3.65)\end{tabular}} &
  \multicolumn{1}{r|}{\begin{tabular}[c]{@{}r@{}}59.29$^{*}$\\ (4.52)\end{tabular}} &
  \multicolumn{1}{r|}{\begin{tabular}[c]{@{}r@{}}69.39\\ (5.86)\end{tabular}} &
  \multicolumn{1}{l|}{\begin{tabular}[c]{@{}l@{}}70.10$^{*}$\\ (5.51)\end{tabular}} &
  \multicolumn{1}{r|}{\begin{tabular}[c]{@{}r@{}}69.74$^{*}$\\ (5.84)\end{tabular}} &
  \multicolumn{1}{r|}{\begin{tabular}[c]{@{}r@{}}75.13\\ (5.64)\end{tabular}} &
  \begin{tabular}[c]{@{}r@{}}69.33$^{*}$\\ (5.54)\end{tabular} \\ \cline{2-12}
 &
  \multicolumn{1}{l|}{WS-RemBERT-TPT} &
  \multicolumn{1}{r|}{\begin{tabular}[c]{@{}r@{}}64.53$^{*}$\\ (6.58)\end{tabular}} &
  \multicolumn{1}{r|}{\begin{tabular}[c]{@{}r@{}}63.66$^{*}$\\ (6.21)\end{tabular}} &
  \multicolumn{1}{r|}{\begin{tabular}[c]{@{}r@{}}64.21$^{*}$\\ (5.84)\end{tabular}} &
  \multicolumn{1}{r|}{\begin{tabular}[c]{@{}r@{}}65.47$^{*}$\\ (4.97)\end{tabular}} &
  \multicolumn{1}{r|}{\begin{tabular}[c]{@{}r@{}}56.75$^{*}$\\ (3.43)\end{tabular}} &
  \multicolumn{1}{r|}{\begin{tabular}[c]{@{}r@{}}69.31\\ (5.56)\end{tabular}} &
  \multicolumn{1}{l|}{\begin{tabular}[c]{@{}l@{}}72.16$^{*}$\\ (5.97)\end{tabular}} &
  \multicolumn{1}{r|}{\begin{tabular}[c]{@{}r@{}}70.71$^{*}$\\ (5.63)\end{tabular}} &
  \multicolumn{1}{r|}{\begin{tabular}[c]{@{}r@{}}74.26$^{*}$\\ (6.38)\end{tabular}} &
  \begin{tabular}[c]{@{}r@{}}69.91$^{*}$\\ (5.61)\end{tabular} \\ \cline{2-12}
 &
  \multicolumn{1}{l|}{WS-UmBERTo} &
  \multicolumn{1}{r|}{\begin{tabular}[c]{@{}r@{}}67.55$^{*}$\\ (6.38)\end{tabular}} &
  \multicolumn{1}{r|}{\begin{tabular}[c]{@{}r@{}}\underline{\textbf{76.40}}\\ (6.25)\end{tabular}} &
  \multicolumn{1}{r|}{\begin{tabular}[c]{@{}r@{}}71.74\\ (5.47)\end{tabular}} &
  \multicolumn{1}{r|}{\begin{tabular}[c]{@{}r@{}}69.98\\ (4.59)\end{tabular}} &
  \multicolumn{1}{r|}{\begin{tabular}[c]{@{}r@{}}63.27\\ (2.66)\end{tabular}} &
  \multicolumn{1}{r|}{\begin{tabular}[c]{@{}r@{}}72.00\\ (6.01)\end{tabular}} &
  \multicolumn{1}{l|}{\begin{tabular}[c]{@{}l@{}}74.23$^{*}$\\ (5.67)\end{tabular}} &
  \multicolumn{1}{r|}{\begin{tabular}[c]{@{}r@{}}73.10\\ (5.73)\end{tabular}} &
  \multicolumn{1}{r|}{\begin{tabular}[c]{@{}r@{}}73.45$^{*}$\\ (5.39)\end{tabular}} &
  \begin{tabular}[c]{@{}r@{}}69.09\\ (5.84)\end{tabular} \\ \cline{2-12}
 &
  \multicolumn{1}{l|}{WS-XLM-RoBERTa} &
  \multicolumn{1}{r|}{\begin{tabular}[c]{@{}r@{}}63.98$^{*}$\\ (5.72)\end{tabular}} &
  \multicolumn{1}{r|}{\begin{tabular}[c]{@{}r@{}}65.04$^{*}$\\ (5.63)\end{tabular}} &
  \multicolumn{1}{r|}{\begin{tabular}[c]{@{}r@{}}64.69$^{*}$\\ (4.98)\end{tabular}} &
  \multicolumn{1}{r|}{\begin{tabular}[c]{@{}r@{}}64.87$^{*}$\\ (4.12)\end{tabular}} &
  \multicolumn{1}{r|}{\begin{tabular}[c]{@{}r@{}}54.30$^{*}$\\ (3.31)\end{tabular}} &
  \multicolumn{1}{r|}{\begin{tabular}[c]{@{}r@{}}68.26\\ (5.64)\end{tabular}} &
  \multicolumn{1}{l|}{\begin{tabular}[c]{@{}l@{}}69.97$^{*}$\\ (5.11)\end{tabular}} &
  \multicolumn{1}{r|}{\begin{tabular}[c]{@{}r@{}}69.10$^{*}$\\ (5.88)\end{tabular}} &
  \multicolumn{1}{r|}{\begin{tabular}[c]{@{}r@{}}70.87$^{*}$\\ (5.42)\end{tabular}} &
  \begin{tabular}[c]{@{}r@{}}65.39$^{*}$\\ (5.15)\end{tabular} \\ \cline{2-12}
 &
  \multicolumn{1}{l|}{WS-RemBERT} &
  \multicolumn{1}{r|}{\begin{tabular}[c]{@{}r@{}}62.25$^{*}$\\ (6.58)\end{tabular}} &
  \multicolumn{1}{r|}{\begin{tabular}[c]{@{}r@{}}60.48$^{**}$\\ (6.96)\end{tabular}} &
  \multicolumn{1}{r|}{\begin{tabular}[c]{@{}r@{}}61.65$^{*}$\\ (5.24)\end{tabular}} &
  \multicolumn{1}{r|}{\begin{tabular}[c]{@{}r@{}}62.21$^{*}$\\ (4.87)\end{tabular}} &
  \multicolumn{1}{r|}{\begin{tabular}[c]{@{}r@{}}55.87$^{*}$\\ (4.71)\end{tabular}} &
  \multicolumn{1}{r|}{\begin{tabular}[c]{@{}r@{}}68.10\\ (5.26)\end{tabular}} &
  \multicolumn{1}{l|}{\begin{tabular}[c]{@{}l@{}}70.16$^{*}$\\ (5.69)\end{tabular}} &
  \multicolumn{1}{r|}{\begin{tabular}[c]{@{}r@{}}69.11$^{*}$\\ (5.54)\end{tabular}} &
  \multicolumn{1}{r|}{\begin{tabular}[c]{@{}r@{}}73.14$^{*}$\\ (5.91)\end{tabular}} &
  \begin{tabular}[c]{@{}r@{}}69.34$^{*}$\\ (5.82)\end{tabular} \\ \cline{2-12}
 &
  \multicolumn{1}{l|}{WS-LSTM} &
  \multicolumn{1}{r|}{\begin{tabular}[c]{@{}r@{}}50.14$^{**}$\\ (7.14)\end{tabular}} &
  \multicolumn{1}{r|}{\begin{tabular}[c]{@{}r@{}}63.97$^{**}$\\ (10.12)\end{tabular}} &
  \multicolumn{1}{r|}{\begin{tabular}[c]{@{}r@{}}56.34$^{**}$\\ (8.65)\end{tabular}} &
  \multicolumn{1}{r|}{\begin{tabular}[c]{@{}r@{}}59.02$^{**}$\\ (6.51)\end{tabular}} &
  \multicolumn{1}{r|}{\begin{tabular}[c]{@{}r@{}}43.30$^{**}$\\ (4.74)\end{tabular}} &
  \multicolumn{1}{r|}{\begin{tabular}[c]{@{}r@{}}58.45$^{**}$\\ (5.51)\end{tabular}} &
  \multicolumn{1}{l|}{\begin{tabular}[c]{@{}l@{}}61.87$^{**}$\\ (5.25)\end{tabular}} &
  \multicolumn{1}{r|}{\begin{tabular}[c]{@{}r@{}}60.11$^{**}$\\ (5.31)\end{tabular}} &
  \multicolumn{1}{r|}{\begin{tabular}[c]{@{}r@{}}66.52$^{**}$\\ (5.37)\end{tabular}} &
  \begin{tabular}[c]{@{}r@{}}60.24$^{**}$\\ (5.36)\end{tabular} \\ \cline{2-12}
 &
  \multicolumn{1}{l|}{WS-XPRESS-UmBERTo-TPT} &
  \multicolumn{1}{r|}{\begin{tabular}[c]{@{}r@{}}16.03$^{**}$\\ (6.52)\end{tabular}} &
  \multicolumn{1}{r|}{\begin{tabular}[c]{@{}r@{}}64.50$^{*}$\\ (12.62)\end{tabular}} &
  \multicolumn{1}{r|}{\begin{tabular}[c]{@{}r@{}}27.36$^{**}$\\ (8.48)\end{tabular}} &
  \multicolumn{1}{r|}{\begin{tabular}[c]{@{}r@{}}52.23$^{**}$\\ (10.12)\end{tabular}} &
  \multicolumn{1}{r|}{\begin{tabular}[c]{@{}r@{}}17.89$^{**}$\\ (2.33)\end{tabular}} &
  \multicolumn{1}{r|}{\begin{tabular}[c]{@{}r@{}}35.47$^{**}$\\ (5.54)\end{tabular}} &
  \multicolumn{1}{l|}{\begin{tabular}[c]{@{}l@{}}50.51$^{**}$\\ (6.24)\end{tabular}} &
  \multicolumn{1}{r|}{\begin{tabular}[c]{@{}r@{}}41.67$^{**}$\\ (5.94)\end{tabular}} &
  \multicolumn{1}{r|}{\begin{tabular}[c]{@{}r@{}}52.28$^{**}$\\ (5.64)\end{tabular}} &
  \begin{tabular}[c]{@{}r@{}}45.43$^{**}$\\ (6.02)\end{tabular} \\ \hline
\multicolumn{1}{|l|}{Unsupervised} &
  \multicolumn{1}{l|}{Rule-based full-text} &
  \multicolumn{1}{r|}{\begin{tabular}[c]{@{}r@{}}62.33$^{**}$\\ (6.78)\end{tabular}} &
  \multicolumn{1}{r|}{\begin{tabular}[c]{@{}r@{}}60.05$^{**}$\\ (8.42)\end{tabular}} &
  \multicolumn{1}{r|}{\begin{tabular}[c]{@{}r@{}}61.15$^{**}$\\ (6.83)\end{tabular}} &
  \multicolumn{1}{r|}{-} &
  \multicolumn{1}{r|}{-} &
  \multicolumn{1}{r|}{\begin{tabular}[c]{@{}r@{}}14.09$^{**}$\\ (5.61)\end{tabular}} &
  \multicolumn{1}{l|}{\begin{tabular}[c]{@{}l@{}}30.00$^{**}$\\ (5.89)\end{tabular}} &
  \multicolumn{1}{r|}{\begin{tabular}[c]{@{}r@{}}19.18$^{**}$\\ (5.76)\end{tabular}} &
  \multicolumn{1}{r|}{-} &
  - \\ \cline{2-12}
 &
  \multicolumn{1}{l|}{LLaMAntino-3} &
  \multicolumn{1}{r|}{\begin{tabular}[c]{@{}r@{}}70.42$^{*}$\\ (5.26)\end{tabular}} &
  \multicolumn{1}{r|}{\begin{tabular}[c]{@{}r@{}}72.98\\ (5.84)\end{tabular}} &
  \multicolumn{1}{r|}{\begin{tabular}[c]{@{}r@{}}71.16\\ (3.98)\end{tabular}} &
  \multicolumn{1}{r|}{\begin{tabular}[c]{@{}r@{}}68.52$^{*}$\\ (4.55)\end{tabular}} &
  \multicolumn{1}{r|}{\begin{tabular}[c]{@{}r@{}}62.24$^{*}$\\ (4.22)\end{tabular}} &
  \multicolumn{1}{r|}{\begin{tabular}[c]{@{}r@{}}60.12$^{*}$\\ (5.01)\end{tabular}} &
  \multicolumn{1}{l|}{\begin{tabular}[c]{@{}l@{}}67.25$^{*}$\\ (5.15)\end{tabular}} &
  \multicolumn{1}{r|}{\begin{tabular}[c]{@{}r@{}}63.49$^{*}$\\ (5.23)\end{tabular}} &
  \multicolumn{1}{r|}{\begin{tabular}[c]{@{}r@{}}68.21$^{*}$\\ (5.02)\end{tabular}} &
  \begin{tabular}[c]{@{}r@{}}59.31$^{*}$\\ (2.74)\end{tabular} \\ \hline
\multicolumn{1}{|l|}{Supervised} &
  \multicolumn{1}{l|}{FS-UmBERTo-TPT} &
  \multicolumn{1}{r|}{\begin{tabular}[c]{@{}r@{}}\textbf{78.25}$^{*}$\\ (5.20)\end{tabular}} &
  \multicolumn{1}{r|}{\begin{tabular}[c]{@{}r@{}}76.21\\ (4.87)\end{tabular}} &
  \multicolumn{1}{r|}{\begin{tabular}[c]{@{}r@{}}\textbf{77.15}\\ (4.01)\end{tabular}} &
  \multicolumn{1}{r|}{\begin{tabular}[c]{@{}r@{}}\textbf{78.51}$^{*}$\\ (3.84)\end{tabular}} &
  \multicolumn{1}{r|}{\begin{tabular}[c]{@{}r@{}}\textbf{75.44}$^{*}$\\ (3.55)\end{tabular}} &
  \multicolumn{1}{r|}{\begin{tabular}[c]{@{}r@{}}\textbf{72.99}\\ (5.36)\end{tabular}} &
  \multicolumn{1}{l|}{\begin{tabular}[c]{@{}l@{}}79.59\\ (5.96)\end{tabular}} &
  \multicolumn{1}{r|}{\begin{tabular}[c]{@{}r@{}}\textbf{76.36}\\ (5.84)\end{tabular}} &
  \multicolumn{1}{r|}{\begin{tabular}[c]{@{}r@{}}\textbf{82.63}\\ (5.64)\end{tabular}} &
  \begin{tabular}[c]{@{}r@{}}\textbf{76.82}\\ (6.37)\end{tabular} \\ \cline{2-12}
 &
  \multicolumn{1}{l|}{FS-XLM-RoBERTa-TPT} &
  \multicolumn{1}{r|}{\begin{tabular}[c]{@{}r@{}}72.54\\ (4.26)\end{tabular}} &
  \multicolumn{1}{r|}{\begin{tabular}[c]{@{}r@{}}73.65\\ (4.16)\end{tabular}} &
  \multicolumn{1}{r|}{\begin{tabular}[c]{@{}r@{}}72.96\\ (3.19)\end{tabular}} &
  \multicolumn{1}{r|}{\begin{tabular}[c]{@{}r@{}}74.08\\ (3.34)\end{tabular}} &
  \multicolumn{1}{r|}{\begin{tabular}[c]{@{}r@{}}70.61\\ (3.39)\end{tabular}} &
  \multicolumn{1}{r|}{\begin{tabular}[c]{@{}r@{}}73.33\\ (5.51)\end{tabular}} &
  \multicolumn{1}{l|}{\begin{tabular}[c]{@{}l@{}}79.38\\ (6.11)\end{tabular}} &
  \multicolumn{1}{r|}{\begin{tabular}[c]{@{}r@{}}76.24\\ (5.43)\end{tabular}} &
  \multicolumn{1}{r|}{\begin{tabular}[c]{@{}r@{}}81.15\\ (5.30)\end{tabular}} &
  \begin{tabular}[c]{@{}r@{}}74.34\\ (5.38)\end{tabular} \\ \cline{2-12}
 &
  \multicolumn{1}{l|}{FS-RemBERT-TPT} &
  \multicolumn{1}{r|}{\begin{tabular}[c]{@{}r@{}}71.49\\ (5.46)\end{tabular}} &
  \multicolumn{1}{r|}{\begin{tabular}[c]{@{}r@{}}70.87\\ (4.12)\end{tabular}} &
  \multicolumn{1}{r|}{\begin{tabular}[c]{@{}r@{}}71.21\\ (3.95)\end{tabular}} &
  \multicolumn{1}{r|}{\begin{tabular}[c]{@{}r@{}}72.68\\ (3.87)\end{tabular}} &
  \multicolumn{1}{r|}{\begin{tabular}[c]{@{}r@{}}64.06\\ (4.59)\end{tabular}} &
  \multicolumn{1}{r|}{\begin{tabular}[c]{@{}r@{}}71.57\\ (5.63)\end{tabular}} &
  \multicolumn{1}{l|}{\begin{tabular}[c]{@{}l@{}}75.26\\ (5.84)\end{tabular}} &
  \multicolumn{1}{r|}{\begin{tabular}[c]{@{}r@{}}73.37\\ (5.69)\end{tabular}} &
  \multicolumn{1}{r|}{\begin{tabular}[c]{@{}r@{}}78.26\\ (5.88)\end{tabular}} &
  \begin{tabular}[c]{@{}r@{}}71.28\\ (5.67)\end{tabular} \\ \hline
\end{tabular}
}
\end{table}

Table \ref{tab:gold-results} reports the cross-validation results for the classification task, including precision, recall, F1-score, AUROC, and AUPRC. The table compares our weakly supervised pipeline with different transformer-based classifiers, fully unsupervised rule-based and zero-shot LLM approaches, and fully supervised versions of the transformer-based classifiers. Results are shown for both datasets and for two experimental settings: keeping the diagnosis string in the discharge letter or removing it.

On the bronchiolitis dataset, our weakly supervised pipeline outperforms fully unsupervised approaches and maintains only a limited gap with the fully supervised models. The TPT step consistently improves performance across models: the differences are limited, but there are no cases in which TPT hurts performance. Among the transformer-based classifiers, UmBERTo generally achieves the best results, outperforming RemBERT and XLM-RoBERTa across most metrics. The LSTM classifier shows a clear performance gap with respect to the transformer-based models.

\textit{Rule-based full-text}, which verifies the presence of keywords in the entire letter, shows lower precision and recall than the majority of the weakly supervised models, emphasizing the need for a more sophisticated approach. The results further decline when the diagnosis string is excluded. \textit{Rule-based diagnosis}, using only diagnosis strings, yields better precision than the rule applied to the complete text, though recall remains similar.

The zero-shot approach with the Italian LLaMAntino-3 LLM does not outperform our best weakly supervised results but shows competitive performance.

The model trained using weak labels derived from the XPRESS framework performs substantially worse than all other approaches in this setting. This is expected, since this implementation of the XPRESS framework applies the same keyword rules defined for our cluster selection across the entire clinical text, generating a larger number of noisy weak labels compared with the more targeted extraction of diagnosis strings.

Results on the bronchitis validation dataset show a broadly similar ranking of models, suggesting consistent trends with the main case study. Despite the substantially smaller dataset size, transformer-based classifiers again outperform rule-based and zero-shot approaches, and UmBERTo-based models achieve the best overall performance among the weakly supervised methods. These findings, however, should be interpreted with caution given the limited number of positive bronchitis cases.

We also repeated the bronchitis experiments without applying the second-level clustering step when generating weak labels. This led to a small decrease in performance (AUPRC reduction of 1-2\%), indicating that the impact of second-level clustering on downstream performance is limited in this setting. Detailed results are reported in \ref{app:classification}.

\begin{figure}[t]
    \centering
    \includegraphics[width=\linewidth]{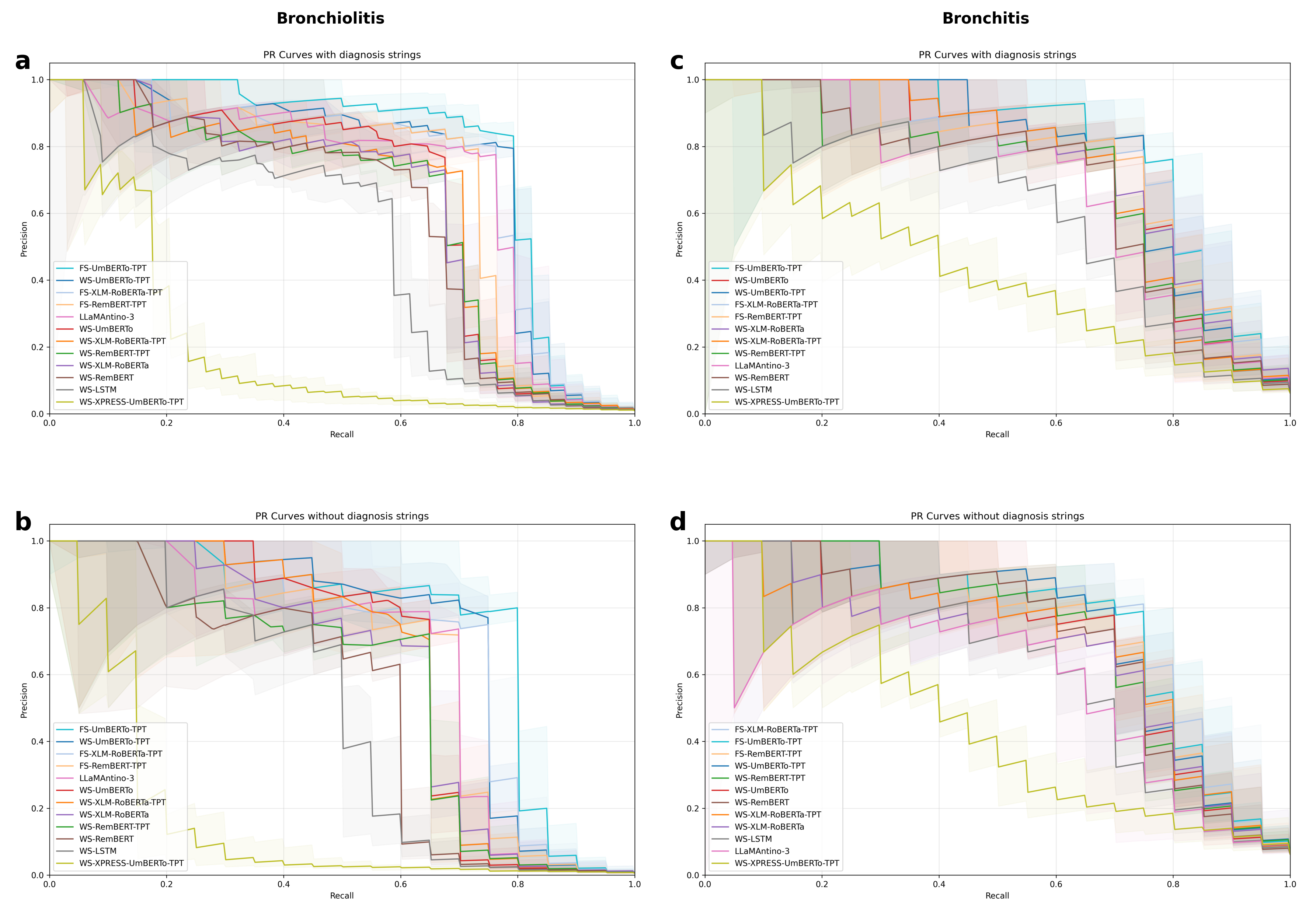}
    \caption{Precision-Recall curves on the bronchiolitis (a,b) and bronchitis (c,d) gold labels with (a,c) and without (b,d) diagnosis strings. Shaded areas represent 95\% C.I.}
    \label{fig:pr-curves}
\end{figure}

Precision–recall curves for the different models are reported in Figure~\ref{fig:pr-curves}, providing a complementary view of model ranking performance across recall levels.

\begin{figure}[ph]
    \centering
    \includegraphics[width=0.8\linewidth]{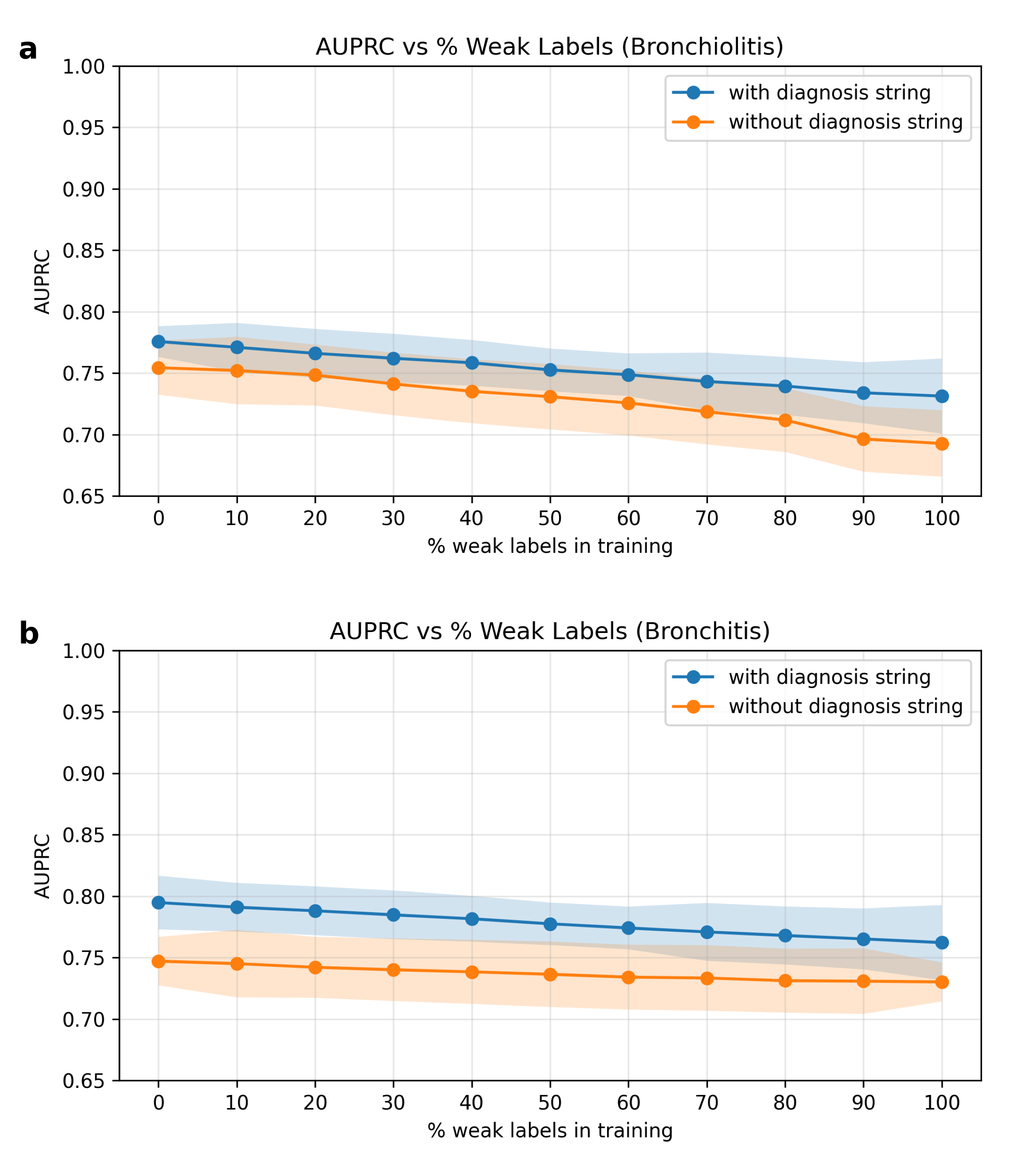}
    \caption{AUPRC with varying percentages of weak labels on the bronchiolitis (a) and bronchitis (b) datasets. Mean AUPRC ($\pm95\%$ CI) on gold labels, with and without diagnosis strings, as different proportions of gold labels are replaced by weak labels. Results are averaged over 5 repetitions of cross-validation. The y-axis is truncated to improve visibility of the trend.}
    \label{fig:auprc-mixedlabels}
\end{figure}

In Figure~\ref{fig:auprc-mixedlabels}, we observe the decrease in performance when replacing increasing percentages of gold labels with weak labels. This scenario is representative of many real-world settings where only a fraction of the data can be manually annotated. The results show a generally smooth and monotonic decrease in AUPRC as the proportion of weak labels increases, with similar trends observed for both bronchiolitis and bronchitis.

\begin{table}[!htbp]
\centering
\caption{Classification results after removing one 1st level cluster for weak labels at a time, on the best weakly supervised classification models. Cross validation results with std. dev. in parenthesis.}
\label{tab:res-sens}
\resizebox{\textwidth}{!}{%
\begin{tabular}{|lrrrrr|}
\hline
\multicolumn{1}{|l|}{\textbf{Clusters for weak labels}} &
  \multicolumn{1}{r|}{\textbf{P {[}\%{]}}} &
  \multicolumn{1}{r|}{\textbf{R {[}\%{]}}} &
  \multicolumn{1}{r|}{\textbf{F1 {[}\%{]}}} &
  \multicolumn{1}{r|}{\textbf{AUROC {[}\%{]}}} &
  \multicolumn{1}{r|}{\textbf{AUPRC {[}\%{]}}} \\ \hline
\multicolumn{6}{|c|}{\textbf{Bronchiolitis}} \\ \hline
\multicolumn{1}{|l|}{\#1. \#2. \#3. \#4.} &
  \multicolumn{1}{r|}{70.75 (6.48)} &
  \multicolumn{1}{r|}{81.27 (6.99)} &
  \multicolumn{1}{r|}{75.65 (4.29)} &
  \multicolumn{1}{r|}{75.49 (5.57)} &
  67.95 (4.32) \\ \hline
\multicolumn{1}{|l|}{\#1. \#2. \#4. \#5.} &
  \multicolumn{1}{r|}{68.42 (8.70)} &
  \multicolumn{1}{r|}{75.09 (8.58)} &
  \multicolumn{1}{r|}{71.60 (7.26)} &
  \multicolumn{1}{r|}{72.67 (6.87)} &
  61.44 (4.75) \\ \hline
\multicolumn{1}{|l|}{\#2. \#3. \#4. \#5.} &
  \multicolumn{1}{r|}{68.54 (6.30)} &
  \multicolumn{1}{r|}{82.35 (4.94)} &
  \multicolumn{1}{r|}{74.81 (3.82)} &
  \multicolumn{1}{r|}{75.83 (5.22)} &
  66.07 (3.64) \\ \hline
\multicolumn{1}{|l|}{\#1. \#3. \#4. \#5.} &
  \multicolumn{1}{r|}{57.29 (12.97)} &
  \multicolumn{1}{r|}{59.03 (15.94)} &
  \multicolumn{1}{r|}{58.15 (12.45)} &
  \multicolumn{1}{r|}{68.90 (10.04)} &
  47.79 (3.39) \\ \hline
\multicolumn{1}{|l|}{\#1. \#2. \#3. \#5.} &
  \multicolumn{1}{r|}{67.86 (6.36)} &
  \multicolumn{1}{r|}{73.42 (8.64)} &
  \multicolumn{1}{r|}{70.53 (5.33)} &
  \multicolumn{1}{r|}{70.22 (3.10)} &
  62.81 (3.19) \\ \hline
\multicolumn{6}{|c|}{\textbf{Bronchitis}} \\ \hline
\multicolumn{1}{|l|}{\#1. \#4. \#6.} &
  \multicolumn{1}{l|}{73.41 (5.84)} &
  \multicolumn{1}{l|}{79.12 (5.71)} &
  \multicolumn{1}{l|}{76.26 (5.19)} &
  \multicolumn{1}{l|}{79.66 (5.22)} &
  75.03 (5.87) \\ \hline
\multicolumn{1}{|l|}{\#1. \#4. \#11.} &
  \multicolumn{1}{l|}{75.67 (5.96)} &
  \multicolumn{1}{l|}{77.23 (6.05)} &
  \multicolumn{1}{l|}{76.64 (5.85)} &
  \multicolumn{1}{l|}{77.22 (5.48)} &
  73.78 (6.09) \\ \hline
\multicolumn{1}{|l|}{\#1. \#6. \#11.} &
  \multicolumn{1}{l|}{66.52 (6.41)} &
  \multicolumn{1}{l|}{70.12 (6.84)} &
  \multicolumn{1}{l|}{68.35 (6.35)} &
  \multicolumn{1}{l|}{70.52 (6.07)} &
  61.91 (5.68) \\ \hline
\multicolumn{1}{|l|}{\#4. \#6. \#11.} &
  \multicolumn{1}{l|}{62.15 (6.96)} &
  \multicolumn{1}{l|}{60.23 (6.84)} &
  \multicolumn{1}{l|}{61.15 (6.40)} &
  \multicolumn{1}{l|}{61.22 (5.45)} &
  56.68 (6.25) \\ \hline
\end{tabular}
}
\end{table}

In Table~\ref{tab:res-sens}, we assess the sensitivity of our best weakly supervised model (\textit{WS-UmBERTo-TPT}) to the selection of clusters used to generate weak labels. In particular, we remove one first-level cluster at a time from the set of positive clusters and evaluate the resulting classification performance.

For the bronchiolitis dataset, excluding clusters \#5 or \#1 has little effect on performance, which is reasonable considering their smaller size. Removing clusters \#3 or \#4 leads to a moderate decrease in performance. In particular, cluster \#4 corresponds to the second bronchiolitis definition ("\textit{bronchospasm}" and "\textit{fever}"); excluding this cluster is therefore equivalent to omitting that definition in the keyword specification step. Only excluding cluster \#2 produces a substantial drop in performance and an increase in variance, which is expected since it accounts for almost half of the bronchiolitis cases.

A similar trend is observed for the bronchitis dataset. Removing cluster \#11 or \#6 produces limited changes in performance, whereas excluding clusters \#1 or \#4 leads to a larger decrease in classification performance. Overall, these results confirm the robustness of the model with respect to moderate changes in the selection of clusters used to generate weak labels and, more generally, to variations in the disease definitions used in the weak label construction step.

Additional analyses evaluating model performance when training and testing on weak labels, together with further stratified results across hospitals, LHUs, and pediatric vs non-pediatric emergency departments, are reported in \ref{app:classification}. Some hospitals are more represented in the dataset, leading to superior performance, but also hospitals with a similar number of records show considerable performance variability, suggesting that reporting practices in discharge letters might influence disease detection.

\section*{Discussion}
\label{sec:discussion}

This study introduces a novel pipeline demonstrating how weakly supervised learning can be effectively applied to clinical NLP. To our knowledge, this is the first work addressing diagnosis identification from Italian discharge letters. Although we primarily validated it using a case study on bronchiolitis, we also conducted a secondary validation on a smaller dataset for bronchitis. 

We observed a good mapping between diagnosis strings and their clusters' descriptions in the manually evaluated subset of strings, and indeed a good match between weak and gold labels, confirming that the SAL step can effectively identify relevant cases. Despite the inevitable noise introduced by weak labels, its magnitude remains limited, allowing the model to perform well when evaluated against gold labels. Broadly similar trends are observed in both datasets; however, results on the bronchitis dataset should be interpreted with caution given the limited number of positive cases. Additionally, the differences between settings where we retained the diagnosis strings and those where we removed them are limited. This indicates that the model is effectively learning from the entire text of the discharge letters, making it particularly advantageous when the automatic extraction of diagnosis strings fails. Our manual exploration of some weak labels further revealed that diagnosis strings do not always mention the disease responsible for hospitalization. In some cases, they only refer to symptoms, and in cases of multiple diseases, not all may be reported. These findings highlight the necessity for a classification model based on the complete text of discharge letters, as the one we developed.

In contrast, rule-based approaches underperformed because diseases like bronchiolitis and bronchitis are not always explicitly mentioned or may appear in negated or hypothetical contexts. While more sophisticated rule-based systems could address some of these issues, such approaches lack the scalability and generalizability of our pipeline. Conversely, our pipeline can be adapted to other diseases, making it a potentially versatile tool for clinical NLP. A similar issue arises when weak labels are generated using the XPRESS framework applied across the entire clinical text, which produces substantially noisier labels compared with the diagnosis-string-based weak labelling used in our pipeline.
The zero-shot LLM performed better than rule-based and some weakly supervised settings, but it did not surpass the best weakly supervised configuration. Despite the reasoning capacities of modern LLMs, their training data rarely include documents and tasks of this type, limiting performance.

Domain pretraining (TPT) improved results across all models. The best performance was obtained with UmBERTo, trained entirely on Italian data, confirming the advantage of language-specific representations even over larger multilingual models.

Compared with fully supervised classifiers, our weakly supervised pipeline shows a reduction in performance of 4–6\% AUPRC. While this difference is statistically significant, its practical relevance depends on the intended application. In the context of cohort selection from large collections of discharge letters, the goal is to identify groups of potentially relevant patients at scale, rather than to achieve perfect classification at the individual level. In this case, such a reduction may be considered acceptable given the substantial decrease in annotation effort, which would otherwise require more than 1,500 hours of expert time in our case study. This highlights the practical trade-off between performance and scalability offered by weakly supervised approaches.

The experiments in which we progressively replaced gold labels with weak labels (Figure~\ref{fig:auprc-mixedlabels}) further support this trade-off in a realistic scenario where only a small portion of the data can be manually annotated. Even in this mixed-label setting, performance degraded smoothly and remained close to the fully supervised baseline as long as a modest fraction of gold labels was available, indicating strong resilience to label noise. 

The pipeline has also shown robustness to the cluster selection used to generate weak labels across both case studies. Performance degradation was proportional to the reduction in weak label coverage, indicating stable behavior under moderate variations in disease definitions. Although second-level clustering provided no or only modest improvements in downstream performance in our case studies, it improved cluster coherence and weak label correctness in the independent evaluation of clustering quality (Figure~\ref{fig:labels-eval}.b), thereby contributing to more interpretable cluster structures.

It is important to clarify that, while the pipeline does not require document-level manual annotation, it relies on disease-specific keyword definitions provided as minimal domain input. This step requires collaboration with clinicians to reflect the terminology commonly used in discharge letters, but it is substantially less demanding than full annotation of individual documents.  As shown in the robustness analysis, moderate variations in cluster selection lead to proportionally limited changes in performance, suggesting that the approach is not overly sensitive to minor imperfections in keyword specification.

While our results are promising, several limitations should be acknowledged. 
First, the gold labels for both datasets were assigned primarily by a single annotator, introducing potential uncertainty in the reference standard. We performed an additional inter-annotator agreement analysis on a stratified subset of discharge letters, but this evaluation was limited in size and does not fully replace a systematic double-annotation process on the complete datasets.
Although the task consisted of binary disease identification, which is less complex than fine-grained clinical annotation, some degree of subjectivity cannot be excluded.
Second, a limitation concerns the additional validation on the bronchitis dataset. Due to the relatively small number of positive cases (97 in total), each test fold in the 5-fold cross-validation contains approximately 19 positive examples. This results in limited statistical power to detect differences between models and increases the variability of performance estimates. Consequently, results on the bronchitis dataset should be primarily viewed as indicative of generalization trends rather than as a definitive comparative evaluation.
Third, the multi-centric nature of our dataset, though a strength in its diversity, also presents an imbalance in the number of records from different hospitals and LHUs, which are also all located within a single Italian region. This is not only due to the geographical distribution of the population, but might also reflect some biases in the distribution of the family pediatricians adhering to the Pedianet network. Testing the pipeline on more balanced datasets and externally validating it on data from other regions would strengthen its generalizability and allow a more precise assessment of performance differences among hospitals.
Finally, the evaluation of intermediate steps relied on the manual review of a relatively small subset of diagnoses strings. Extending this annotation to a larger subset and assessing entire clusters would require substantial effort but could enable a more detailed analysis of the performance of each component in the pipeline.

Future research can extend this work along multiple directions.
Although we evaluated the pipeline on two different disease-specific datasets, further validation on additional clinical conditions and datasets is required to fully assess its generalizability. Future studies should also include complete double annotation to formally assess inter-annotator agreement and better quantify the reliability of reference labels. Incorporating hospital or LHU identifiers in hierarchical classification models could further improve performance by accounting for site-specific documentation practices.
Moreover, enhancing the interpretability of model predictions will be crucial for widespread clinical adoption. Identifying which portions of each discharge letter most influence the classifier's decision would help clinicians assess reliability and build trust. Techniques from explainable artificial intelligence could be applied to reveal the reasoning process behind model outputs and possibly recognize potential biases present in the training data.

These future developments would further strengthen both the methodological robustness and clinical utility of weakly supervised NLP systems for healthcare text, supporting scalable and trustworthy solutions for real-world applications.

\section*{Conclusions}
We presented a weakly supervised pipeline for disease identification from clinical discharge letters, designed to minimize the need for manual annotation while maintaining competitive classification performance. The proposed approach combines automatic extraction of diagnosis strings, clustering-based weak label generation, and transformer-based text classification.

In a case study on bronchiolitis using Italian discharge letters, the pipeline outperformed unsupervised rule-based approaches and achieved performance close to fully supervised models. Similar model rankings were observed in a secondary validation on a smaller bronchitis dataset. The results also highlight the importance of targeted weak label generation: simpler strategies, such as in the XPRESS-based baseline, produced substantially noisier labels and markedly worse performance.

Experiments with mixed gold and weak labels further showed that the proposed approach is resilient to label noise and can maintain strong performance even when only a limited portion of the dataset is manually annotated. In addition, the classification models were able to effectively learn from the entire discharge letter text, reducing dependence on perfectly extracted diagnosis strings.

By substantially reducing the annotation burden while maintaining robust performance, this approach offers a practical and scalable framework for applying NLP to large collections of clinical documents.

\section*{Declaration of competing interests}
The authors declare that they have no competing interests.

\section*{Data statement}
Data used in this study are not publicly available due to their confidential nature. Data are however available upon reasonable request to the corresponding author, subject to approval by the Internal Scientific Committee of Società Servizi Telematici Srl, the legal owner of Pedianet. 

The code developed for the analysis is available at \url{https://github.com/vittot/weakly-supervised-classification-italian-discharge-letters} .

\section*{CRediT authorship contribution statement}
\textbf{Vittorio Torri}: Methodology, Software, Investigation, Writing - Original Draft
\textbf{Elisa Barbieri}: Conceptualization, Data Curation, Writing - Review \& Editing
\textbf{Anna Cantarutti}: Conceptualization, Data Curation, Writing - Review \& Editing
\textbf{Carlo Giaquinto}: Conceptualization, Supervision
\textbf{Francesca Ieva}: Conceptualization, Methodology, Writing - Review \& Editing, Supervision

\section*{Acknowledgments}
The present research has been supported by MUR, grant "Dipartimento di Eccellenza 2023-2027". 

Francesca Ieva acknowledges the National Plan for NRRP Complementary Investments "Advanced Technologies for Human-centred Medicine" (PNC0000003). 

Elisa Barbieri declares that her position is funded within the project “INF-ACT - One Health Basic and translational Research Actions addressing Unmet Needs on Emerging Infectious Diseases”, ID MUR PE\_00000007, under the National Recovery and Resilience Plan (NRRP), Mission 4, Component 2 – CUP C93C22005170007.

The authors thank all the family paediatricians collaborating in Pedianet: Eva Alfieri, Michela Alfiero Bordigato, Angelo Alongi, Biagio Amoroso, Rosaria Ancarola, Barbara Andreola, Giampaolo Anese, Roberta Angelini, Maria Grazia Apostolo, Giovanna Argo, Giovanni Avarello, Lucia Azzoni, Maria Carolina Barbazza, Patrizia Barbieri, Gabriele Belluzzi, Eleonora Benetti, Filippo Biasci, Franca Boe, Stefano Bollettini, Francesco Bonaiuto, Anna Maria Bontempelli, Matteo Bonza, Sara Bozzetto, Andrea Bruna, Ivana Brusaterra, Massimo Caccini, Laura Calì, Sonia Camposilvan, Laura Cantalupi, Luigi Cantarutti, Chiara Cardarelli, Giovanna Carli, Sylvia Carnazza, Massimo Castaldo, Stefano Castelli, Monica Cavedagni, Giuseppe Egidio Cera, Chiara Chillemi, Francesca Cichello, Giuseppe Cicione, Carla Ciscato, Mariangela Clerici Schoeller, Samuele Cocchiola, Giuseppe Collacciani, Valeria Conte, Roberta Corro', Rosaria Costagliola, Nicola Costanzo, Sandra Cozzani, Giancarlo Cuboni, Giorgia Curia, Caterina D'alia, Vito Francesco D'Amanti, Antonio D'Avino, Roberto De Clara, Lorenzo De Giovanni, Annamaria De Marchi, Gigliola Del Ponte, Tiziana Di Giampietro, Giuseppe Di Mauro, Giuseppe Di Santo, Piero Di Saverio, Mattea Dieli, Marco Dolci, Mattia Doria, Dania El Mazloum, Maria Carmen Fadda, Pietro Falco, Mario Fama, Marco Faraci, Maria Immacolata Farina, Tania Favilli, Mariagrazia Federico, Michele Felice, Maurizio Ferraiuolo, Michele Ferretti, Mauro Gabriele Ferretti, Paolo Forcina, Patrizia Foti, Luisa Freo, Ezio Frison, Fabrizio Fusco, Giovanni Gallo, Roberto Gallo, Andrea Galvagno, Alberta Gentili, Pierfrancesco Gentilucci, Giuliana Giampaolo, Francesco Gianfredi, Isabella Giuseppin, Laura Gnesi, Costantino Gobbi, Renza Granzon, Mauro Grelloni, Mirco Grugnetti, Antonina Isca, Urania Elisabetta Lagrasta, Maria Rosaria Letta, Giuseppe Lietti, Cinzia Lista, Ricciardo Lucantonio, Francesco Luise, Enrico Marano, Francesca Marine, Lorenzo Mariniello, Gabriella Marostica, Sergio Masotti, Stefano Meneghetti, Massimo Milani, Stella Vittoria Milone, Donatella Moggia, Angela Maria Monteleone, Pierangela Mussinu, Anna Naccari, Immacolata Naso, Flavia Nicoloso, Cristina Novarini, Laura Maria Olimpi, Riccardo Ongaro, Maria Maddalena Palma, Angela Pasinato, Andrea Passarella, Pasquale Pazzola, Monica Perin, Danilo Perri, Silvana Rosa Pescosolido, Giovanni Petrazzuoli, Giuseppe Petrotto, Patrizia Picco, Ambrogina Pirola, Lorena Pisanello, Daniele Pittarello, Elena Porro, Antonino Puma, Maria Paola Puocci, Andrea Righetti, Rosaria Rizzari, Cristiano Rosafio, Paolo Rosas, Bruno Ruffato, Lucia Ruggieri, Annamaria Ruscitti, Annarita Russo, Pietro Salamone, Daniela Sambugaro, Luigi Saretta, Vittoria Sarno, Valentina Savio, Nico Maria Sciolla, Rossella Semenzato, Paolo Senesi, Carla Silvan, Giorgia Soldà, Valter Spanevello, Sabrina Spedale, Francesco Speranza, Sara Stefani, Francesco Storelli, Paolo Tambaro, Giacomo Toffol, Gabriele Tonelli, Silvia Tulone, Angelo Giuseppe Tummarello, Cristina Vallongo, Sergio Venditti, Maria Grazia Vitale, Concetta Volpe, Francescopaolo Volpe, Aldo Vozzi, Giulia Zanon, Maria Luisa Zuccolo.

 \bibliographystyle{elsarticle-num} 
 \bibliography{cas-refs}

 \renewcommand\thefigure{\thesection.\arabic{figure}}    
\renewcommand{\thetable}{\thesection.\arabic{table}}

\clearpage
\appendix

\section{Dataset}
\label{sec:data-appendix}
\setcounter{figure}{0}  
\setcounter{table}{0}
\subsection{Records distributions among hospitals and LHUs}
\label{app:data-plots}
\begin{figure}[h!]
    \centering    \includegraphics[width=.6\textwidth]{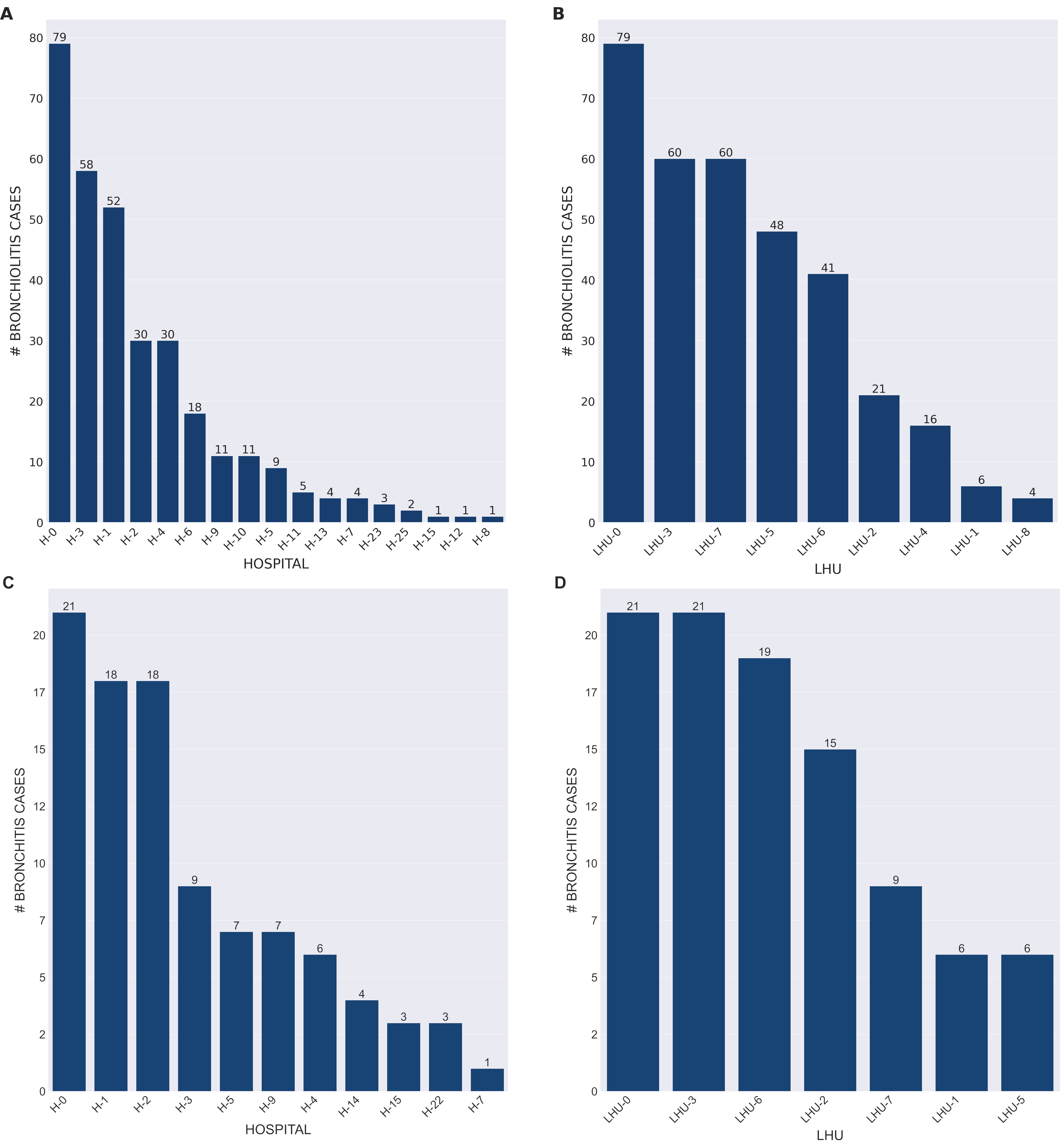}
    \caption{Distribution of cases among hospitals (A,C) and LHUs (B,D) for the bronchiolitis (A,B) and bronchitis datasets (C,D).}
    \label{fig:bronch_dist}
\end{figure}

\begin{figure}[htp]
    \centering    \includegraphics[width=\textwidth]{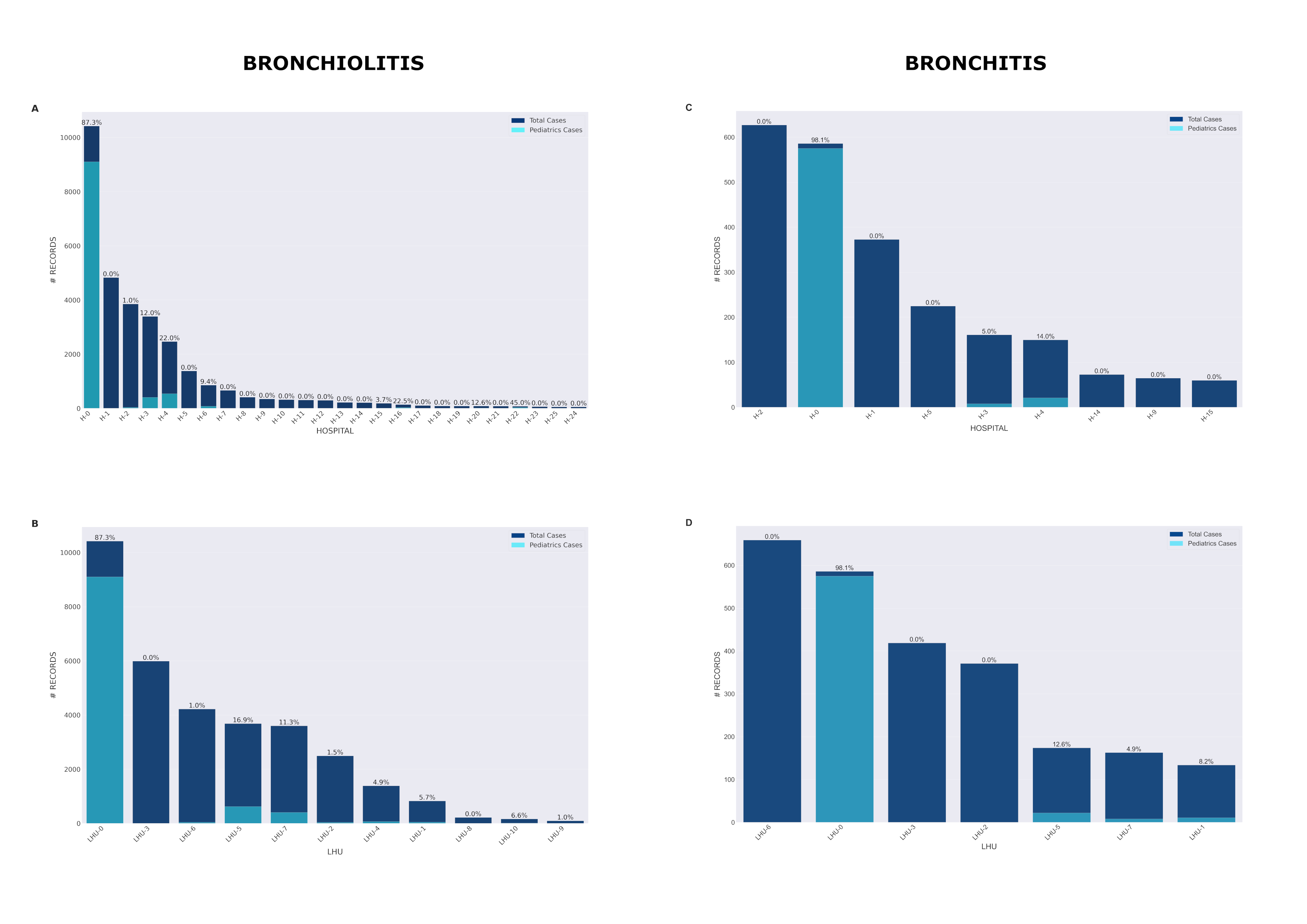}
    \caption{Distribution of records among hospitals (only hospitals with $> 50$ records) (A,C) and distribution of records among LHUs (only LHUs with $> 50$ cases) (B,D) for the bronchiolitis (A,B) and bronchitis (C,D) datasets. For each hospital/LHU, the blue bar indicates the total number of records, the light blue bar the records from pediatric ERs/departments and the value on top of the bars is the \% of pediatric records in the hospital/LHU.}
    \label{fig:dist}
\end{figure}

Figures \ref{fig:bronch_dist} and \ref{fig:dist} show the distribution of records across hospitals and LHUs, highlighting the number of positive cases and the percentage of records from pediatric ERs/departments.

\break

\subsection{Discharge letters example}
\label{app:letter-example}
An example of a discharge letter is reported below. This letter is a synthetic example similar in the type of content to those present in the analyzed dataset. We report it in both Italian and English.

\begin{lstlisting}[
    basicstyle=\tiny, %or \small or \footnotesize etc.
]
    
                               SERVIZIO SANITARIO NAZIONALE - REGIONE VENETO
                                            Azienda ULSS n.
                                      - 
                                           Presidio Ospedaliero di 
                                   Unita' Operativa di  Pediatria  Degenze
       Data: 
       Egregio Collega,
       si dimette in data odierna:   nato/a il 
       Data e ora di ricovero:  
       Diagnosi:
        INSUFFICIENZA RESPIRATORIA IN BRONCOSPASMO
       Decorso Clinico e Conclusioni:
        E.O. all'ingresso: condizioni generali sufficienti,  Sat.O2 = 96%,   FC = 154/m'. F.R.= 50/m', P.A. = 100/60. Kg = 13,250, T = 36,3C
        Paziente pallido, polipnoica, al torace ridotta penetrazione d'aria, con ronchi e gemiti espiratori sparsi. Azione cardiaca valida e
        ritmica, addome piano, trattabile, non dolente ne' dolorabile sia alla palpazione superficiale che profonda in tutti i  quadranti, organi
        ipocondriaci nei limiti, peristalsi vivace. Faringe leggermente iperemico. Es. neurologico negativo. Non segni meningei.
        Sono stati eseguiti i seguenti esami, risultati negativi:
        Es. emocromocitometrico, azotemia, creatininemi, elettroliti, transaminasi, PCR, emogasanalisi.
        E' stato riscontrato un modesto aumento della glicemia, dovuta allasomministrazione di cortisone.
        E' stata iniziata terapia con O2 e con aerosol con broncovaleas 8 gtt + sol. fisiologica una somministrazione ogni 30 m' per 3 volte,
        seguita da somministrazione dello stesso prodotto ( 8 gtt) 8  volte al di' (ogni 3 h), con buona risposta da parte della sintomatologia
        respiratoria.  Alla terapia aerosolica e' stata associata terapia cortisonica per os con bentelan, 1,0 mg 2 volte al di'.
                                                             i
       Programma alla dimissione:
        A domicilio   si   consiglia:
        - Aerosol con broncovaleas 6 gtt + sol. fisiol. 3 - 4 volte al di' per 3 - 5 giorni.
        - Lavaggi nasali con sol. fisiol. sei volte al di' per 3 - 5 giorni.
        - Bentelan 2 cp da 0,5 mg alla sera, a stomaco pieno, per 2 giorni.
       Consigli Clinici:
        Si consiglia inoltre controllo presso il pediatra curante nei prossimi giorni.
         A  disposizione per ulteriori chiarimenti, inviamo cordiali saluti
       Cordiali saluti
       Dr. 
       PED1D Pediatria  Degenze
       INFORMAZIONE
       Gentile signore/ signora, desideriamo renderla partecipe che per il suo percorso di cura e stato stimato un impiego di risorse
       economiche da parte del Servizio Sanitario Regionale pari a Euro 1.945,69
       Paziente:   ( ID:  )
       Episodio N.: /01 ( Unita' Operativa di PED1D Pediatria  Degenze )
       stampato il                                                                          Pag.1 of 1
                                             
                                                                                   Referto firmato digitalmente.
\end{lstlisting}

\begin{lstlisting}[
    basicstyle=\tiny, %or \small or \footnotesize etc.
]
                               NATIONAL HEALTH SERVICE - VENETO REGION 
                                            ULSS Company No.
                                      - 
                                           Hospital of 
                                   Pediatric Unit Hospitalization
       Date: 
       Dear Colleague,
       discharged today:   born on 
       Date and time of admission:  
       Diagnosis:
        RESPIRATORY FAILURE IN BRONCHOSPASM
       Clinical Course and Conclusions:
        Upon admission: general condition sufficient, Sat.O2 = 96%,   HR = 154/m', RR = 50/m', BP = 100/60. Weight = 13,250 kg, T = 36.3C
        Patient pale, tachypneic, with reduced air entry in the chest, with scattered wheezing and expiratory moaning. Valid and rhythmic
        cardiac action, flat abdomen, treatable, not painful or tender on both superficial and deep palpation in all quadrants, hypochondriac
        organs within limits, active peristalsis. Pharynx slightly hyperemic. Neurological exam negative. No meningeal signs.
        The following tests were performed, all negative:
        Blood count, urea, creatinine, electrolytes, transaminases, CRP, blood gas analysis.
        A slight increase in blood glucose was found, due to the administration of cortisone.
        Therapy was started with O2 and aerosol with broncovaleas 8 drops + saline solution, one administration every 30 minutes for 3 times,
        followed by administration of the same product (8 drops) 8 times a day (every 3 hours), with good response from respiratory symptoms.
        Aerosol therapy was combined with oral cortisone therapy with bentelan, 1.0 mg twice a day.
                                                             i
       Discharge Plan:
        At home, it is advised:
        - Aerosol with broncovaleas 6 drops + saline solution 3 - 4 times a day for 3 - 5 days.
        - Nasal washes with saline solution six times a day for 3 - 5 days.
        - Bentelan 2 tablets of 0.5 mg in the evening, on a full stomach, for 2 days.
       Clinical Recommendations:
        It is also advised to check with the primary care pediatrician in the coming days.
         Available for further clarification, we send our best regards.
       Best regards,
       Dr. 
       PED1D Pediatric Hospitalization
       INFORMATION
       Dear sir/madam, we wish to inform you that an estimated expenditure of resources by the Regional Health Service for your care is
       Euro 1,945.69.
       Patient:   ( ID:  )
       Episode No.: /01 ( Pediatric Unit of PED1D Hospitalization )
       printed on                                                                          Page 1 of 1
                                             
                                                                                   Digitally signed report.
\end{lstlisting}

\subsection{E3C Corpus}
\label{app:e3c}
Table \ref{tab:e3c} reports the details about the composition of the E3C corpus.

\begin{table}[h!]
\caption{Number of documents in the E3C corpus for the different languages and types of sources}
\label{tab:e3c}
\begin{tabular}{llllll}
\hline
\multicolumn{1}{c}{\textbf{Source\textbackslash{}Language}} & \textbf{English} & \textbf{French} & \textbf{Italian} & \textbf{Spanish} & \textbf{Basque} \\
\hline
\textit{pubmed}                                             & 9318             & 132             & 0                & 0                & 0               \\
\textit{journal}                                            & 716              & 1637            & 798              & 504              & 201             \\
\textit{SPACCC corpus}                                      & 0                & 0               & 0                & 1645             & 0               \\
\textit{other}                                              & 0                & 24220           & 9675             & 0                & 1232 \\
\hline
\end{tabular}
\end{table}

\section{Methods notation summary}
\setcounter{table}{0}
\setcounter{figure}{0}
\label{app:B}
Table~\ref{tab:notation} reports a summary of the main notation used throughout the Methods sections of the paper.
\begin{table}[p]
\centering
\small
\caption{Summary of the main notation used throughout the Methods section of the paper.}
\label{tab:notation}
\begin{tabular}{ll}
\hline
\textbf{Notation} & \textbf{Description} \\
\hline
$\mathcal{X}$ & Space of textual documents \\
$x_i$ & Single discharge letter/document \\
$\mathcal{D}$ & Set of discharge letters \\
$\mathcal{C}$ & Full E3C corpus \\
$\mathcal{C}_{\ell}$ & Subset of E3C documents in language $\ell$ \\
$T_\ell(\cdot)$ & Translation operator from language $\ell$ to Italian \\
$\mathcal{C}^{it}$ & Final Italian medical corpus used for TPT \\
$f_{\theta}(\cdot)$ & Transformer-based text encoder \\
$f_{\hat{\theta}}(\cdot)$ & Domain-adapted transformer encoder after TPT \\
$\mathcal{L}_{\text{MLM}}$ & Masked language modeling loss \\
$\mathcal{S}$ & Space of diagnosis strings \\
$s_i$ & Diagnosis string extracted from $x_i$ \\
$\bar{s}_i$ & Normalized diagnosis string \\
$g(\cdot)$ & Diagnosis string extraction function \\
$p_s(\cdot)$ & Diagnosis string preprocessing function \\
$z_i$ & Embedding of diagnosis string $\bar{s}_i$ \\
$\tilde{z}_i$ & PCA-projected embedding of $z_i$ \\
$W$ & PCA projection matrix \\
$I$ & Documents with extracted diagnosis strings \\
$h(\cdot)$ & HDBSCAN cluster assignment function \\
$C_k$ & Cluster of diagnosis strings \\
$\mathrm{Keywords}(C_k)$ & Representative keywords of cluster $C_k$ \\
$u_k$ & Embedding of cluster keyword representation \\
$I_c$ & Documents assigned to a cluster \\
$P_d$ & Set of positive keywords in disease definition $d$ \\
$N_d$ & Set of negative keywords in disease definition $d$ \\
$L$ & Set of clusters selected for a disease \\
$\tilde{y}_i$ & Weak label associated with document $x_i$ \\
$\tilde{\mathcal{D}}$ & Set of weakly labelled documents \\
$p(\cdot)$ & Document preprocessing function \\
$\bar{x}_i$ & Preprocessed discharge letter \\
$\tau_{512}(\cdot)$ & Truncation operator to 512 tokens \\
$x_i^{*}$ & Final classifier input document \\
$\hat{p}_i$ & Predicted probability of the disease \\
$\mathcal{L}_{\text{WLC}}$ & Weak label classification loss \\
\hline
\end{tabular}
\end{table}

\break

\section{Pre-processing}
\label{app:preproc}
\setcounter{table}{0}
\setcounter{figure}{0}
From diagnosis strings we remove punctuation, dates and times, and we trim new lines and spaces in excess.

To remove headers and footers from the text of whole letters we report below the strings used as identifiers of sentences to be removed.
\begin{lstlisting}
regione
azienda
dipartimento*
pronto soccorso pediatrico*
u.o.c*
UOC *
operativa*
accettazione*
u.o*
UO *
uo.*
U.L.S.S.*
ULSS*
ospedale*
presidio ospedaliero*
via*
punto di primo intervento
pediatria*
verbale*
pag.
pag 
pagina
nato*
tess.san
codice fiscale
comune di nascita
cap
indirizzo
cartella dea
documento firmato digitalmente
informazione ai sensi
il medico dimettente
gentile signor
copia di documento firmato e conservato
desideriamo renderLa partecipe
modulo di pronto soccorso
direttore*
ai genitori
al medico
indirizzo
residente
residenza
nome
cognome
firma
consegnare al proprio pediatra
l'orario di alcune prestazioni
verbale di pronto soccorso
della cartella*
modulo di
numero di certificato
firmatario
il referto e\ conservato
ID Documento
gentile signore
informazione
dettagli paziente
verbale n
cartella DEA
priorita*
tel
fax
indirizzo
residenza
domicilio
segreteria
data e ora
\end{lstlisting}
Those with * must be removed only when they are strictly at the beginning of the document. Otherwise, they might not necessarily indicate a header sentence.

\begin{figure}[H]
    \centering
    \includegraphics[width=\linewidth]{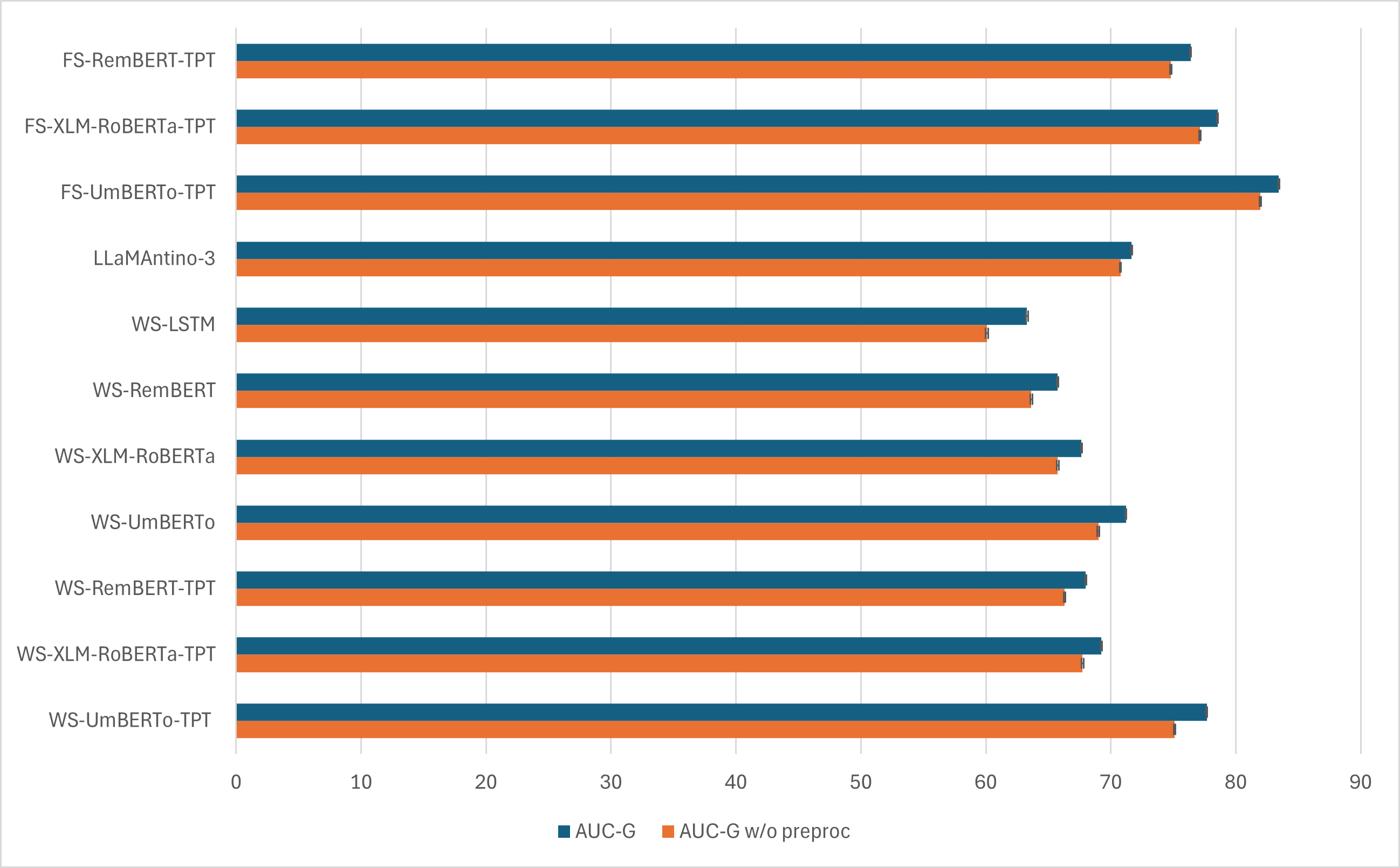}
    \caption{AUC ($\pm 95\%$ C.I.) on gold labels with and without pre-processing}
    \label{fig:abl-preproc}
\end{figure}

Figure~\ref{fig:abl-preproc} reports the difference in results with and without this preprocessing.

\section{Diagnosis extraction}
\label{app:diagnosis}
\setcounter{table}{0}
To extract diagnosis strings, we initially consider the following keywords to be looked for at the beginning of a sentence
\begin{lstlisting}
diagnosi
diagnosi di dimissione
diagnosi alla dimissione
diagnosi testuale
\end{lstlisting}
We initially take as a diagnosis string the sentence that follows the keywords we identified among the ones above. It can be on the same line, if present, or on the following line.
Then, this candidate diagnosis string is cleaned by removing the portion(s) related to the following expressions:
\begin{lstlisting}
decorso clinico
consigli terapeutici
consiglio
controllo
a domicilio
paziente: ( id: \d+)
\end{lstlisting}
These keywords are followed by indications related to therapies to be followed after discharge or additional details of the hospital stay, which are not strictly part of the diagnosis string.
In particular, if we identify these keywords, we remove the portion of the sentence that starts from them up to the next major punctuation sign (dot or semicolon). The remaining portion, if any, is extracted as a diagnosis string. 

\noindent Table \ref{tab:diagnosis-extracted} reports the most frequently extracted diagnosis sentences. 

\begin{table}[!htbp]
\caption{Most common diagnosis sentences extracted from discharge letters \\ (frequency $>$100)}
\centering
\label{tab:diagnosis-extracted}
\resizebox{.85\textwidth}{!}{
\begin{tabular}{p{6cm}p{6cm}r}
\hline
\textbf{Extracted Diagnosis {[}ITA{]}}                                 & \textbf{Extracted Diagnosis {[}ENG{]}}                                    & \textbf{N. Occ.} \\
\hline
virosi                                                                 & virosis                                                                   & 598              \\
079.99   infezioni virali,non specificate                              & 079.99 viral infections,unspecified                                       & 351              \\
nato   singolo,nato in ospedale senza menzione di taglio cesareo       & single-born child,born in hospital without mention of cesarean section          & 324              \\
gastroenterite                                                         & gastroenteritis                                                           & 323              \\
trauma cranico minore                                                  & minor head injury                                                         & 212              \\
febbre                                                                 & fever                                                                     & 190              \\
trauma cranico non commotivo                                           & non-concussive head injury                                                & 174              \\
dolore addominale                                                      & abdominal pain                                                            & 170              \\
780.6 febbre                                                           & 780.6 fever                                                               & 139              \\
orticaria                                                              & hives                                                                     & 135              \\
broncospasmo                                                           & bronchospasm                                                              & 117              \\
519.11   broncospasmo acuto                                            & 519.11 acute bronchospasm                                                 & 116              \\
febbre di recente insorgenza                                           & fever of recent onset                                                     & 112              \\
gastroenterite acuta                                                   & acute gastroenteritis                                                     & 109              \\
otite                                                                  & otitis                                                                    & 106              \\
trauma cranico                                                         & head injury                                                               & 106              \\
009.1   colite,enterite e gastroenterite di presunta origine infettiva & 009.1 colitis,enteritis and gastroenteritis of presumed infectious origin & 104              \\
vomito                                                                 & vomit                                                                     & 104              \\
flogosi delle alte vie respiratorie                                    & upper respiratory tract inflammation                                      & 102              \\
bronchiolite                                                           & bronchiolitis                                                             & 102             \\
\hline
\end{tabular}
}
\end{table}

\section{Clustering}
\setcounter{table}{0}
\setcounter{figure}{0}
\label{app:clustering}
\subsection{Clustering hyperparameters}

\begin{figure}
    \centering
    \includegraphics[height=.86\textheight]{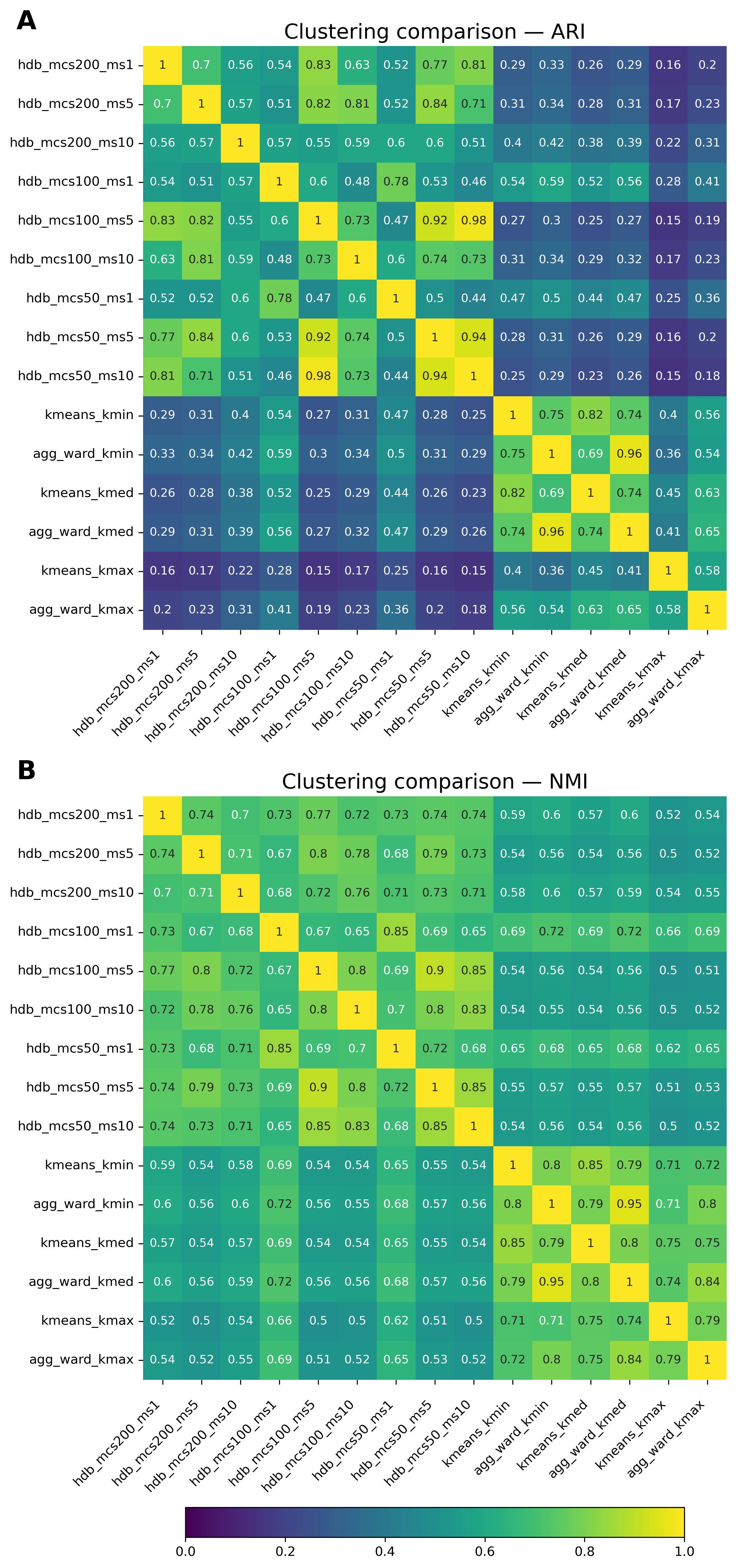}
    \caption{Adjusted Rand Index (ARI) (A) and Normalized Mutual Information (NMI) (B) among clusters obtained with different algorithms (HDBSCAN, K-Means, Agglomerative Hierarchical Clustering), with different hyperparameters. For HDBSCAN \textit{mcs} = minimum cluster size, \textit{ms}=minimum number of samples. For K-Means and Agglomerative Hierarchical Clustering, \textit{kmin}, \textit{kmed} and \textit{kmax} use as number of clusters the minimum, median or maximum number of clusters found by HDBSCAN, respectively.}
    \label{fig:ari_nmi}
\end{figure}

\begin{figure}
    \centering
    \includegraphics[width=\textwidth]{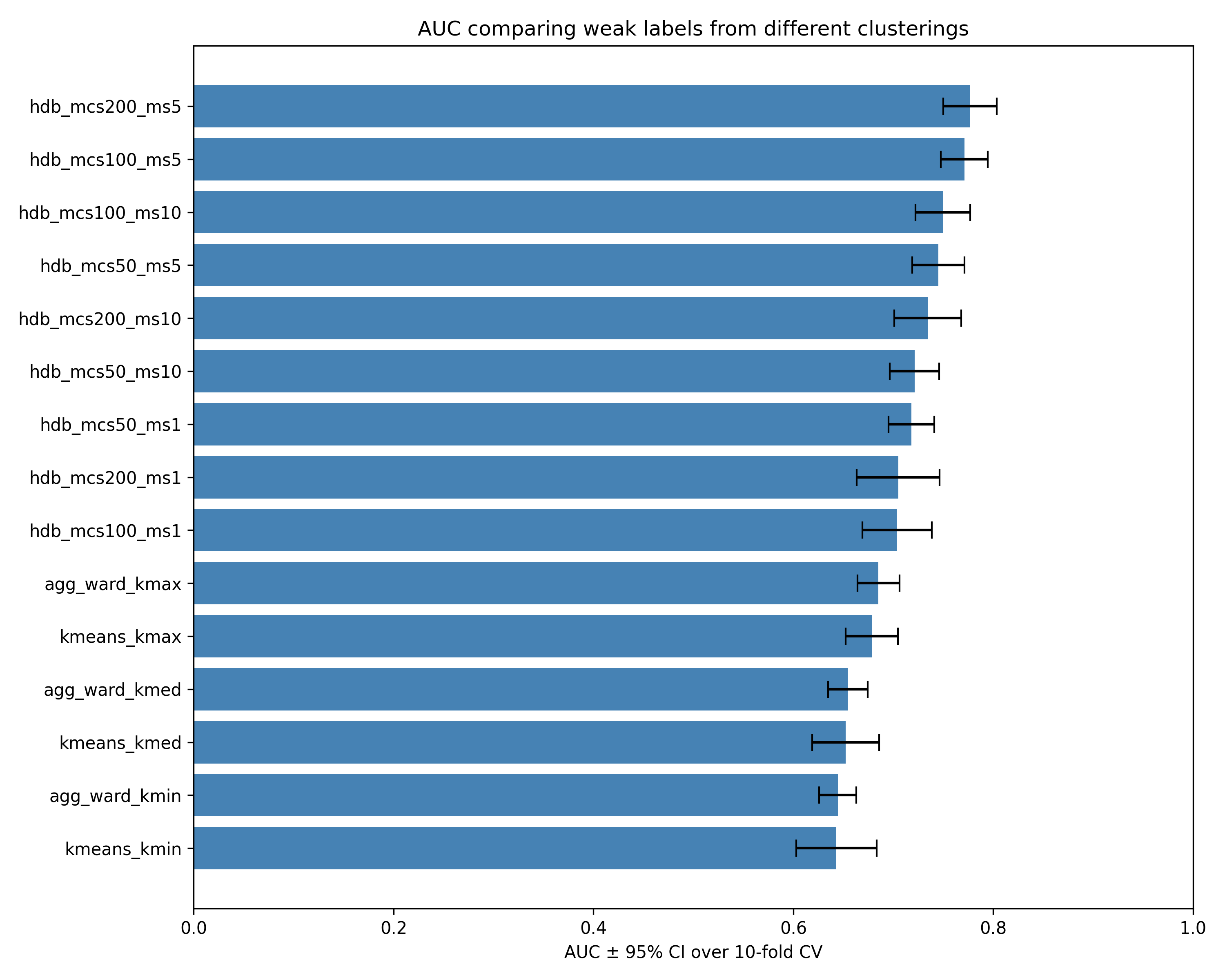}
    \caption{AUROC ($\pm 95\%$ C.I.) obtained in 10-fold cross validation using weak labels derived by applying different clustering algorithms (HDBSCAN, K-Means, Agglomerative Hierarchical Clustering), with different hyperparameters. For HDBSCAN \textit{mcs} = minimum cluster size, \textit{ms}=minimum number of samples. For K-Means and Agglomerative Hierarchical Clustering, \textit{kmin}, \textit{kmed} and \textit{kmax} use as number of clusters the minimum, median or maximum number of clusters found by HDBSCAN, respectively.}
    \label{fig:auc_cluster_comp}
\end{figure}

To assess the robustness of the weak label generation process, we compared clusterings obtained using multiple HDBSCAN, k-means, and agglomerative hierarchical clustering configurations, on the bronchiolitis dataset. As shown in Figure~\ref{fig:ari_nmi}, HDBSCAN solutions exhibited very high internal consistency across parameter settings (ARI and NMI often $>0.7$), indicating that the underlying structure captured by the algorithm is stable. In contrast, clusterings from k-means and agglomerative hierarchical clustering formed separate blocks with substantially lower agreement with HDBSCAN (ARI 0.2–0.4), suggesting that these methods partition the data according to different inductive biases. We then evaluated the downstream utility of the derived weak labels in a supervised classification task (Figure~\ref{fig:auc_cluster_comp}). Weak labels obtained from HDBSCAN consistently yielded the highest AUROC values, whereas those from k-means and Ward resulted in lower performance. Although HDBSCAN configurations displayed broadly similar performance, some parameter choices led to noticeably larger uncertainty in downstream AUROC estimates. This suggests that larger minimum cluster sizes and intermediate values of min\_samples provide more stable weak labels and should be preferred as a starting point.

\subsection{Clusters representation}
\begin{algorithm}[htbp]
\caption{Keyword extraction for cluster representation with global-frequency filtering}
\label{alg:cluster-keywords}
\begin{algorithmic}[1]
\REQUIRE Diagnosis strings $\{s_i\}_{i \in \mathcal{I}}$, cluster assignment $h(i)\in\{1,\dots,K\}\cup\{-1\}$, stopword set $\mathcal{S}_{stop}$, tokenization function $\texttt{Tok}(\cdot)$
\REQUIRE Hyperparameters: initial threshold $\tau_0$, minimum keywords $m$, lower bound $\tau_{\min}$, global-frequency cutoff $\gamma$
\ENSURE Keyword sets $\{\mathrm{Keywords}(C_k)\}_{k=1}^K$

\vspace{0.2em}
\STATE \textbf{// Precompute global token frequencies across all extracted diagnosis strings}
\STATE $\mathcal{W} \leftarrow \emptyset$ \COMMENT{multiset of all tokens}
\FORALL{$i \in \mathcal{I}$}
    \STATE $W_i \leftarrow \texttt{Tok}(\texttt{lower}(s_i))$
    \STATE $W_i \leftarrow \{ w \in W_i : w \notin \mathcal{S}_{stop} \}$
    \STATE $\mathcal{W} \leftarrow \mathcal{W} \uplus W_i$ \COMMENT{$\uplus$ is multiset union}
\ENDFOR
\STATE $\mathrm{G}(w) \leftarrow \#(w \in \mathcal{W}) / |\mathcal{W}|$ \COMMENT{global relative frequency}

\vspace{0.2em}
\STATE \textbf{// Extract cluster keywords}
\FOR{$k = 1$ \TO $K$}
    \STATE $C_k \leftarrow \{ i \in \mathcal{I} : h(i) = k \}$
    \IF{$C_k = \emptyset$}
        \STATE $\mathrm{Keywords}(C_k) \leftarrow \emptyset$
        \STATE \textbf{continue}
    \ENDIF

    \STATE $\mathcal{W}_k \leftarrow \emptyset$
    \FORALL{$i \in C_k$}
        \STATE $W_i \leftarrow \texttt{Tok}(\texttt{lower}(s_i))$
        \STATE $W_i \leftarrow \{ w \in W_i : w \notin \mathcal{S}_{stop} \}$
        \STATE $W_i \leftarrow \{ w \in W_i : \mathrm{G}(w) \le \gamma \}$ \COMMENT{global-frequency filter}
        \STATE $\mathcal{W}_k \leftarrow \mathcal{W}_k \uplus W_i$
    \ENDFOR

    \IF{$|\mathcal{W}_k| = 0$}
        \STATE $\mathrm{Keywords}(C_k) \leftarrow \emptyset$
        \STATE \textbf{continue}
    \ENDIF

    \STATE $f_k(w) \leftarrow \#(w \in \mathcal{W}_k) / |\mathcal{W}_k|$ \COMMENT{within-cluster relative frequency}
    \STATE $\tau \leftarrow \tau_0$
    \STATE $K_k \leftarrow \emptyset$
    \WHILE{$|K_k| < m$ \AND $\tau \ge \tau_{\min}$}
        \STATE $K_k \leftarrow \{ w : f_k(w) > \tau \}$
        \STATE $\tau \leftarrow \tau / 2$
    \ENDWHILE
    \STATE $\mathrm{Keywords}(C_k) \leftarrow K_k$
\ENDFOR

\RETURN $\{\mathrm{Keywords}(C_k)\}_{k=1}^K$
\end{algorithmic}
\end{algorithm}

\noindent Algorithm~\ref{alg:cluster-keywords} describes the algorithm for keyword representation of clusters. In our implementation we used $\tau_0=0.8$, $m=2$, $\tau_{\min}=0.001$, and selected $\gamma=0.20$ to remove highly generic tokens (e.g., administrative words or very common clinical function words not fully captured by stopwords).

\subsection{Cluster with other embeddings}
\label{sec:cluster-umberto}

Another additional comparison that we report here is obtained by varying the embedding model. In particular, we experiment using the embeddings produced by the original UmBERTo model. This leads to 488 clusters at the first level and 119 at the second level. Tables \ref{tab:clusters} and \ref{tab:clusters-umb} describe the twenty largest clusters identified with the UmBERTo-TPT and the original UmBERTo, respectively. Table~\ref{tab:clusters-umb-resp} reports the main first-level clusters related to respiratory infections and Table~\ref{tab:clusters-umb-resp-2liv} the corresponding second-level clusters, using the original UmBERTo model.

\begin{table}[h!]
\caption{Keyword-based representation of the twenty largest first-level diagnosis clusters obtained with Umberto-TPT}
\label{tab:clusters}
\begin{tabular}{lll}
\toprule
\textbf{Cluster size} & \textbf{Cluster keywords {[}ITA{]}} & \textbf{Cluster keywords {[}ENG{]}} \\
\midrule
270                   & flogosi vie                         & phlogosis airways                   \\
145                   & trauma cranico                      & head injury                         \\
143                   & contusione dito mano                & contusion finger hand               \\
116                   & gastroenterite verosimile           & probable gastroenteritis            \\
114                   & episodio febbre                     & fever episode                       \\
94                    & vie infezione                       & tract infection                     \\
93                    & pronazione dolorosa                 & painful pronation                   \\
78                    & ferita labbro                       & lip wound                           \\
75                    & otite media acuta                   & acute otitis media                  \\
70                    & ferita lacero-contusa               & lacerocontused wound                \\
66                    & trauma contusivo                    & blunt trauma                        \\
66                    & broncospasmo flogosi vie            & bronchospasm pathway inflammation   \\
64                    & puntura insetto                     & insect bite                         \\
62                    & ferita lacero-contusa               & lacerocontused wound                \\
59                    & trauma dx                           & right-handed trauma                 \\
58                    & ustione grado                       & burn degree                         \\
54                    & frattura ferita                     & wound fracture                      \\
54                    & ferita lacero contusa destra        & right-handed lacerocontused wound   \\
53                    & insorgenza recente febbre           & recent onset fever                  \\
53                    & contusione piede destro dito        & right foot toe contusion \\
\bottomrule
\end{tabular}
\end{table}

\begin{table}[h!]
\caption{Keyword-based representation of the twenty largest first-level diagnosis clusters from the original Umberto representation}
\label{tab:clusters-umb}
\begin{tabular}{rp{5cm}p{5cm}}
\hline
\multicolumn{1}{l}{\textbf{Cluster size}} & \textbf{Cluster keywords [ITA]}                      & \textbf{Cluster keywords ENG}                    \\ \hline
615                                       & trauma cranico                                       & head injury                                            \\
498                                       & nato taglio                                              & born section                                           \\
405                                       & diagnosi reazione locale vaccino                         & diagnosis specific vaccine reaction                       \\
368                                       & infezioni specificate                                    & specified infections                                   \\
333                                       & ansia cellulite diarrea febbre frattura influenza vomito & anxiety cellulite diarrhea fever fracture flu vomiting \\
238                                       & aeree flogosi vie                                        & flogosis airways                                       \\
232                                       & congiuntivite specificata                                & painful pronation                                      \\
226                                       & flogosi vie                                              & flogosis airways                                       \\
206                                       & appendicite congiuntivite enterite rinite                & appendicitis conjunctivitis enteritis rhinitis         \\
199                                       & allergica orticaria pregressa                            & previous allergic orticaria                        \\
195                                       & coxalgia otalgia                                         & coxalgia otalgia                                       \\
178                                       & broncospasmo episodio                                    & bronchospasm episode                                   \\
178                                       & broncospasmo laringospasmo                               & bronchospasm laryngospasm                              \\
170                                       & d insetto puntura                                        & insect bite                                            \\
148                                       & acuta gastroenterite                                     & acute gastroenteritis                                  \\
144                                       & alte infezione respiratorie vie                          & high airways infection                                 \\
140                                       & enterite gastroenterite                                  & enteritis gastroenteritis                              \\
136                                       & faringotonsillite virale                                 & viral pharyngotonsillitis                              \\
135                                       & insorgenza recente                                       & recent onset                                           \\
133                                       & acuta bronchiolite bronchite                             & acute bronchiolitis bronchitis                         \\ \hline
\end{tabular}%
\end{table}

\begin{landscape}
\begin{table}[]
\caption{Keyword-based representation of the first-level clusters related to respiratory
infections, with the \% of bronchiolitis cases in each cluster and the \% to the total number
of cases in the dataset from the original Umberto representation}
\label{tab:clusters-umb-resp}
\scriptsize
\begin{tabular}{p{1cm}p{6cm}p{6cm}rr}
\hline
\textbf{Cluster \#} & \textbf{Cluster keywords [ITA]}                                                                          & \textbf{Cluster keywords [ENG]}                                                           & \textbf{Within cluster bronch. \%} & \textbf{\begin{tabular}[t]{@{}r@{}}\% of total bronch. \\ covered\end{tabular}} \\ \hline 
1                   & acuta bronchiolite lieve severa                                                                                      & acute mild severe bronchiolitis                                                                       & 87.76                              & 12.84                                                                           \\
2                   & acuta bronchiolite bronchite                                                                                         & acute bronchiolitis bronchitis                                                                        & 72.93                              & 28.96                                                                           \\
3                   & bronchiolite respiratorio virus                                                                                      & bronchiolitis respiratory virus                                                                       & 66.66                              & 2.99                                                                            \\
4                   & bronchiolite positiva vrs                                                                                            & bronchiolitis rsv positive                                                                            & 44.44                              & 1.19                                                                            \\
5                   & bronchiolite complicata corso ex gastroenterite paziente                                                             & bronchiolitis complicated course former gastroenteritis patient                                       & 40.00                              & 0.60                                                                            \\
6                   & bronchite broncospasmo lieve                                                                                         & bronchitis mild bronchospasm                                                                          & 20.00                              & 0.30                                                                            \\
7                   & bronchiolite broncospasmo lieve                                                                                      & bronchiolitis mild bronchospasm                                                                       & 16.67                              & 0.30                                                                            \\
8                   & allarme bronchiolite caratteristiche d epistassi insetto locale lussazione medicazione pregressa reazione sdr sn vrs & bronchiolitis alarm features epistaxis local insect dislocation dressing previous reaction sdr sn rsv & 16.67                              & 0.30                                                                            \\
9                   & acuta insufficienza respiratoria                                                                                     & acute respiratory failure                                                                             & 14.04                              & 2.39                                                                            \\
10                  & broncospasmo respiratoria                                                                                            & bronchospasm respiratory                                                                              & 12.50                              & 2.99                                                                            \\
11                  & broncospasmo laringospasmo                                                                                           & bronchospasm laryngospasm                                                                             & 1.12                               & 0.60                                                                            \\
12                  & broncospasmo episodio                                                                                                & bronchospasm episode                                                                                  & 0.56                               & 0.30                                                                            \\
13                  & febbrile respiratoria                                                                                                & respiratory fever                                                                                     & 0.0                                & 0.0                                                                             \\
14                  & febbre recente respiratorie vie                                                                                      & recent fever airways                                                                                  & 0.0                                & 0.0                                                                             \\
15                  & corso infezione respiratoria                                                                                         & ongoing respiratory infection                                                                         & 0.0                                & 0.0                                                                             \\
16                  & broncospasmo vie                                                                                                     & bronchospasm airways                                                                                 & 0.0                                & 0.0                                                                             \\
17                  & broncospasmo paziente ricorrente                                                                                     & bronchospasm recurrent patient                                                                        & 0.0                                & 0.0                                                                             \\
18                  & broncospasmo parainfettivo                                                                                           & parainfective bronchospasm                                                                            & 0.0                                & 0.0                                                                             \\
19                  & broncospasmo noto                                                                                                    & known bronchospasm                                                                                    & 0.0                                & 0.0                                                                             \\
20                  & broncospasmo infettivo parainfettivo                                                                                 & parainfective infectious bronchospasm                                                                 & 0.0                                & 0.0                                                                             \\
21                  & broncospasmo corso virosi                                                                                            & bronchospasm ongoing virosis                                                                         & 0.0                                & 0.0                                                                             \\
22                  & broncospasmo corso respiratorie vie                                                                                  & bronchospasm ongoing airways                                                                           & 0.0                                & 0.0                                                                             \\
23                  & broncospasmo corso infezione respiratoria                                                                            & bronchospasm ongoing respiratory infection                                                            & 0.0                                & 0.0                                                                             \\
24                  & broncospasmo corso flogosi vie                                                                                       & bronchospasm ongoing flogosis airways                                                                 & 0.0                                & 0.0                                                                             \\
25                  & broncospasmo corso flogosi lieve respiratoria                                                                        & bronchospasm ongoing mild flososis respiratory                                                        & 0.0                                & 0.0                                                                             \\
26                  & broncopolmonite destra iniziale                                                                                      & initial right bronchopneumonia                                                                        & 0.0                                & 0.0                                                                             \\
27                  & batterica broncopolmonite esterna infettiva otite polmonite specificata                                              & bacterial bronchopneumonia external infectious otitis pneumonia specified                             & 0.0                                & 0.0                                                                             \\
28                  & alte vie respiratorie infezione                                                                                      & high airways infection                                                                                & 0.0                                & 0.0                                                                             \\
29                  & alte vie respiratorie flogosi                                                                                        & high airways flogosis                                                                                 & 0.0                                & 0.0                                                                             \\
30                  & corso alte vie respiratorie flogosi                                                                                  & ongoing high airways flogosis                                                                         & 0.0                                & 0.0                                                                             \\
31                  & corso alte vie respiratorie virosi febbre                                                                            & ongoing high airways virosis fever                                                                    & 0.0                                & 0.0                                                                             \\
32                  & acuto broncospasmo lieve moderato                                                                                    & acute mild moderate bronchospasm                                                                      & 0.0                                & 0.0                                                                             \\
33                  & acute infezioni vie respiratorie superiori                                                                           & acute high airways infection                                                                          & 0.0                                & 0.0                                                                             \\
34                  & acuta alte vie respiratorie rinite virosi                                                                            & acute high airways virosis rhinitis                                                                   & 0.0                                & 0.0                                                                             \\ \hline

\end{tabular}
\end{table}
\end{landscape}

\begin{landscape}
\begin{table}[]
\caption{Keyword-based representation of the second level clusters related to respiratory infections from the original Umberto representation}
\label{tab:clusters-umb-resp-2liv}
\tiny
\begin{tabular}{p{1cm}p{4cm}p{4cm}p{2cm}p{4cm}p{4cm}}
\hline
\textbf{2nd level cluster \#} & \textbf{2nd level clusters keywords [ITA]}                                                               & \textbf{2nd level clusters keywords [ENG]}                                                         & \textbf{1st level cluster \#} & \textbf{1st level clusters keywords [ITA]}                                                               & \textbf{1st level clusters keywords [ENG]}                                                \\ \hline
\multirow{2}{*}{1}            & \multirow{2}{*}{\begin{tabular}[c]{@{}l@{}}acuta corso infezione\\ insufficienza respiratoria\end{tabular}}          & \multirow{2}{*}{\begin{tabular}[c]{@{}l@{}}acute bronchiolitis undergoing \\ respiratory failure\end{tabular}} & \# 9                          & acuta insufficienza respiratoria                                                                                     & acute respiratory failure                                                                             \\
                              &                                                                                                                      &                                                                                                                & \# 15                         & corso infezione respiratoria                                                                                         & ongoing respiratory infection                                                                         \\
\multirow{3}{*}{2}            & \multirow{3}{*}{broncospasmo episodio noto vie}                                                                      & \multirow{3}{*}{known bronchospasm episode airways}                                                           & \# 16                         & broncospasmo vie                                                                                                     & bronchospasm airways                                                                                  \\
                              &                                                                                                                      &                                                                                                                & \# 19                         & broncospasmo noto                                                                                                    & known bronchospasm                                                                                    \\
                              &                                                                                                                      &                                                                                                                & \# 12                         & broncospasmo episodio                                                                                                & episode bronchospasm                                                                                  \\
\multirow{2}{*}{3}            & \multirow{2}{*}{\begin{tabular}[c]{@{}l@{}}broncospasmo corso \\ paziente ricorrente virosi\end{tabular}}            & \multirow{2}{*}{\begin{tabular}[c]{@{}l@{}}bronchospasm ongoing\\ recurrent patient virosis\end{tabular}}      & \# 21                         & broncospasmo corso virosi                                                                                            & bronchospasm ongoing virosis                                                                          \\
                              &                                                                                                                      &                                                                                                                & \# 17                         & broncospasmo paziente ricorrente                                                                                     & bronchospasm recurrent patient                                                                        \\
\multirow{3}{*}{4}            & \multirow{3}{*}{broncospasmo lieve}                                                                                  & \multirow{3}{*}{bronchospasm mild}                                                                            & \# 6                          & bronchiolite broncospasmo lieve                                                                                      & bronchiolitis bronchospasm mild                                                                       \\
                              &                                                                                                                      &                                                                                                                & \# 7                          & bronchiolite broncospasmo lieve                                                                                      & bronchiolitis bronchospasm mild                                                                       \\
                              &                                                                                                                      &                                                                                                                & \# 32                         & acuto broncospasmo lieve moderato                                                                                    & acute bronchospasm mild moderate                                                                     \\
\multirow{2}{*}{5}            & \multirow{2}{*}{acuta bronchiolite}                                                                                  & \multirow{2}{*}{acute bronchiolitis}                                                                           & \# 2                          & acuta bronchiolite bronchite                                                                                         & acute bronchiolitis bronchitis                                                                        \\
                              &                                                                                                                      &                                                                                                                & \# 1                          & acuta bronchiolite lieve severa                                                                                      & acute bronchiolitis mild severe                                                                       \\
\multirow{5}{*}{6}            & \multirow{5}{*}{respiratorie vie}                                                                                    & \multirow{5}{*}{airways}                                                                                       & \# 28                         & alte infezione respiratorie vie                                                                                      & high airways infection                                                                                \\
                              &                                                                                                                      &                                                                                                                & \# 31                         & alte vie respiratorie corso febbre virosi                                                                            & high airways ongoing fever virosis                                                                    \\
                              &                                                                                                                      &                                                                                                                & \# 33                         & acute infezioni vie respiratorie superiori                                                                           & acute high airways infection                                                                          \\
                              &                                                                                                                      &                                                                                                                & \# 34                         & acuta alte respiratorie rinite vie virosi                                                                            & acute high airways rhinitis virosis                                                                   \\
                              &                                                                                                                      &                                                                                                                & \# 14                         & febbre recente respiratorie vie                                                                                      & recent fever airways                                                                                  \\
\multirow{3}{*}{8}            & \multirow{3}{*}{broncospasmo respiratoria}                                                                           & \multirow{3}{*}{bronchospasm respiratory}                                                                      & \# 23                         & broncospasmo corso infezione respiratoria                                                                            & bronchospasm ongoing respiratory infection                                                            \\
                              &                                                                                                                      &                                                                                                                & \# 25                         & broncospasmo corso flogosi lieve respiratoria                                                                        & bronchospasm ongoing flogosis mild respiratory                                                        \\
                              &                                                                                                                      &                                                                                                                & \# 10                         & broncospasmo respiratoria                                                                                            & bronchospasm respiratory                                                                              \\
\multirow{2}{*}{9}            & \multirow{2}{*}{broncospasmo corso vie}                                                                              & \multirow{2}{*}{bronchospasm ongoing airways}                                                                   & \# 24                         & broncospasmo corso flogosi vie                                                                                       & bronchospasm ongoing flogosis                                                                         \\
                              &                                                                                                                      &                                                                                                                & \# 22                         & broncospasmo corso respiratorie vie                                                                                  & bronchospasm ongoing airways                                                                          \\
\multirow{2}{*}{10}           & \multirow{2}{*}{alte flogosi respiratorie vie}                                                                       & \multirow{2}{*}{flogosis high airways}                                                                         & \# 30                         & corso flogosi alte vie respiratorie                                                                                  & flogosis high airways ongoing                                                                         \\
                              &                                                                                                                      &                                                                                                                & \# 29                         & flogosi alte vie respiratorie                                                                                        & flogosis high airways                                                                                 \\
11                            & bronchiolite respiratorio virus                                                                                      & bronchiolitis respiratory virus                                                                                & \# 3                          & bronchiolite respiratorio virus                                                                                      & bronchiolitis respiratory virus                                                                       \\
12                            & bronchiolite positiva vrs                                                                                            & bronchiolitis rsv positive                                                                                     & \# 4                          & bronchiolite positiva vrs                                                                                            & bronchiolitis rsv positive                                                                            \\
13                            & bronchiolite complicata corso ex gastroenterite paziente                                                             & bronchiolitis complicated course former gastroenteritis patient                                                & \# 5                          & bronchiolite complicata corso ex gastroenterite paziente                                                             & bronchiolitis complicated course former gastroenteritis patient                                       \\
14                            & allarme bronchiolite caratteristiche d epistassi insetto locale lussazione medicazione pregressa reazione sdn sn vrs & bronchiolitis alarm features epistaxis local insect dislocation dressing previous reaction sdn sn rsv          & \# 8                          & allarme bronchiolite caratteristiche d epistassi insetto locale lussazione medicazione pregressa reazione sdn sn vrs & bronchiolitis alarm features epistaxis local insect dislocation dressing previous reaction sdn sn rsv \\
15                            & febbrile respiratoria                                                                                                & respiratory fever                                                                                              & \# 13                         & febbrile respiratoria                                                                                                & respiratory fever                                                                                     \\
16                            & broncopolmonite destra iniziale                                                                                      & initial right bronchopneumonia                                                                                 & \# 26                         & broncopolmonite destra iniziale                                                                                      & initial right bronchopneumonia                                                                        \\
17                            & batterica broncopolmonite esterna infettiva otite polmonite specificata                                              & bacterial bronchopneumonia external infectious otitis pneumonia specified                                      & \# 27                         & batterica broncopolmonite esterna infettiva otite polmonite specificata                                              & bacterial bronchopneumonia external infectious otitis pneumonia specified \\
\bottomrule
\end{tabular}%
\end{table}
\end{landscape}

\section{Additional classification results}
\label{app:classification}
In this section, we report additional classification results to assess the effect and the differences among hospitals and LHUs, in Tables \ref{tab:res-hospitals} and \ref{tab:res-lhus}, respectively. 
Since the number of cases for each hospital and LHU is limited, we consider those with a minimum number of cases (15) and measure the results using a Leave One Group Out (LOGO) approach. For each hospital/LHU to be evaluated, we train the model on the data from the other hospitals/LHUs, and we measure the performance on the data from the hospital/LHU of interest. 
We perform these analyses only on the bronchiolitis dataset, due to the smaller numbers of the bronchitis dataset.

In Table \ref{tab:res-pediatric} we also compare results between the groups of pediatric vs non-pediatric ERs/departments admissions. In this case, we adopt the same 5/10-fold stratified cross-validation used to compare different classification models, dividing the test set between pediatric and non-pediatric ERs/departments at each iteration and computing the metrics on these two groups separately.

We identified the records related to pediatric ERs/departments by looking at the presence of specific keywords (\textit{pediatria, pediatrico, pediatrica}) in the header of the letters. This marked 10,340 records (31.17\%) as related to pediatric ERs/departments in the bronchiolitis dataset, and 767 (24.07\%) in the bronchitis dataset.

Table~\ref{tab:ablation1stlevel} reports results for the bronchitis dataset using weak labels obtained at the first-level clustering, while Table~\ref{tab:weak-results} reports classification performance evaluated on weak labels.

Table~\ref{tab:undersampling} reports additional results with different undersampling levels and models, on the bronchiolitis dataset.

\setcounter{table}{0}

\begin{table}[h!]
\caption{10-fold CV results for classification, comparing different hospitals (only hospitals with more than 15 bronchiolitis cases). G = gold labels, W = weak labels.}
\label{tab:res-hospitals}
\resizebox{\textwidth}{!}{%
\begin{tabular}{lrrrrrrrrrrr}
\toprule
\textbf{Hospital} &
  \multicolumn{1}{l}{\textbf{Dataset \%}} &
  \multicolumn{1}{l}{\textbf{\begin{tabular}[c]{@{}l@{}}Pediatric ER \\ (\% records)\end{tabular}}} &
  \multicolumn{1}{l}{\textbf{\begin{tabular}[c]{@{}l@{}}Bronchiolitis\\ (\% records)\end{tabular}}} &
  \multicolumn{1}{l}{\textbf{\begin{tabular}[c]{@{}l@{}}P  {[}\%{]}\\ (W)\end{tabular}}} &
  \multicolumn{1}{l}{\textbf{\begin{tabular}[c]{@{}l@{}}R {[}\%{]}\\ (W)\end{tabular}}} &
  \multicolumn{1}{l}{\textbf{\begin{tabular}[c]{@{}l@{}}F1 {[}\%{]}\\ (W)\end{tabular}}} &
  \multicolumn{1}{l}{\textbf{\begin{tabular}[c]{@{}l@{}}AUC {[}\%{]}\\ (W)\end{tabular}}} &
  \multicolumn{1}{l}{\textbf{\begin{tabular}[c]{@{}l@{}}P {[}\%{]}\\ (G)\end{tabular}}} &
  \multicolumn{1}{l}{\textbf{\begin{tabular}[c]{@{}l@{}}R {[}\%{]}\\ (G)\end{tabular}}} &
  \multicolumn{1}{l}{\textbf{\begin{tabular}[c]{@{}l@{}}F1 {[}\%{]}\\ (G)\end{tabular}}} &
  \multicolumn{1}{l}{\textbf{\begin{tabular}[c]{@{}l@{}}AUC {[}\%{]}\\ (G)\end{tabular}}} \\ \midrule
\textit{H-0} & 31.41   & 87.30 & 0.76 & 77.46  & 78.57 & 78.01 & 90.40 & 81.40  & 63.64 & 71.43 & 86.56 \\
\textit{H-1} & 14.56   & 0.00  & 1.08 & 63.64  & 71.79 & 79.43 & 85.32 & 66.67  & 66.67 & 66.67 & 77.95 \\
\textit{H-2} & 11.63   & 1.00  & 0.78 & 80.00  & 72.73 & 76.19 & 92.16 & 50.00  & 71.43 & 58.82 & 83.98 \\
\textit{H-3} & 10.23   & 12.00 & 1.71 & 88.68  & 88.68 & 88.68 & 89.45 & 81.40  & 81.40 & 81.40 & 88.12 \\
\textit{H-4} & 7.43 & 22.00 & 1.22 & 92.31  & 35.29 & 51.06 & 64.32 & 84.21  & 37.21 & 51.61 & 57.34 \\
\textit{H-6} & 2.59 & 9.40  & 2.09 & 100.00 & 31.25 & 47.62 & 65.63 & 100.00 & 33.33 & 50.00 & 70.48 \\ \bottomrule
\end{tabular}
}
\end{table}

\begin{table}[h!]
\caption{10-fold CV results for classification, comparing different LHUs (only LHUs with more than 15 cases). Std. Dev. in parenthesis. G = gold labels, W = weak labels.}
\label{tab:res-lhus}
\resizebox{\textwidth}{!}{%
\begin{tabular}{@{}lrrrrrrrr@{}}
\hline
\textbf{LHU}                & \textbf{\begin{tabular}[c]{@{}l@{}}P {[}\%{]}\\ (W)\end{tabular}} & \textbf{\begin{tabular}[c]{@{}l@{}}R {[}\%{]} \\ (W)\end{tabular}} & \textbf{\begin{tabular}[c]{@{}l@{}}F1 {[}\%{]}\\ (W)\end{tabular}} & \textbf{\begin{tabular}[c]{@{}l@{}}AUC {[}\%{]}\\ (W)\end{tabular}} & \textbf{\begin{tabular}[c]{@{}l@{}}P {[}\%{]}\\ (G)\end{tabular}} & \textbf{\begin{tabular}[c]{@{}l@{}}R {[}\%{]}\\ (G)\end{tabular}} & \textbf{\begin{tabular}[c]{@{}l@{}}F1 {[}\%{]}\\ (G)\end{tabular}} & \textbf{\begin{tabular}[c]{@{}l@{}}AUC {[}\%{]}\\ (G)\end{tabular}} \\ \hline
\textit{LHU-3}      & 75.56                                                             & 65.38                                                              & 70.10                                                              & 78.81                                                               & 69.70                                                             & 53.49                                                             & 60.53                                                              & 73.94                                                               \\
\textit{LHU-5} & 83.33                                                             & 40.00                                                              & 54.05                                                              & 65.77                                                               & 45.83                                                             & 64.71                                                             & 53.66                                                              & 58.25                                                               \\
\textit{LHU-2}          & 94.12                                                             & 64.00                                                              & 76.19                                                              & 88.89                                                               & 51.61                                                             & 94.12                                                             & 66.67                                                              & 59.95                                                               \\
\textit{LHU-6}      & 64.29                                                             & 81.81                                                              & 72.00                                                              & 91.17                                                               & 51.67                                                             & 71.43                                                             & 59.96                                                              & 83.52                                                               \\
\textit{LHU-7}           & 86.27                                                             & 83.02                                                              & 84.62                                                              & 87.06                                                               & 82.50                                                             & 76.74                                                             & 79.52                                                              & 86.22                                                               \\
\textit{LHU-0}             & 77.46                                                             & 78.57                                                              & 78.01                                                              & 90.40                                                               & 81.40                                                             & 63.64                                                             & 71.43                                                              & 86.56                                                               \\ \hline
\end{tabular}%
}
\end{table}

\begin{table}[h!]
\caption{10-fold CV results for classification, comparing records from pediatric ERs/Departments with records from non-pediatric ERs/Departments. Std. Dev. in parenthesis. G = gold labels, W = weak labels.}
\label{tab:res-pediatric}
\resizebox{\textwidth}{!}{%
\begin{tabular}{lllllllll}
\hline
\textbf{Pediatrics ER}                & \textbf{\begin{tabular}[c]{@{}l@{}}P [\%]\\ (W)\end{tabular}} & \textbf{\begin{tabular}[c]{@{}l@{}}R [\%]\\ (W)\end{tabular}} & \textbf{\begin{tabular}[c]{@{}l@{}}F1 [\%]\\ (W)\end{tabular}} & \textbf{\begin{tabular}[c]{@{}l@{}}AUC [\%]\\ (W)\end{tabular}} & \textbf{\begin{tabular}[c]{@{}l@{}}P [\%]\\ (G)\end{tabular}} & \textbf{\begin{tabular}[c]{@{}l@{}}R [\%]\\ (G)\end{tabular}} & \textbf{\begin{tabular}[c]{@{}l@{}}F1 [\%]\\ (G)\end{tabular}} & \textbf{\begin{tabular}[c]{@{}l@{}}AUC [\%]\\ (G)\end{tabular}} \\ \hline
\textit{Yes} & 87.79 (7.12)                                                                           & 83.31 (4.30)                                                                         & 85.13 (6.56)                                                                           & 86.04 (5.29)                                                                           & 75.89 (9.85)                                                                        & 88.57 (8.23)   & 81.74 (7.71) & 80.89 (6.37)                                                                       \\
\textit{No} & 72.47 (7.28)                                                                           & 80.52 (7.87)                                                                         & 76.08 (5.64)                                                                          &  75.93 (4.04)                                                                          & 60.29 (8.29)                                                                        & 85.21 (6.11) & 69.37 (4.36) & 71.07 (4.57)                                                                          \\ \hline
\end{tabular}
}
\end{table}

\begin{table}[]
\centering
\caption{CV results for classification trained with first-level weak labels, evaluated on gold labels, on the bronchitis dataset. Std. Dev. in parenthesis}
\label{tab:ablation1stlevel}
\resizebox{\textwidth}{!}{%
\begin{tabular}{llrrrrr}
\toprule
\multicolumn{1}{l}{\multirow{2}{*}{\textbf{Learning}}} &
  \multicolumn{1}{l}{\multirow{2}{*}{\textbf{Classifier}}} &
  \multicolumn{5}{c}{\textbf{Bronchitis}} \\
\multicolumn{1}{l}{} &
  \multicolumn{1}{l}{} &
  \multicolumn{1}{r}{\textbf{P [\%]}} &
  \multicolumn{1}{r}{\textbf{R [\%]}} &
  \multicolumn{1}{r}{\textbf{F1 [\%]}} &
  \multicolumn{1}{r}{\textbf{AUROC [\%]}} &
  \textbf{AUPRC [\%]} \\ \midrule

\multicolumn{7}{c}{\textbf{With diagnosis string}} \\ \midrule

\multicolumn{1}{l}{Weakly supervised} &
  \multicolumn{1}{l}{WS-UmBERTo-TPT} &
  \multicolumn{1}{r}{\begin{tabular}[c]{@{}r@{}}74.41\\ (5.24)\end{tabular}} &
  \multicolumn{1}{r}{\begin{tabular}[c]{@{}r@{}}79.51\\ (5.63)\end{tabular}} &
  \multicolumn{1}{r}{\begin{tabular}[c]{@{}r@{}}76.85\\ (5.26)\end{tabular}} &
  \multicolumn{1}{r}{\begin{tabular}[c]{@{}r@{}}79.69\\ (4.98)\end{tabular}} &
  \begin{tabular}[c]{@{}r@{}}75.02\\ (5.85)\end{tabular} \\ 

\multicolumn{1}{l}{} &
  \multicolumn{1}{l}{WS-XLM-RoBERTa-TPT} &
  \multicolumn{1}{r}{\begin{tabular}[c]{@{}r@{}}72.14\\ (5.36)\end{tabular}} &
  \multicolumn{1}{r}{\begin{tabular}[c]{@{}r@{}}72.41\\ (5.47)\end{tabular}} &
  \multicolumn{1}{r}{\begin{tabular}[c]{@{}r@{}}72.23\\ (5.30)\end{tabular}} &
  \multicolumn{1}{r}{\begin{tabular}[c]{@{}r@{}}74.15\\ (4.86)\end{tabular}} &
  \begin{tabular}[c]{@{}r@{}}69.21\\ (8.73)\end{tabular} \\ 

\multicolumn{1}{l}{} &
  \multicolumn{1}{l}{WS-RemBERT-TPT} &
  \multicolumn{1}{r}{\begin{tabular}[c]{@{}r@{}}71.14\\ (4.94)\end{tabular}} &
  \multicolumn{1}{r}{\begin{tabular}[c]{@{}r@{}}71.85\\ (5.84)\end{tabular}} &
  \multicolumn{1}{r}{\begin{tabular}[c]{@{}r@{}}71.48\\ (5.46)\end{tabular}} &
  \multicolumn{1}{r}{\begin{tabular}[c]{@{}r@{}}74.86\\ (4.98)\end{tabular}} &
  \begin{tabular}[c]{@{}r@{}}66.34\\ (4.70)\end{tabular} \\ 

\multicolumn{1}{l}{} &
  \multicolumn{1}{l}{WS-UmBERTo} &
  \multicolumn{1}{r}{\begin{tabular}[c]{@{}r@{}}73.89\\ (5.65)\end{tabular}} &
  \multicolumn{1}{r}{\begin{tabular}[c]{@{}r@{}}73.51\\ (5.98)\end{tabular}} &
  \multicolumn{1}{r}{\begin{tabular}[c]{@{}r@{}}73.67\\ (5.54)\end{tabular}} &
  \multicolumn{1}{r}{\begin{tabular}[c]{@{}r@{}}77.21\\ (5.36)\end{tabular}} &
  \begin{tabular}[c]{@{}r@{}}73.70\\ (3.75)\end{tabular} \\ 

\multicolumn{1}{l}{} &
  \multicolumn{1}{l}{WS-XLM-RoBERTa} &
  \multicolumn{1}{r}{\begin{tabular}[c]{@{}r@{}}70.95\\ (5.76)\end{tabular}} &
  \multicolumn{1}{r}{\begin{tabular}[c]{@{}r@{}}70.54\\ (5.88)\end{tabular}} &
  \multicolumn{1}{r}{\begin{tabular}[c]{@{}r@{}}70.71\\ (5.62)\end{tabular}} &
  \multicolumn{1}{r}{\begin{tabular}[c]{@{}r@{}}75.26\\ (5.45)\end{tabular}} &
  \begin{tabular}[c]{@{}r@{}}70.62\\ (4.44)\end{tabular} \\ 

\multicolumn{1}{l}{} &
  \multicolumn{1}{l}{WS-RemBERT} &
  \multicolumn{1}{r}{\begin{tabular}[c]{@{}r@{}}69.16\\ (5.69)\end{tabular}} &
  \multicolumn{1}{r}{\begin{tabular}[c]{@{}r@{}}70.94\\ (5.46)\end{tabular}} &
  \multicolumn{1}{r}{\begin{tabular}[c]{@{}r@{}}70.08\\ (5.51)\end{tabular}} &
  \multicolumn{1}{r}{\begin{tabular}[c]{@{}r@{}}74.69\\ (5.46)\end{tabular}} &
  \begin{tabular}[c]{@{}r@{}}65.75\\ (4.82)\end{tabular} \\ 

\multicolumn{1}{l}{} &
  \multicolumn{1}{l}{WS-LSTM} &
  \multicolumn{1}{r}{\begin{tabular}[c]{@{}r@{}}61.75\\ (5.52)\end{tabular}} &
  \multicolumn{1}{r}{\begin{tabular}[c]{@{}r@{}}61.23\\ (5.91)\end{tabular}} &
  \multicolumn{1}{r}{\begin{tabular}[c]{@{}r@{}}61.47\\ (5.46)\end{tabular}} &
  \multicolumn{1}{r}{\begin{tabular}[c]{@{}r@{}}68.84\\ (5.91)\end{tabular}} &
  \begin{tabular}[c]{@{}r@{}}58.11\\ (3.86)\end{tabular} \\ \midrule

\multicolumn{7}{c}{\textbf{Without diagnosis string}} \\ \midrule 

\multicolumn{1}{l}{Weakly supervised} &
  \multicolumn{1}{l}{WS-UmBERTo-TPT} &
  \multicolumn{1}{r}{\begin{tabular}[c]{@{}r@{}}73.15\\ (5.67)\end{tabular}} &
  \multicolumn{1}{r}{\begin{tabular}[c]{@{}r@{}}76.89\\ (5.84)\end{tabular}} &
  \multicolumn{1}{r}{\begin{tabular}[c]{@{}r@{}}74.56\\ (5.23)\end{tabular}} &
  \multicolumn{1}{r}{\begin{tabular}[c]{@{}r@{}}76.51\\ (5.42)\end{tabular}} &
  \begin{tabular}[c]{@{}r@{}}74.53\\ (4.92)\end{tabular} \\ 

\multicolumn{1}{l}{} &
  \multicolumn{1}{l}{WS-XLM-RoBERTa-TPT} &
  \multicolumn{1}{r}{\begin{tabular}[c]{@{}r@{}}70.54\\ (5.48)\end{tabular}} &
  \multicolumn{1}{r}{\begin{tabular}[c]{@{}r@{}}67.94\\ (5.64)\end{tabular}} &
  \multicolumn{1}{r}{\begin{tabular}[c]{@{}r@{}}68.97\\ (5.32)\end{tabular}} &
  \multicolumn{1}{r}{\begin{tabular}[c]{@{}r@{}}73.56\\ (5.47)\end{tabular}} &
  \begin{tabular}[c]{@{}r@{}}64.45\\ (6.89)\end{tabular} \\ 

\multicolumn{1}{l}{} &
  \multicolumn{1}{l}{WS-RemBERT-TPT} &
  \multicolumn{1}{r}{\begin{tabular}[c]{@{}r@{}}70.36\\ (5.84)\end{tabular}} &
  \multicolumn{1}{r}{\begin{tabular}[c]{@{}r@{}}70.28\\ (5.52)\end{tabular}} &
  \multicolumn{1}{r}{\begin{tabular}[c]{@{}r@{}}69.96\\ (5.59)\end{tabular}} &
  \multicolumn{1}{r}{\begin{tabular}[c]{@{}r@{}}72.11\\ (5.67)\end{tabular}} &
  \begin{tabular}[c]{@{}r@{}}64.95\\ (5.40)\end{tabular} \\ 

\multicolumn{1}{l}{} &
  \multicolumn{1}{l}{WS-UmBERTo} &
  \multicolumn{1}{r}{\begin{tabular}[c]{@{}r@{}}73.48\\ (5.69)\end{tabular}} &
  \multicolumn{1}{r}{\begin{tabular}[c]{@{}r@{}}71.12\\ (5.45)\end{tabular}} &
  \multicolumn{1}{r}{\begin{tabular}[c]{@{}r@{}}72.56\\ (5.53)\end{tabular}} &
  \multicolumn{1}{r}{\begin{tabular}[c]{@{}r@{}}71.69\\ (5.49)\end{tabular}} &
  \begin{tabular}[c]{@{}r@{}}73.70\\ (2.83)\end{tabular} \\ 

\multicolumn{1}{l}{} &
  \multicolumn{1}{l}{WS-XLM-RoBERTa} &
  \multicolumn{1}{r}{\begin{tabular}[c]{@{}r@{}}69.56\\ (5.64)\end{tabular}} &
  \multicolumn{1}{r}{\begin{tabular}[c]{@{}r@{}}67.41\\ (5.55)\end{tabular}} &
  \multicolumn{1}{r}{\begin{tabular}[c]{@{}r@{}}68.46\\ (5.72)\end{tabular}} &
  \multicolumn{1}{r}{\begin{tabular}[c]{@{}r@{}}68.84\\ (5.38)\end{tabular}} &
  \begin{tabular}[c]{@{}r@{}}67.93\\ (5.77)\end{tabular} \\ 

\multicolumn{1}{l}{} &
  \multicolumn{1}{l}{WS-RemBERT} &
  \multicolumn{1}{r}{\begin{tabular}[c]{@{}r@{}}69.42\\ (5.41)\end{tabular}} &
  \multicolumn{1}{r}{\begin{tabular}[c]{@{}r@{}}68.06\\ (5.66)\end{tabular}} &
  \multicolumn{1}{r}{\begin{tabular}[c]{@{}r@{}}68.64\\ (5.37)\end{tabular}} &
  \multicolumn{1}{r}{\begin{tabular}[c]{@{}r@{}}72.09\\ (5.42)\end{tabular}} &
  \begin{tabular}[c]{@{}r@{}}65.55\\ (4.51)\end{tabular} \\ 

\multicolumn{1}{l}{} &
  \multicolumn{1}{l}{WS-LSTM} &
  \multicolumn{1}{r}{\begin{tabular}[c]{@{}r@{}}59.68\\ (5.43)\end{tabular}} &
  \multicolumn{1}{r}{\begin{tabular}[c]{@{}r@{}}62.84\\ (5.64)\end{tabular}} &
  \multicolumn{1}{r}{\begin{tabular}[c]{@{}r@{}}61.20\\ (5.36)\end{tabular}} &
  \multicolumn{1}{r}{\begin{tabular}[c]{@{}r@{}}65.02\\ (5.12)\end{tabular}} &
  \begin{tabular}[c]{@{}r@{}}59.26\\ (4.92)\end{tabular} \\ \bottomrule

\end{tabular}
}
\end{table}

\begin{table}[]
\caption{CV results for classification on weak labels, for the bronchiolitis dataset. Std. Dev. in parenthesis.}
\label{tab:weak-results}
\resizebox{\textwidth}{!}{%
\begin{tabular}{llrrrrr}
\toprule
\multicolumn{1}{l}{\textbf{Learning}} &
  \multicolumn{1}{l}{\textbf{Classifier}} &
  \multicolumn{1}{r}{\textbf{P {[}\%{]}}} &
  \multicolumn{1}{r}{\textbf{R {[}\%{]}}} &
  \multicolumn{1}{r}{\textbf{F1 {[}\%{]}}} &
  \multicolumn{1}{r}{\textbf{AUROC {[}\%{]}}} &
  \textbf{AUPRC {[}\%{]}} \\ \midrule
\multicolumn{7}{c}{\textbf{With diagnosis string}} \\  \midrule
\multicolumn{1}{l}{Weakly supervised} &
  \multicolumn{1}{l}{WS-UmBERTo-TPT} &
  \multicolumn{1}{r}{\begin{tabular}[c]{@{}r@{}}82.25\\ (5.52)\end{tabular}} &
  \multicolumn{1}{r}{\begin{tabular}[c]{@{}r@{}}80.24\\ (7.01)\end{tabular}} &
  \multicolumn{1}{r}{\begin{tabular}[c]{@{}r@{}}81.35\\ (4.10)\end{tabular}} &
  \multicolumn{1}{r}{\begin{tabular}[c]{@{}r@{}}82.30\\ (3.06)\end{tabular}} &
  \begin{tabular}[c]{@{}r@{}}80.59\\ (3.07)\end{tabular} \\ 
\multicolumn{1}{l}{} &
  \multicolumn{1}{l}{WS-XLM-RoBERTa-TPT} &
  \multicolumn{1}{r}{\begin{tabular}[c]{@{}r@{}}76.74\\ (3.69)\end{tabular}} &
  \multicolumn{1}{r}{\begin{tabular}[c]{@{}r@{}}74.91\\ (7.08)\end{tabular}} &
  \multicolumn{1}{r}{\begin{tabular}[c]{@{}r@{}}75.89\\ (3.14)\end{tabular}} &
  \multicolumn{1}{r}{\begin{tabular}[c]{@{}r@{}}75.53\\ (3.44)\end{tabular}} &
  \begin{tabular}[c]{@{}r@{}}72.49\\ (4.45)\end{tabular} \\ 
\multicolumn{1}{l}{} &
  \multicolumn{1}{l}{WS-RemBERT-TPT} &
  \multicolumn{1}{r}{\begin{tabular}[c]{@{}r@{}}76.34\\ (5.58)\end{tabular}} &
  \multicolumn{1}{r}{\begin{tabular}[c]{@{}r@{}}75.57\\ (6.14)\end{tabular}} &
  \multicolumn{1}{r}{\begin{tabular}[c]{@{}r@{}}75.85\\ (5.19)\end{tabular}} &
  \multicolumn{1}{r}{\begin{tabular}[c]{@{}r@{}}74.51\\ (3.87)\end{tabular}} &
  \begin{tabular}[c]{@{}r@{}}73.06\\ (2.79)\end{tabular} \\ 
\multicolumn{1}{l}{} &
  \multicolumn{1}{l}{WS-UmBERTo} &
  \multicolumn{1}{r}{\begin{tabular}[c]{@{}r@{}}80.67\\ (5.21)\end{tabular}} &
  \multicolumn{1}{r}{\begin{tabular}[c]{@{}r@{}}75.48\\ (6.45)\end{tabular}} &
  \multicolumn{1}{r}{\begin{tabular}[c]{@{}r@{}}78.01\\ (4.02)\end{tabular}} &
  \multicolumn{1}{r}{\begin{tabular}[c]{@{}r@{}}78.67\\ (2.97)\end{tabular}} &
  \begin{tabular}[c]{@{}r@{}}74.77\\ (4.56)\end{tabular} \\ 
\multicolumn{1}{l}{} &
  \multicolumn{1}{l}{WS-XLM-RoBERTa} &
  \multicolumn{1}{r}{\begin{tabular}[c]{@{}r@{}}75.53\\ (4.56)\end{tabular}} &
  \multicolumn{1}{r}{\begin{tabular}[c]{@{}r@{}}70.73\\ (6.44)\end{tabular}} &
  \multicolumn{1}{r}{\begin{tabular}[c]{@{}r@{}}73.24\\ (4.28)\end{tabular}} &
  \multicolumn{1}{r}{\begin{tabular}[c]{@{}r@{}}73.82\\ (3.22)\end{tabular}} &
  \begin{tabular}[c]{@{}r@{}}71.46\\ (3.95)\end{tabular} \\ 
\multicolumn{1}{l}{} &
  \multicolumn{1}{l}{WS-RemBERT} &
  \multicolumn{1}{r}{\begin{tabular}[c]{@{}r@{}}74.15\\ (5.62)\end{tabular}} &
  \multicolumn{1}{r}{\begin{tabular}[c]{@{}r@{}}65.35\\ (6.94)\end{tabular}} &
  \multicolumn{1}{r}{\begin{tabular}[c]{@{}r@{}}70.41\\ (4.14)\end{tabular}} &
  \multicolumn{1}{r}{\begin{tabular}[c]{@{}r@{}}72.51\\ (3.74)\end{tabular}} &
  \begin{tabular}[c]{@{}r@{}}69.18\\ (4.33)\end{tabular} \\ 
\multicolumn{1}{l}{} &
  \multicolumn{1}{l}{WS-LSTM} &
  \multicolumn{1}{r}{\begin{tabular}[c]{@{}r@{}}68.57\\ (6.48)\end{tabular}} &
  \multicolumn{1}{r}{\begin{tabular}[c]{@{}r@{}}64.97\\ (9.96)\end{tabular}} &
  \multicolumn{1}{r}{\begin{tabular}[c]{@{}r@{}}65.52\\ (5.56)\end{tabular}} &
  \multicolumn{1}{r}{\begin{tabular}[c]{@{}r@{}}67.50\\ (4.99)\end{tabular}} &
  \begin{tabular}[c]{@{}r@{}}63.67\\ (4.93)\end{tabular} \\ \midrule
\multicolumn{1}{l}{Unsupervised} &
  \multicolumn{1}{l}{Rule-based full-text} &
  \multicolumn{1}{r}{\begin{tabular}[c]{@{}r@{}}70.18\\ (2.42)\end{tabular}} &
  \multicolumn{1}{r}{\begin{tabular}[c]{@{}r@{}}87.05\\ (3.93)\end{tabular}} &
  \multicolumn{1}{r}{\begin{tabular}[c]{@{}r@{}}77.74\\ (2.61)\end{tabular}} &
  \multicolumn{1}{r}{-} &
  - \\ 
\multicolumn{1}{l}{} &
  \multicolumn{1}{l}{Rule-based diagnosis} &
  \multicolumn{1}{r}{\begin{tabular}[c]{@{}r@{}}73.88\\ (2.26)\end{tabular}} &
  \multicolumn{1}{r}{\begin{tabular}[c]{@{}r@{}}86.38\\ (3.11)\end{tabular}} &
  \multicolumn{1}{r}{\begin{tabular}[c]{@{}r@{}}79.85\\ (2.33)\end{tabular}} &
  \multicolumn{1}{r}{-} &
  - \\ 
\multicolumn{1}{l}{} &
  \multicolumn{1}{l}{LLaMAntino-3} &
  \multicolumn{1}{r}{\begin{tabular}[c]{@{}r@{}}57.39\\ (8.31)\end{tabular}} &
  \multicolumn{1}{r}{\begin{tabular}[c]{@{}r@{}}63.78\\ (4.47)\end{tabular}} &
  \multicolumn{1}{r}{\begin{tabular}[c]{@{}r@{}}79.64\\ (5.96)\end{tabular}} &
  \multicolumn{1}{r}{\begin{tabular}[c]{@{}r@{}}70.87\\ (2.37)\end{tabular}} &
  \begin{tabular}[c]{@{}r@{}}56.44\\ (5.05)\end{tabular} \\ \midrule
\multicolumn{7}{c}{\textbf{Without diagnosis string}} \\ \midrule
\multicolumn{1}{l}{Weakly supervised} &
  \multicolumn{1}{l}{WS-UmBERTo-TPT} &
  \multicolumn{1}{r}{\begin{tabular}[c]{@{}r@{}}84.43\\ (6.02)\end{tabular}} &
  \multicolumn{1}{r}{\begin{tabular}[c]{@{}r@{}}77.63\\  (7.57)\end{tabular}} &
  \multicolumn{1}{r}{\begin{tabular}[c]{@{}r@{}}80.79\\  (5.47)\end{tabular}} &
  \multicolumn{1}{r}{\begin{tabular}[c]{@{}r@{}}80.57\\ (3.58)\end{tabular}} &
  \begin{tabular}[c]{@{}r@{}}80.30\\ (3.82)\end{tabular} \\ 
\multicolumn{1}{l}{} &
  \multicolumn{1}{l}{WS-XLM-RoBERTa-TPT} &
  \multicolumn{1}{r}{\begin{tabular}[c]{@{}r@{}}74.47\\ (3.95)\end{tabular}} &
  \multicolumn{1}{r}{\begin{tabular}[c]{@{}r@{}}71.68\\  (7.24)\end{tabular}} &
  \multicolumn{1}{r}{\begin{tabular}[c]{@{}r@{}}73.11\\ (3.54)\end{tabular}} &
  \multicolumn{1}{r}{\begin{tabular}[c]{@{}r@{}}75.62\\ (3.67)\end{tabular}} &
  \begin{tabular}[c]{@{}r@{}}69.51\\ (4.34)\end{tabular} \\ 
\multicolumn{1}{l}{} &
  \multicolumn{1}{l}{WS-RemBERT-TPT} &
  \multicolumn{1}{r}{\begin{tabular}[c]{@{}r@{}}72.13\\ (5.64)\end{tabular}} &
  \multicolumn{1}{r}{\begin{tabular}[c]{@{}r@{}}72.64\\  (6.02)\end{tabular}} &
  \multicolumn{1}{r}{\begin{tabular}[c]{@{}r@{}}72.40\\ (5.28)\end{tabular}} &
  \multicolumn{1}{r}{\begin{tabular}[c]{@{}r@{}}72.27\\ (3.96)\end{tabular}} &
  \begin{tabular}[c]{@{}r@{}}68.92\\ (2.92)\end{tabular} \\ 
\multicolumn{1}{l}{} &
  \multicolumn{1}{l}{WS-UmBERTo} &
  \multicolumn{1}{r}{\begin{tabular}[c]{@{}r@{}}80.08\\ (5.89)\end{tabular}} &
  \multicolumn{1}{r}{\begin{tabular}[c]{@{}r@{}}72.15\\ (6.74)\end{tabular}} &
  \multicolumn{1}{r}{\begin{tabular}[c]{@{}r@{}}76.02\\ (4.95)\end{tabular}} &
  \multicolumn{1}{r}{\begin{tabular}[c]{@{}r@{}}76.16\\ (3.55)\end{tabular}} &
  \begin{tabular}[c]{@{}r@{}}73.26\\ (4.77)\end{tabular} \\ 
\multicolumn{1}{l}{} &
  \multicolumn{1}{l}{WS-XLM-RoBERTa} &
  \multicolumn{1}{r}{\begin{tabular}[c]{@{}r@{}}72.21\\ (4.77)\end{tabular}} &
  \multicolumn{1}{r}{\begin{tabular}[c]{@{}r@{}}65.56\\ (7.88)\end{tabular}} &
  \multicolumn{1}{r}{\begin{tabular}[c]{@{}r@{}}68.74\\ (5.04)\end{tabular}} &
  \multicolumn{1}{r}{\begin{tabular}[c]{@{}r@{}}69.62\\ (3.66)\end{tabular}} &
  \begin{tabular}[c]{@{}r@{}}68.55\\ (4.38)\end{tabular} \\ 
\multicolumn{1}{l}{} &
  \multicolumn{1}{l}{WS-RemBERT} &
  \multicolumn{1}{r}{\begin{tabular}[c]{@{}r@{}}69.86\\ (5.96)\end{tabular}} &
  \multicolumn{1}{r}{\begin{tabular}[c]{@{}r@{}}68.32\\ (7.02)\end{tabular}} &
  \multicolumn{1}{r}{\begin{tabular}[c]{@{}r@{}}69.25\\ (4.87)\end{tabular}} &
  \multicolumn{1}{r}{\begin{tabular}[c]{@{}r@{}}70.34\\ (4.21)\end{tabular}} &
  \begin{tabular}[c]{@{}r@{}}67.05\\ (4.52)\end{tabular} \\ 
\multicolumn{1}{l}{} &
  \multicolumn{1}{l}{WS-LSTM} &
  \multicolumn{1}{r}{\begin{tabular}[c]{@{}r@{}}64.98\\ (6.89)\end{tabular}} &
  \multicolumn{1}{r}{\begin{tabular}[c]{@{}r@{}}56.22\\ (10.88)\end{tabular}} &
  \multicolumn{1}{r}{\begin{tabular}[c]{@{}r@{}}60.54\\ (6.52)\end{tabular}} &
  \multicolumn{1}{r}{\begin{tabular}[c]{@{}r@{}}63.00\\ (5.24)\end{tabular}} &
  \begin{tabular}[c]{@{}r@{}}56.14\\ (5.12)\end{tabular} \\ \midrule
\multicolumn{1}{l}{Unsupervised} &
  \multicolumn{1}{l}{Rule-based full-text} &
  \multicolumn{1}{r}{\begin{tabular}[c]{@{}r@{}}65.54\\ (2.42)\end{tabular}} &
  \multicolumn{1}{r}{\begin{tabular}[c]{@{}r@{}}79.16\\ (3.93)\end{tabular}} &
  \multicolumn{1}{r}{\begin{tabular}[c]{@{}r@{}}71.28\\ (2.61)\end{tabular}} &
  \multicolumn{1}{r}{-} &
  - \\ 
\multicolumn{1}{l}{} &
  \multicolumn{1}{l}{LLaMAntino-3} &
  \multicolumn{1}{r}{\begin{tabular}[c]{@{}r@{}}49.81\\ (5.27)\end{tabular}} &
  \multicolumn{1}{r}{\begin{tabular}[c]{@{}r@{}}79.91\\ (6.35)\end{tabular}} &
  \multicolumn{1}{r}{\begin{tabular}[c]{@{}r@{}}61.12\\ (4.98)\end{tabular}} &
  \multicolumn{1}{r}{\begin{tabular}[c]{@{}r@{}}65.32\\ (6.40)\end{tabular}} &
  \begin{tabular}[c]{@{}r@{}}57.54\\ (3.83)\end{tabular} \\ \bottomrule
\end{tabular}
}
\end{table}

\begin{table}[t]
\centering
\caption{CV results for classification against gold labels on the bronchiolitis dataset, with diagnosis strings, under different undersampling strategies. Std. Dev. in parenthesis.}
\label{tab:undersampling}
\resizebox{\textwidth}{!}{%
\begin{tabular}{llrrrrr}
\toprule
\textbf{Model} & \textbf{Undersampling} & \textbf{P [\%]} & \textbf{R [\%]} & \textbf{F1 [\%]} & \textbf{AUROC [\%]} & \textbf{AUPRC [\%]} \\ \midrule

\multirow{3}{*}{WS-UmBERTo-TPT}
& None &
\makecell[r]{4.78\\(0.8)} &
\makecell[r]{81.70\\(6.6)} &
\makecell[r]{9.04\\(1.5)} &
\makecell[r]{90.22\\(3.1)} &
\makecell[r]{23.43\\(8.3)} \\

& 1:2 &
\makecell[r]{78.34\\(6.8)} &
\makecell[r]{77.57\\(6.9)} &
\makecell[r]{78.14\\(4.9)} &
\makecell[r]{77.68\\(4.3)} &
\makecell[r]{73.13\\(4.9)} \\

& 1:1 &
\makecell[r]{53.21\\(4.8)} &
\makecell[r]{81.60\\(8.3)} &
\makecell[r]{64.22\\(4.7)} &
\makecell[r]{86.33\\(3.9)} &
\makecell[r]{65.61\\(10.6)} \\ \midrule

\multirow{3}{*}{FS-UmBERTo-TPT}
& None &
\makecell[r]{5.11\\(0.6)} &
\makecell[r]{86.20\\(7.6)} &
\makecell[r]{9.55\\(1.0)} &
\makecell[r]{92.32\\(1.8)} &
\makecell[r]{23.94\\(6.3)} \\

& 1:2 &
\makecell[r]{80.30\\(6.2)} &
\makecell[r]{81.52\\(6.6)} &
\makecell[r]{80.83\\(4.1)} &
\makecell[r]{83.45\\(4.0)} &
\makecell[r]{77.57\\(2.0)} \\

& 1:1 &
\makecell[r]{65.21\\(6.2)} &
\makecell[r]{85.63\\(6.0)} &
\makecell[r]{73.78\\(4.7)} &
\makecell[r]{90.19\\(2.6)} &
\makecell[r]{74.04\\(4.6)} \\

\bottomrule

\end{tabular}
}
\end{table}





\end{document}